\documentclass{article}

\usepackage{arxiv}

\usepackage[utf8]{inputenc} 
\usepackage[T1]{fontenc}    
\usepackage{hyperref}       
\usepackage{url}            
\usepackage{booktabs}       
\usepackage{amsfonts}       
\usepackage{nicefrac}       
\usepackage{microtype}      
\usepackage{lipsum}
\usepackage{graphicx}
\usepackage{hyperref}
\usepackage{epstopdf, epsfig}
\usepackage{bm}
\usepackage{amsthm}
\usepackage{amsmath}
\usepackage{amssymb}
\usepackage[utf8]{inputenc}
\usepackage[T1]{fontenc}
\usepackage{mathptmx}
\usepackage[numbers]{natbib}
\usepackage{eucal}
\usepackage{color}
\usepackage{float}
\usepackage{subcaption}
\usepackage[ruled,linesnumbered]{algorithm2e}
\usepackage{natbib}
\usepackage{caption}
\SetKwComment{Comment}{$\triangleright$\ }{}

\title{Gated Linear Model induced U-net for surrogate modeling and uncertainty quantification}

\author{
  Sai Krishna Mendu \\
  Department of Electronics and Electrical Engineering\\
  Indian Institute of Technology Guwahati\\
  Guwahati - 781 039, Assam, India. \\
  \texttt{msk.mendu@gmail.com} \\
   \And
 Souvik Chakraborty \\
  Department of Applied Mechanics\\
  School of Artificial Intelligence\\
  Indian Institute of Technology Delhi\\
  Hauz Khas - 110 016, New Delhi, India \\
  \texttt{souvik@am.iitd.ac.in} \\
}

\begin{document}
\maketitle

\begin{abstract}
We propose a novel deep learning based surrogate model for solving high-dimensional uncertainty quantification and uncertainty propagation problems. The proposed deep learning architecture is developed by integrating the well-known U-net architecture with the Gaussian Gated Linear Network (GGLN) and referred to as the Gated Linear Network induced U-net or GLU-net. The proposed GLU-net treats the uncertainty propagation problem as an image to image regression and hence, is extremely data efficient. Additionally, it also provides estimates of the predictive uncertainty. The network architecture of GLU-net is less complex with 44\% fewer parameters than the contemporary works. We illustrate the performance of the proposed GLU-net in solving the Darcy flow problem under uncertainty under the sparse data scenario. We consider the stochastic input dimensionality to be up to $4225$. Benchmark results are generated using the vanilla Monte Carlo simulation. We observe the proposed GLU-net to be accurate and extremely efficient even when no information about the structure of the inputs is provided to the network. Case studies are performed by varying the training sample size and stochastic input dimensionality to illustrate the robustness of the proposed approach.   
\end{abstract}


\section{Introduction}\label{sec:intro}
If humans could visualize high dimensions, solving problems involving high dimensions (dimensions higher than three usually) would not be as challenging a task as it is now. Providing an uncertainty estimate for the solution of a high-dimensional problem by considering both the uncertainty in the input and the prediction is even more complicated. Due to the curse of dimensionality, specific thumb rules are proposed to look for while dealing with high dimensional problems, namely at least ten training points for every single dimension in the input. In such complex tasks, there is a need to know how confident we are in the prediction, and hence we argue that Uncertainty Quantification(UQ) is needed trying to solve a higher-dimensional problem with lesser data available. An uncertainty estimate for every spatial dimension estimate is helpful in making informed decisions while relying on the prediction.


We follow \cite{caldeira2020deeply} and \cite{jcgm2008evaluation} for defining uncertainty in a system. Epistemic uncertainty accounts for the uncertainty in the model prediction (or confidence in the prediction), and aleatoric uncertainty accounts for all types of uncertainties in the prediction arising due to the uncertainty in the input data. Whilst physicists follow a different convention and define statistical and systematic uncertainty, with statistical uncertainty as the uncertainty quantified by statistical analysis of a series of experimental measurements (variation observed by repeating the same experiment under similar conditions) and systematic uncertainty as possible unknown variation in measurement and are not altered by repeating the experiment under the same conditions. We summarize three types of uncertainties, aleatoric systematic, aleatoric statistical, and epistemic systematic uncertainties. We present the existing techniques and contemporary works for quantifying uncertainties below.

The sampling-based approach is a popular traditional approach for obtaining uncertainty estimates.
One variant is the well-known Monte Carlo Sampling (MCS) \cite{rubinstein2016simulation}. In MCS, the output uncertainty estimate is obtained by performing simulations at many sample points, and the Probability Density Function (PDF) is estimated. Although MCS is easy to implement, the convergence rate is quite slow, and computationally inefficient. Several developments to MCS include Latin Hypercube Sampling (LHS) \cite{tang1993orthogonal} and stratified sampling \cite{ericson1965optimum}; however, both the methods often suffer from the curse of dimensionality. A popular alternative is to use surrogate model based approaches. Popular surrogate models in the literature includes polynomial chaos expansion \cite{blatman2011adaptive}, Gaussian process \cite{bilionis2013multi,chakraborty2019graph}, analysis-of-variance decomposition \cite{chakraborty2017polynomial,chakraborty2017efficient}, and support vector machine \cite{roy2019support} among others.

Whilst conventional uncertainty estimation models have many disadvantages solving higher dimensional problems, deep learning offers a reliable solution \cite{chakraborty2021transfer,chakraborty2020simulation,kumar2021grade}. Uncertainty quantification and estimation is an active area of research in the machine learning community. Previous works broadly can be classified into, but obviously not limited to, Bayesian Neural Networks (Bayesian Deep Learning), dropout techniques (spatial dropout \cite{tompson2015efficient} and Bernoulli dropout \cite{boluki2020learnable}), ensemble techniques \cite{lakshminarayanan2016simple, valdenegro2019deep,wen2020batchensemble}, test time augmentation techniques \cite{shorten2019survey, wen2020time},  a model with an uncertainty module inscribed in it \cite{tsiligkaridis2021information,sensoy2018evidential,NEURIPS2019_73c03186}, reinforcement learning based UQ models \cite{zhao2019uncertainty,lee2018bayesian}. More about these works are given in  \cite{gawlikowski2021survey,abdar2021review}.

One of the main advantages of Bayesian deep learning resides in the fact that it allows us to quantify uncertainty and even study different forms of uncertainty. Frequentist learning approaches, on the other hand, provides point estimates of the unknown parameters and are only capable of quantifying the aleatoric uncertainty. We note that any frequentist approach can be converted into a Bayesian approach by assigning prior distribution to the unknown parameters and then updating it using Bayesian statistics. Unfortunately, the exact posterior is often intractable and hence, one has to rely on either rely on approximate algorithms (e.g., variational inference \cite{hoffman2013stochastic} or sample the posterior using Monte Carlo sampling (e.g., Markov Chain Monte Carlo or Hamiltonian Monte Carlo sampling methods \cite{papamarkou2019challenges,cobb2020scaling}).
This often renders the Bayesian deep learning model computationally expensive to train. On the other hand, Bayesian deep learning models are less prone to overfitting, robust to sparse training data, and provides an estimate of the predictive uncertainty due to limited and noisy data. Although Bayesian methods to solving inverse problems are on the rise, we want to bring forward the point that there is scarcity in the Bayesian deep learning based uncertainty quantification and uncertainty propagation literature \cite{lataniotis2019data}, specifically when dealing with a high-dimensional input space. 
The objective of this paper is fill this void by developing deep learning model that is (a) efficient to train, (b) scalable to high-dimensional problems, (c) works with sparse dataset, and (d) provides an estimate of the predictive uncertainty due to limited and noisy data.

To address the difficulties in the current literature for solving the high dimensional uncertainty quantification problems, we propose a new deep learning based surrogate model which produces accurate predictions and also provides a reliable estimate of the predictive uncertainty for the output (due to uncertainty in the model and the data). Our solution requires less than one training point per dimension to train (thereby vastly decreasing the training time) and predicts the solution accurately. 
Since often engineers and practitioners treat neural networks as black box models, there is a dire need of knowing the certainty of a prediction obtained using the same. We estimate the Probability Density Function (PDF) at each dimension for all the output fields given the input field. Our solution is less complex (architecture-wise) and has 44 $\%$ lesser parameters than the current SOTA method \cite{zhu2018bayesian}.

The remaining of this paper is organized as follows: In Section \ref{sec:ps}, we present the exact problem statement we are trying to solve. In Section \ref{sec:pf} we introduce GLU-Net by building it from scratch. In Section \ref{sec:iandr}, we analyze the use of GLU-Net in solving stochastic PDEs.
Finally, Section \ref{sec:conc} provides the concluding remarks.


\section{Problem setup}\label{sec:ps}


\subsection{Problem Statement}
Let us first define a probability space $(\Omega ,\mathcal F,{\mathcal  P})$, where $\Omega $ is sample space, $\mathcal F$ is a $\sigma$ algebra of subsets of $\Omega$ and ${\mathcal P}: \mathcal F \to [0,1]$, be a probability measure on $\Omega$. 
Let us assume a spatial $d$-dimensional bounded domain D $\subset {\Re ^d}$ with a boundary $\partial D$. A general form of an SPDE can be expressed as \begin{equation}\label{eq:eq1}
    \Gamma (\bm x,\bm \omega ;\bm y) = 0, \forall \bm x \in D,
\end{equation}
with a Boundary Condition (BC), 
\begin{equation}\label{eq:eq2}
    {\mathcal B}(\bm x,\bm \omega ;\bm y) = 0,\forall \bm x \in \partial D,
\end{equation}
where $\Gamma$  represents a partial differential operator and ${\mathcal B}$ is a boundary operator expressing the BC and $\bm \omega  \in \Omega$ is an event in the sample space, and $\bm y$ is the response for the SPDE. Usually the solution to the PDEs are infinite (thereby having a degree of freedom) and a BC is provided which constraints the infinite set of solutions to a finite (usually a single) number of solutions. The number of solutions to this SPDE may be finite, infinite or may not exist, which can be controlled by the BCs. We are particularly interested in solving physical systems which are theoretically modelled by SPDEs. 

Now, consider a spatial location set $\bm S \subset {\Re^{{d_s}}}$, be our interest locations for the solution (referred as index set). Physical systems modelled by SPDEs with a solution $\bm y(s, \bm \xi( s, \bm \omega))$, where $ s\in S$, with $\bm \xi(s, \bm \omega)$ being a realization of the input random field $\{ \bm \xi(s,\bm \omega ),s \in S,\omega  \in \Omega \}$, where $\Omega$ is the sample space. Here, the random field is the variable appearing in the general form of an SPDE as defined above.

We consider a very interesting problem that is, steady state fluid flow through a porous media (non uniform variable permeability field), in a two dimensional unit square spatial domain. We formally present the problem statement here. Let us consider a random permeability field ${ K}$ on a unit square space domain $S = {[0,1]^2}$ taken in the first quadrant on the $X-Y$ plane. The single phase, steady state flow of the fluid through porous media has two properties: the pressure field $p$ and the velocity field $\bm u$, with ${K}$ being permeability of the porous media on $S$. The Darcy's Law is given by 
\begin{equation}\label{eq:eq3}
\begin{split}
    \bm u(s) & =  -  K(s) \bm {\nabla} p(s),\qquad s \in S \\ 
    \nabla .\bm u(s)  & =  f(s), \qquad s \in S \\  
    \bm u(s).\hat{\bm n}(s) & = 0, \qquad s \in \partial S  \\ 
    \int\limits_S {p(s)ds}  & = 0, 
\end{split}
\end{equation}
where $\hat{\bm n}$ is the unit normal vector to the boundary of $S$ and $\bm f$ is an injection well model on the bottom left corner of $S$ and a production well on the top right corner of $S$. We enforce two Boundary Conditions, the first one being no-flux boundary condition and second one leading to the solution uniqueness, as defined below. So, 
\begin{equation}\label{eq:eq4}
  f(s)=\begin{cases}
    r, & \text{if $|{s_i} - \frac{1}{2}w| \le \frac{1}{2}w$, for $i = 1,2$}.\\
    -r, & \text{if $|{s_i} - 1 + \frac{1}{2}w| \le \frac{1}{2}w$, for $i = 1,2$}.\\
    0, & \text{otherwise}.
  \end{cases}
\end{equation}
where $r$ is the rate of wells and $w$ is the size. 
We note that the permeability $K(s) $ in Eq. \eqref{eq:eq4} is modeled as a stochastic field, and hence, the underlying governing equation is an SPDE. The objective here is to develop an efficient data-driven deep learning algorithm for solving the SPDE shown in Eq. \eqref{eq:eq4}.

\subsection{Data generation}
For data generation, we consider the Log  permeability field to be a Gaussian random field such that
\begin{equation}\label{eq:eq5}
    K(s) = \exp (G(s)), \qquad G(.) \sim {\mathcal N}(m,k(\cdot,\cdot))
\end{equation}
where $m$ is the mean and $k$ is the covariance function such that
\begin{equation}\label{eq:eq6}
    k(s,{s'}) = \exp ( - ||s - {s'}||/l)    
\end{equation}
where $||s||$ is the L2-Norm. 
we discretize over the square space domain $S$, by forming a grid of 65X65 over the unit square. During this process, the permeability field does jump due to the nature of the kernel function that is  chosen. We use the dataset provided by \cite{zhu2018bayesian}, with $l = 0.1$, mean of the kernel $m = 0$, rate of the source $r = 10$ and size $w=0.125$. 

Now that we have defined the exact problem we are trying to solve, it is important to look into the mapping we are trying to establish.
The permeability $K$ modelled as a log-normal random field discretized over the grid $S$ is our input.
Redefining $\bm S$, $\bm S = \{ {s_{{n_1}}},...{s_{{n_s}}}\}$ to be a set of points in the spatial 2-D domain over the $65 \times 65$ grid, the random permeability field $K$ can be classified as a random vector $\bm \Xi$ where $\bm \Xi \in {\Re ^{{n_s}\cdot{d_x}}}$. Our objective is to learn a mapping $ \bm \Xi \mapsto \mathbf Y$. $\bm Y = \left[ \bm P, \bm U, \bm V \right] \in \Re^{{n_s}\cdot{d_y}}$ represents the response vector with $\bm P \in \Re ^{{n_s}\cdot{d_x}}$, $\bm U \in \Re ^{{n_s}\cdot{d_x}}$,
and $\bm V \in \Re ^{{n_s}\cdot{d_x}}$ representing random output vectors obtained by discretizing the random fields $p$, $u$, and $v$ over the spatial grid $\bm S$. For the Darcy flow problem, we have one input field $K$ and three output fields, $u$, $v$, and $p$. Therefore, $d_x = 1$ and $d_y = 3$. Since we are dealing with a $65\times 65$ grid, $N_s = 65\times 65$. Since, we are interested in developing a data efficient Bayesian surrogate model, we treat the underlying problem as an image to image regression task, 
\begin{equation}\label{eq:obj}
    \mathcal M_1: \Re ^{65\times 65 \times 1} \mapsto \Re ^{65\times 65 \times 3} .
\end{equation}
With this, the objective is to develop a Bayesian deep learning algorithm that can accomplish the task in Eq. \eqref{eq:obj}.

\begin{figure}[ht!]
    \centering
    \subfloat[\centering KLE50 Permeability]{{\includegraphics[width=4.1cm,height=4.1cm]{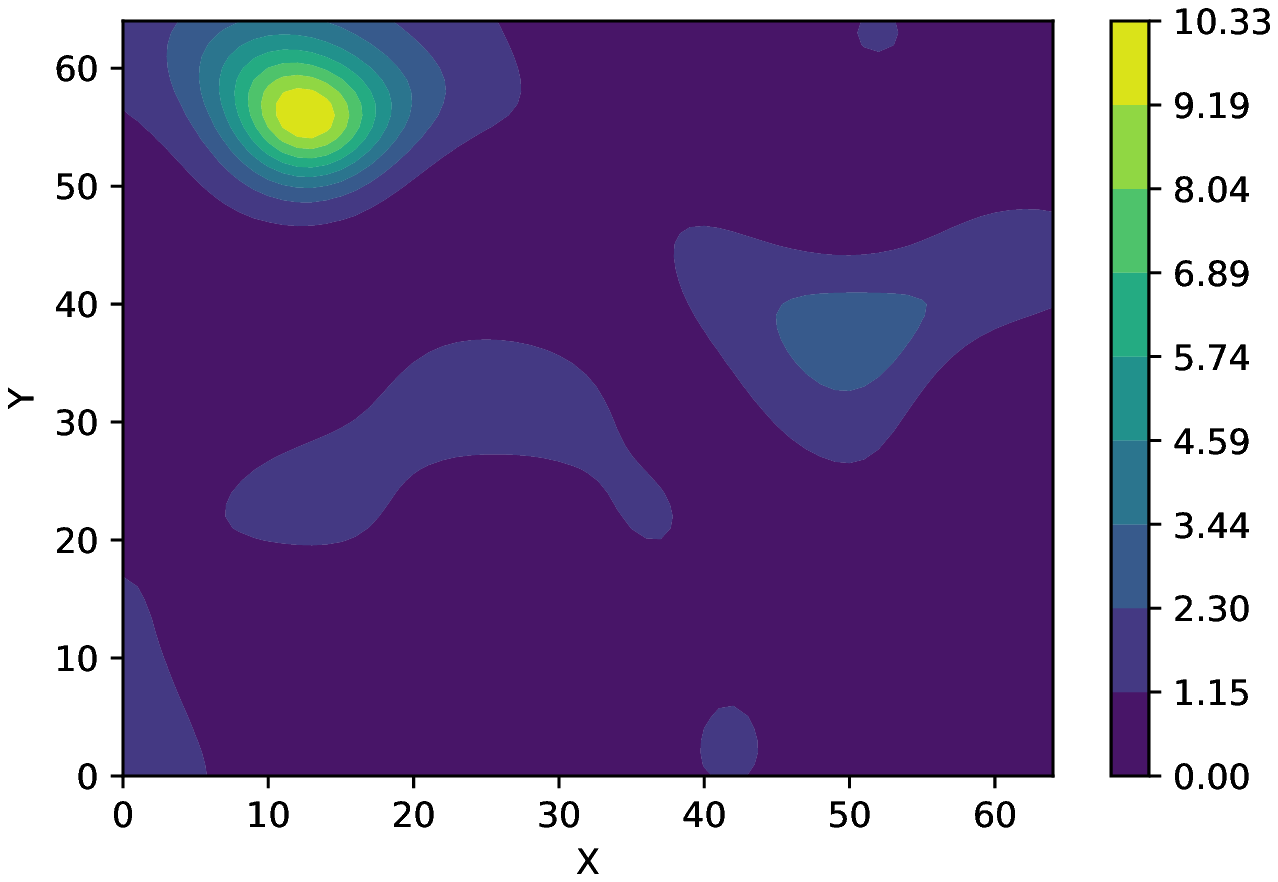}}}%
    \hspace{0.01cm}
    \subfloat[\centering KLE50 Pressure]{{\includegraphics[width=4.1cm,height=4.1cm]{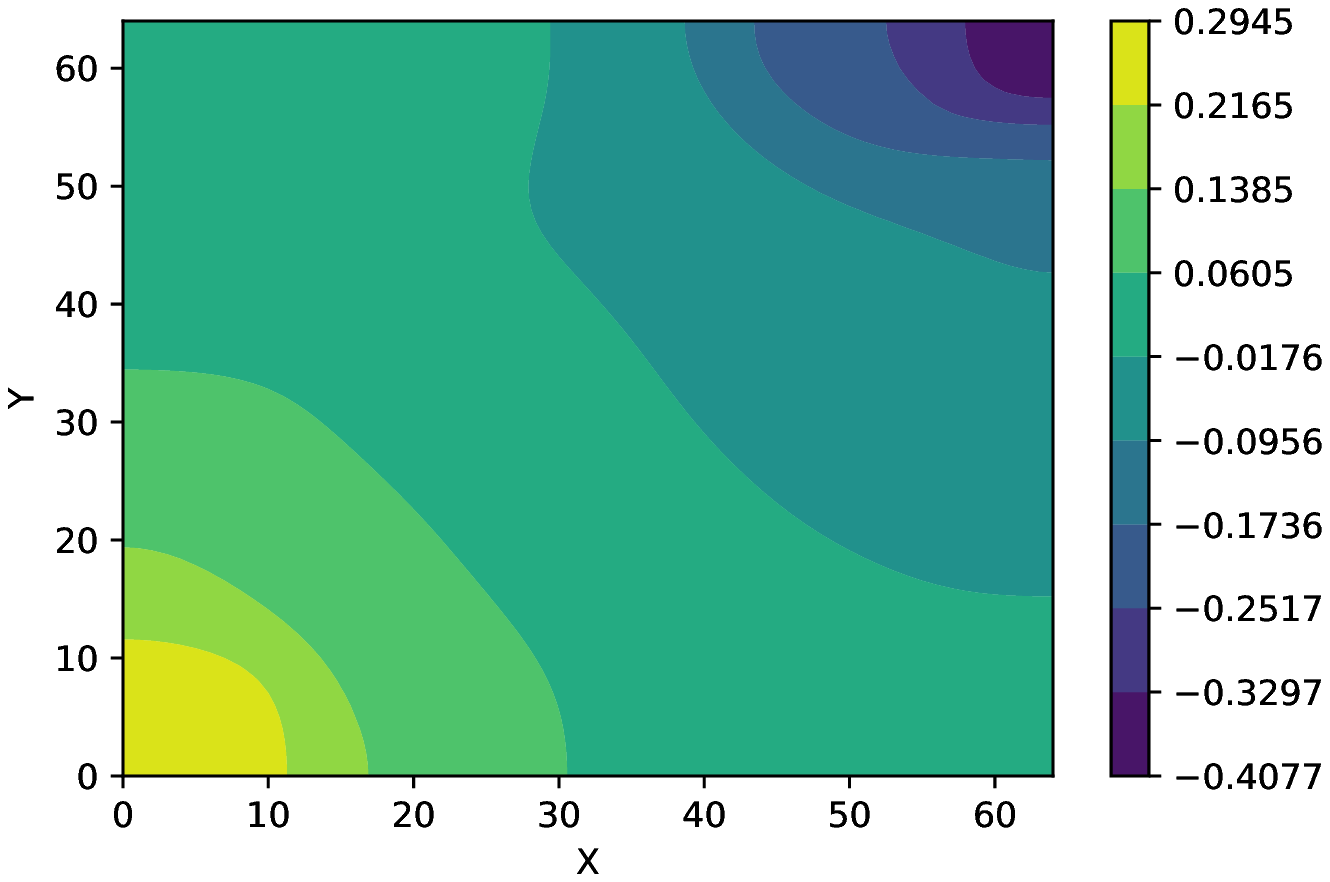}}}%
    \hspace{0.01cm}
    \subfloat[\centering KLE50 Velocity(X)]{{\includegraphics[width=4.1cm,height=4.1cm]{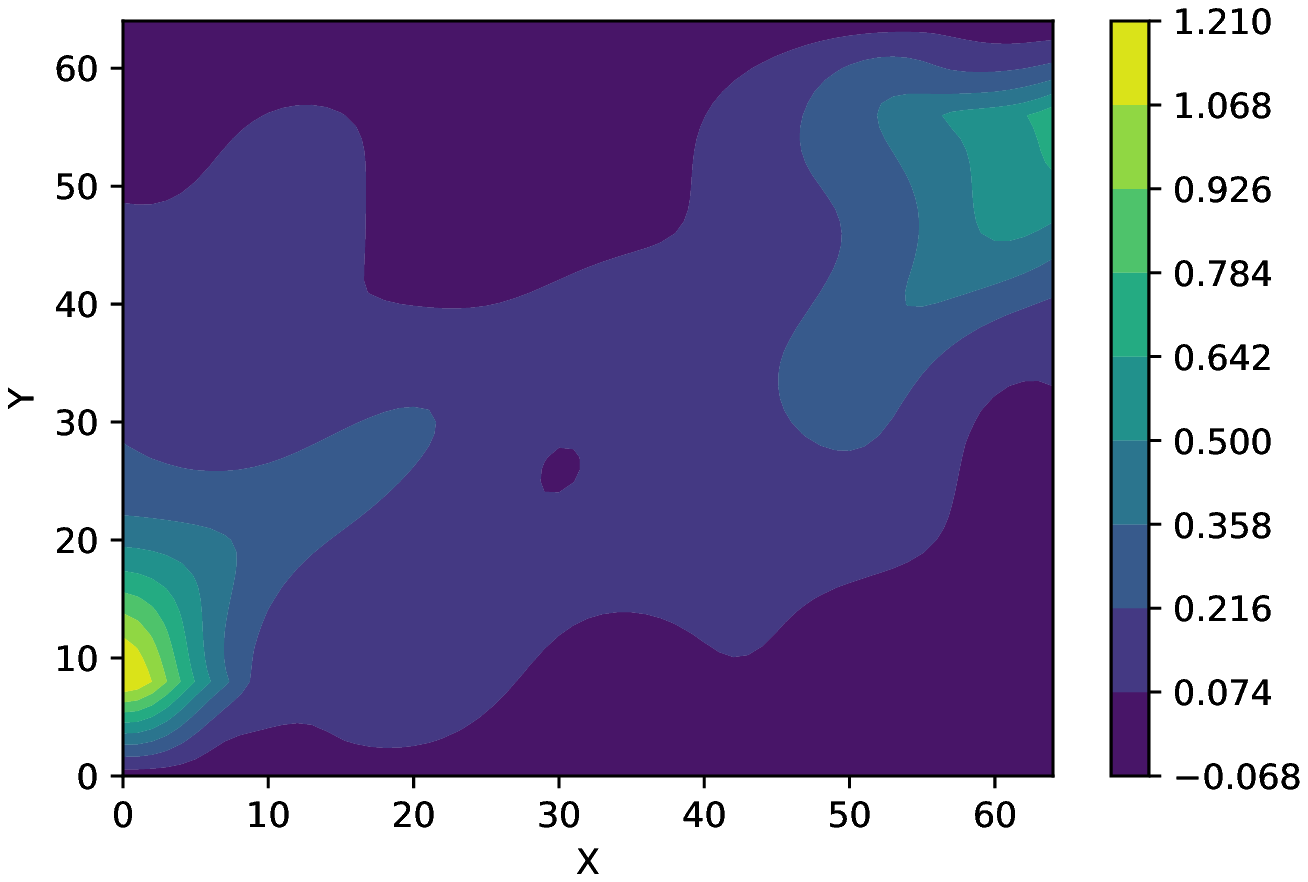}}}%
    \hspace{0.01cm}
    \subfloat[\centering KLE50 Velocity(Y)]{{\includegraphics[width=4.1cm,height=4.1cm]{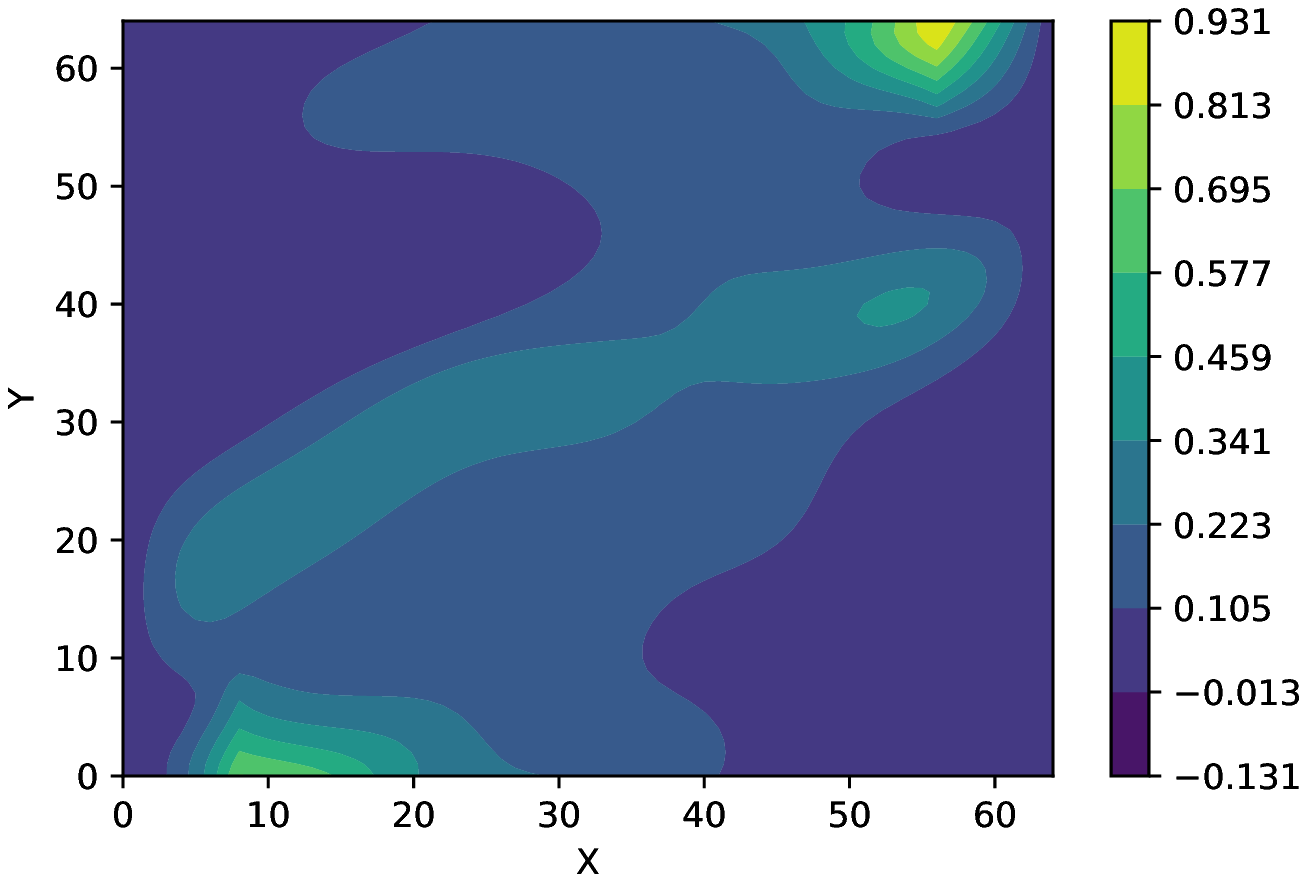}}}%
    \hspace{0.01cm}
    \subfloat[\centering KLE500 Permeability]{{\includegraphics[width=4.1cm,height=4.1cm]{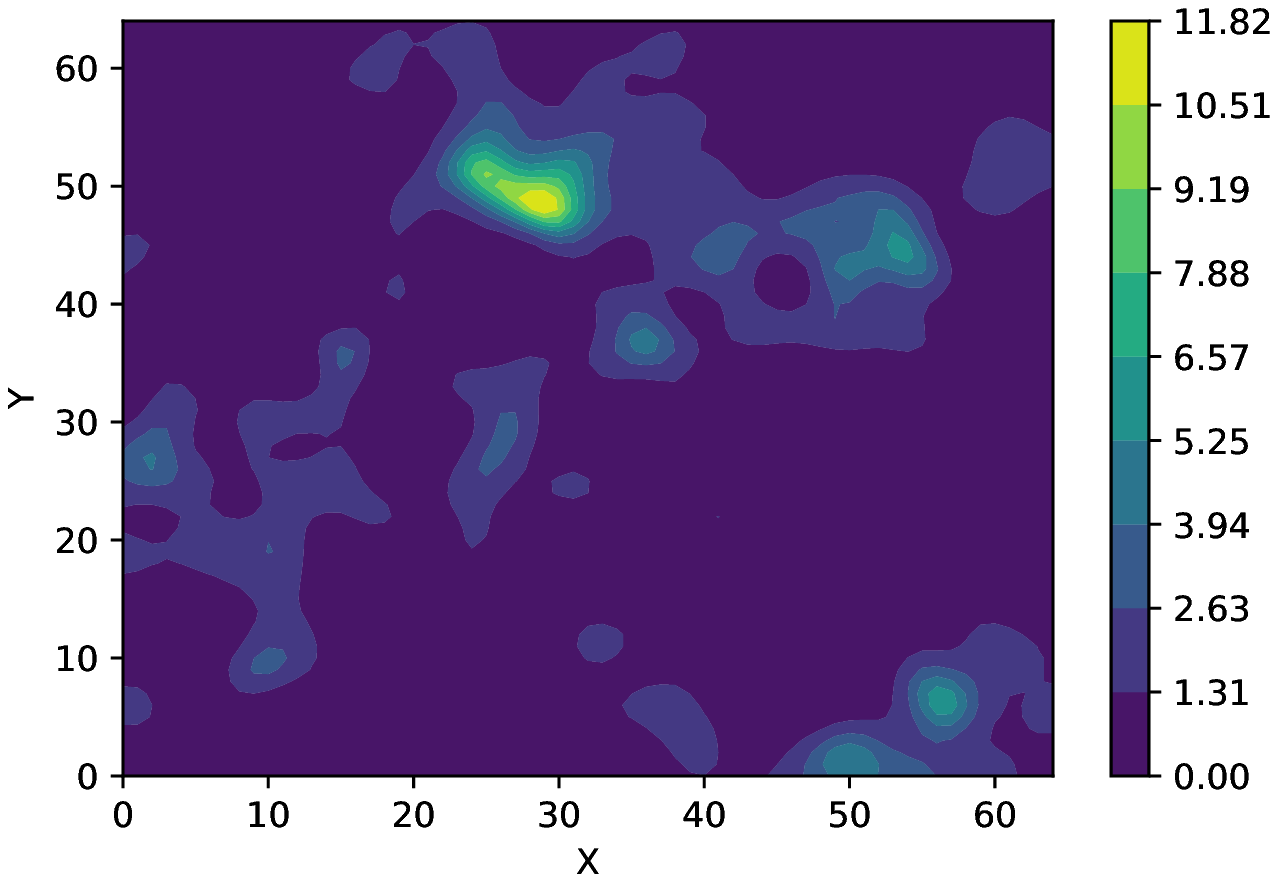}}}%
    \hspace{0.01cm}
    \subfloat[\centering KLE500 Pressure]{{\includegraphics[width=4.1cm,height=4.1cm]{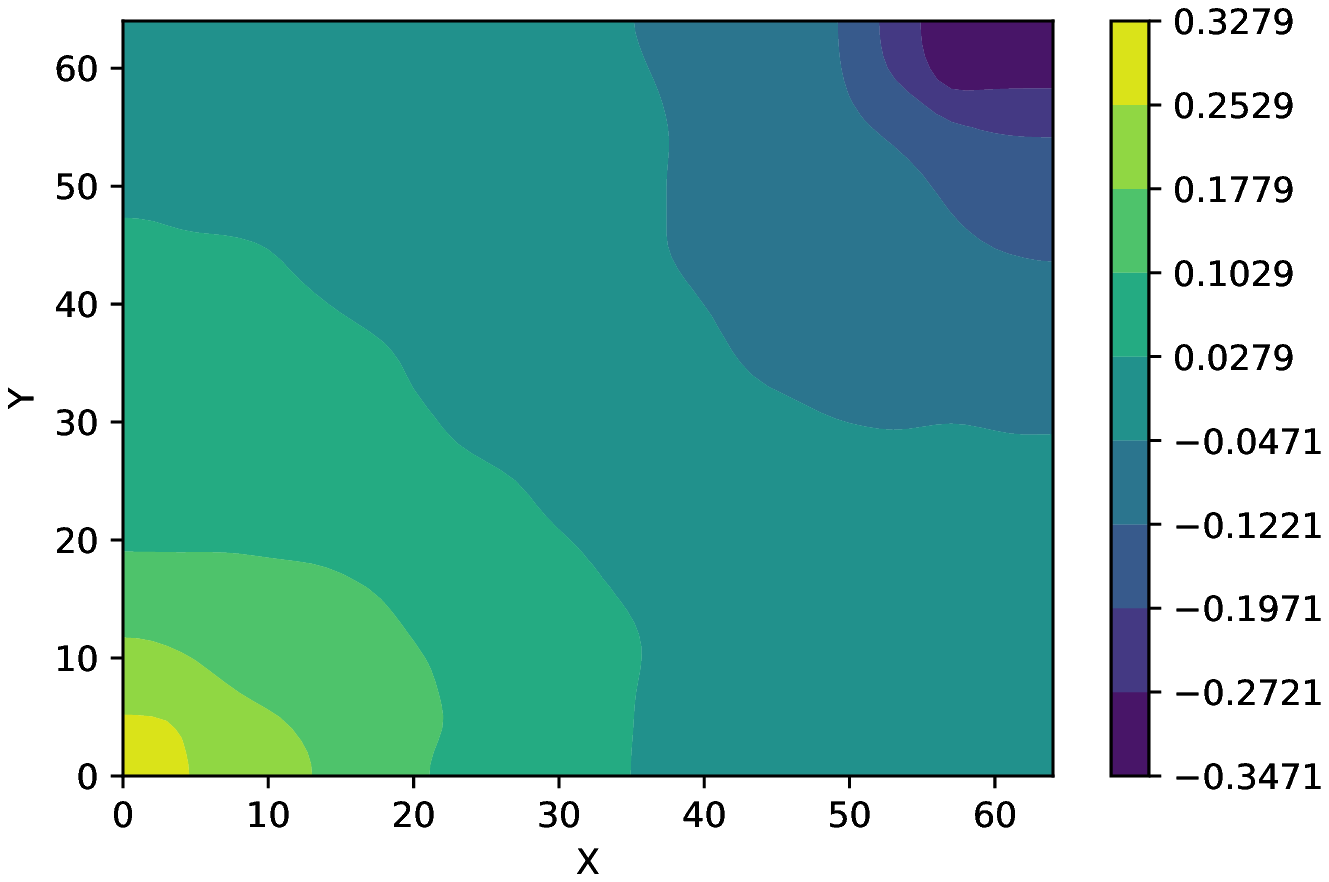}}}%
    \hspace{0.01cm}
    \subfloat[\centering KLE500 Velocity(X)]{{\includegraphics[width=4.1cm,height=4.1cm]{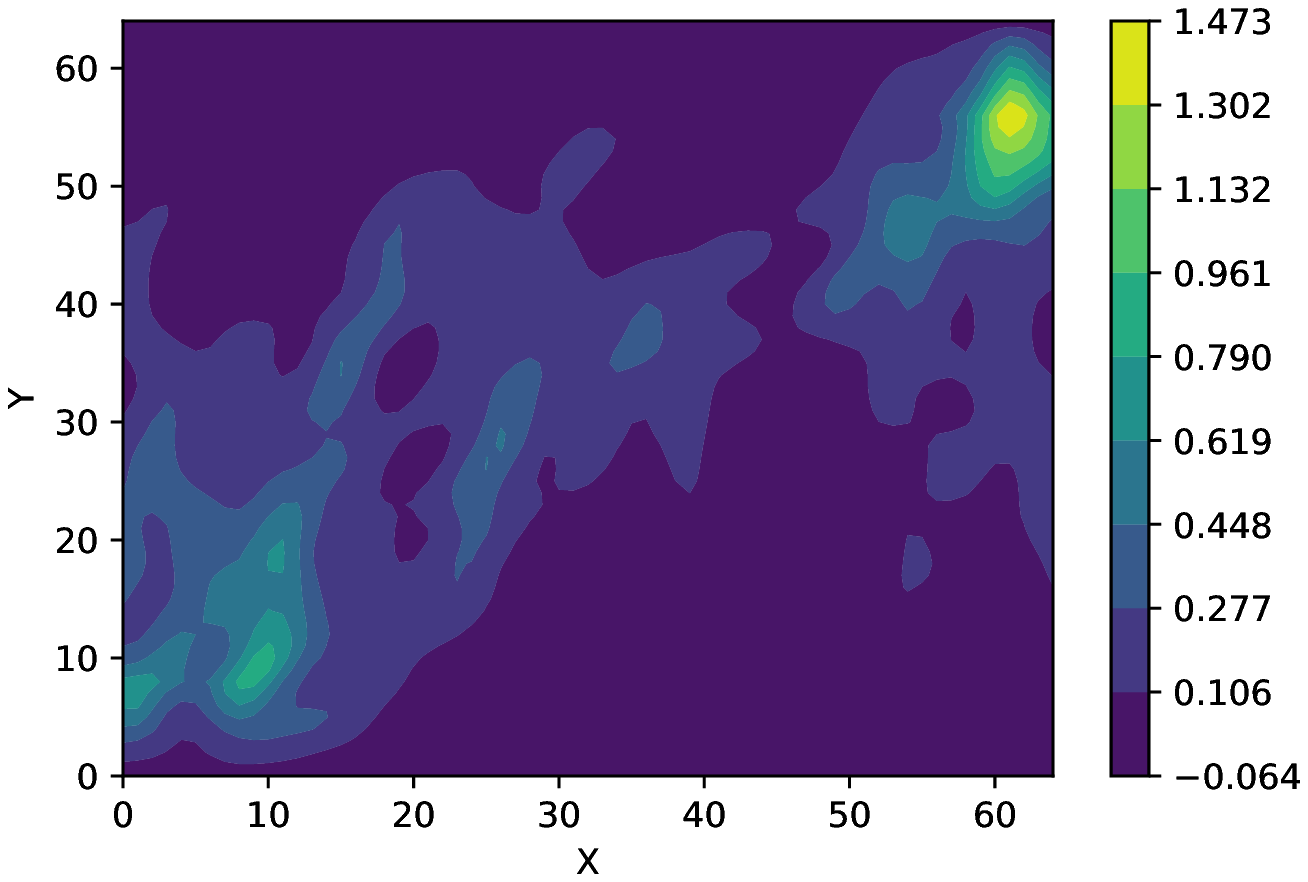}}}%
    \hspace{0.01cm}
    \subfloat[\centering KLE500 Velocity(Y)]{{\includegraphics[width=4.1cm,height=4.1cm]{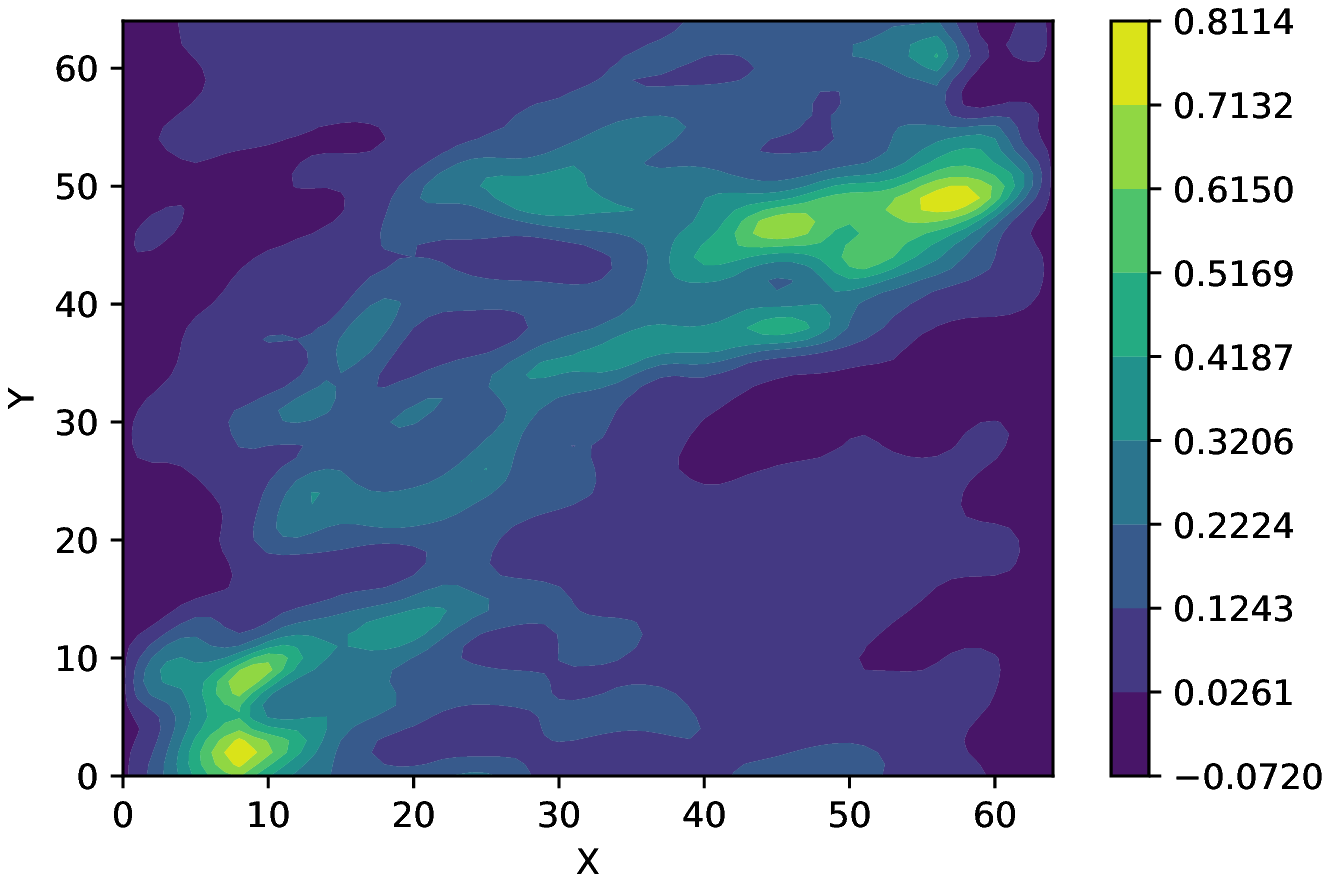}}}%
    \hspace{0.01cm}
    \subfloat[\centering KLE4225 Permeability]{{\includegraphics[width=4.1cm,height=4.1cm]{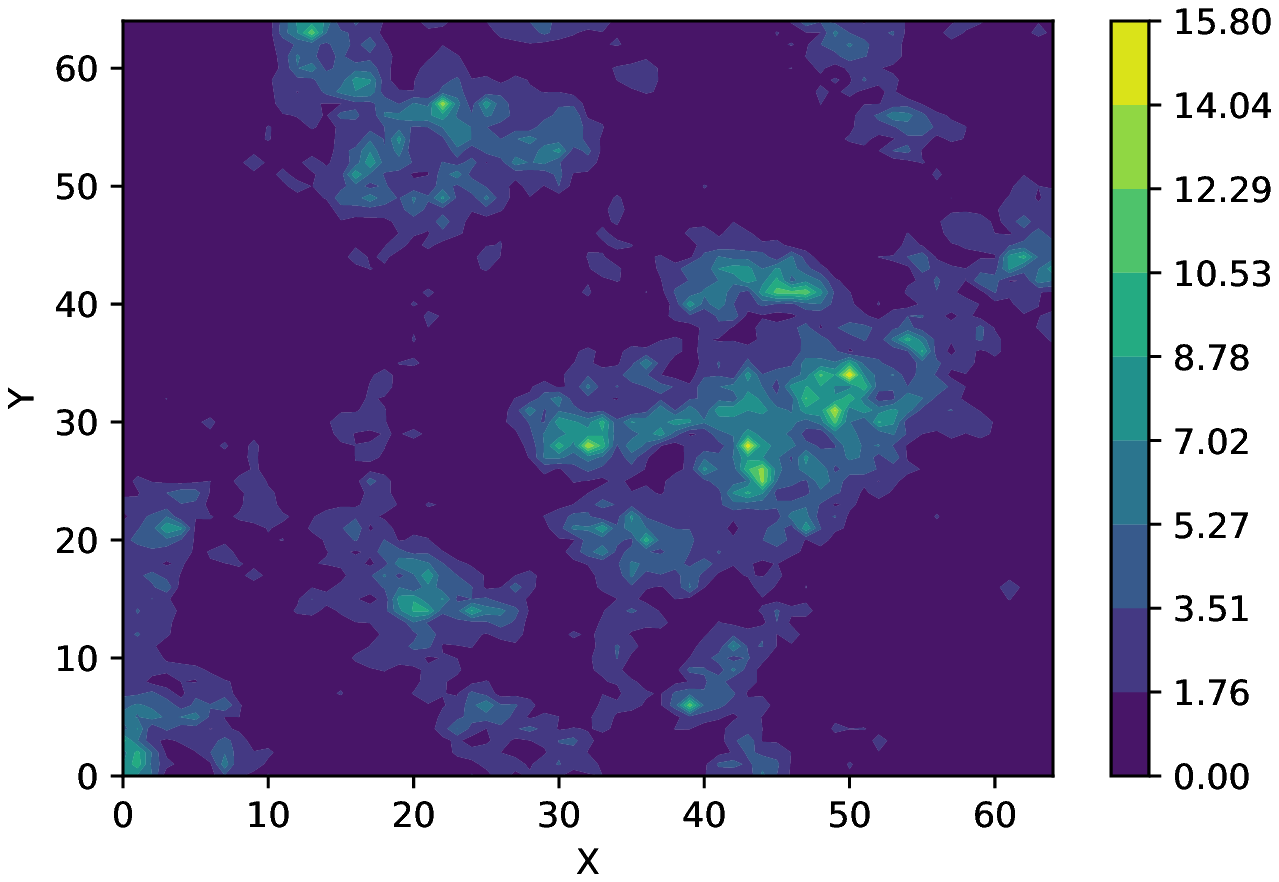}}}%
    \hspace{0.01cm}
    \subfloat[\centering KLE4225 Pressure]{{\includegraphics[width=4.1cm,height=4.1cm]{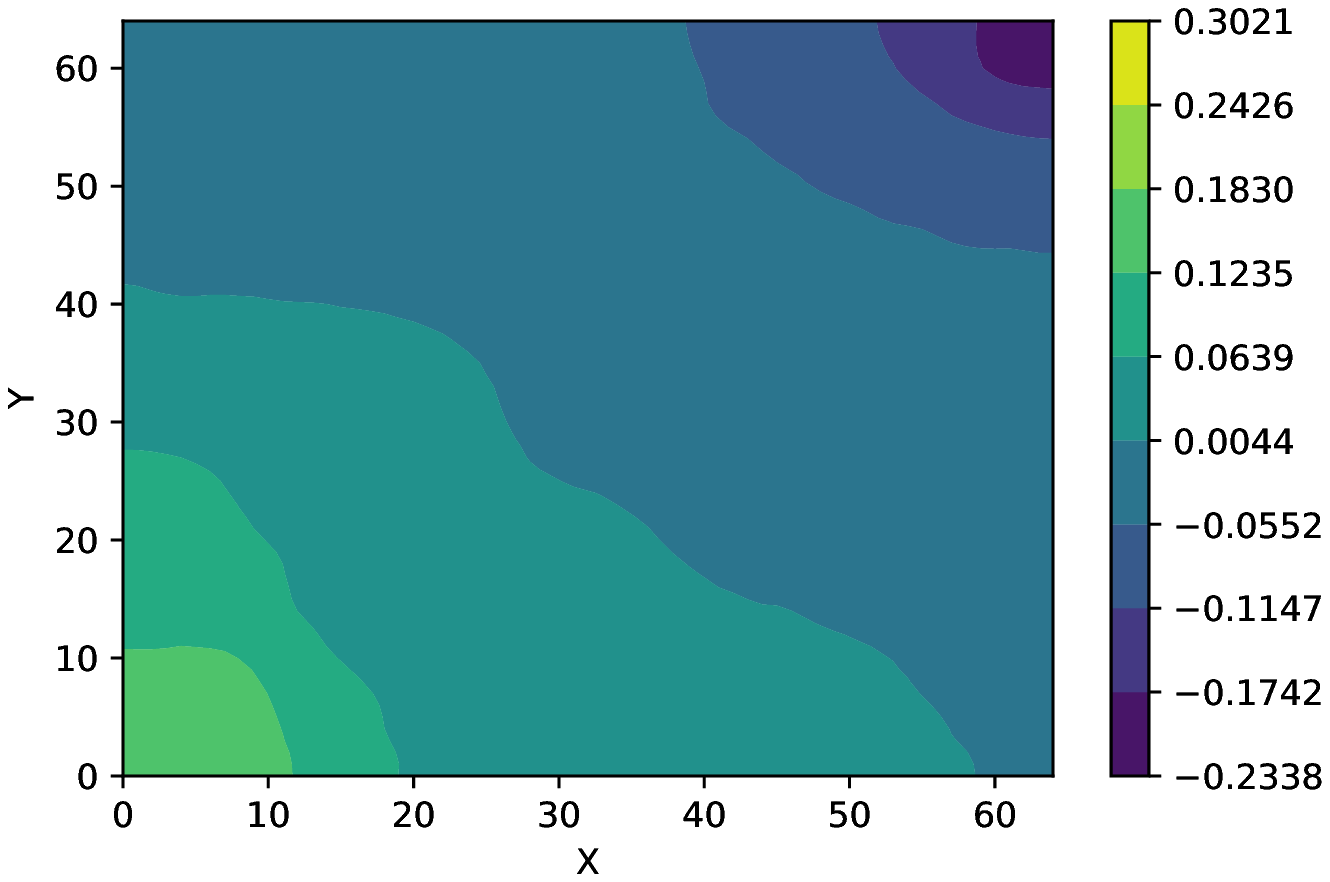}}}%
    \hspace{0.01cm}
    \subfloat[\centering KLE4225 Velocity(X)]{{\includegraphics[width=4.1cm,height=4.1cm]{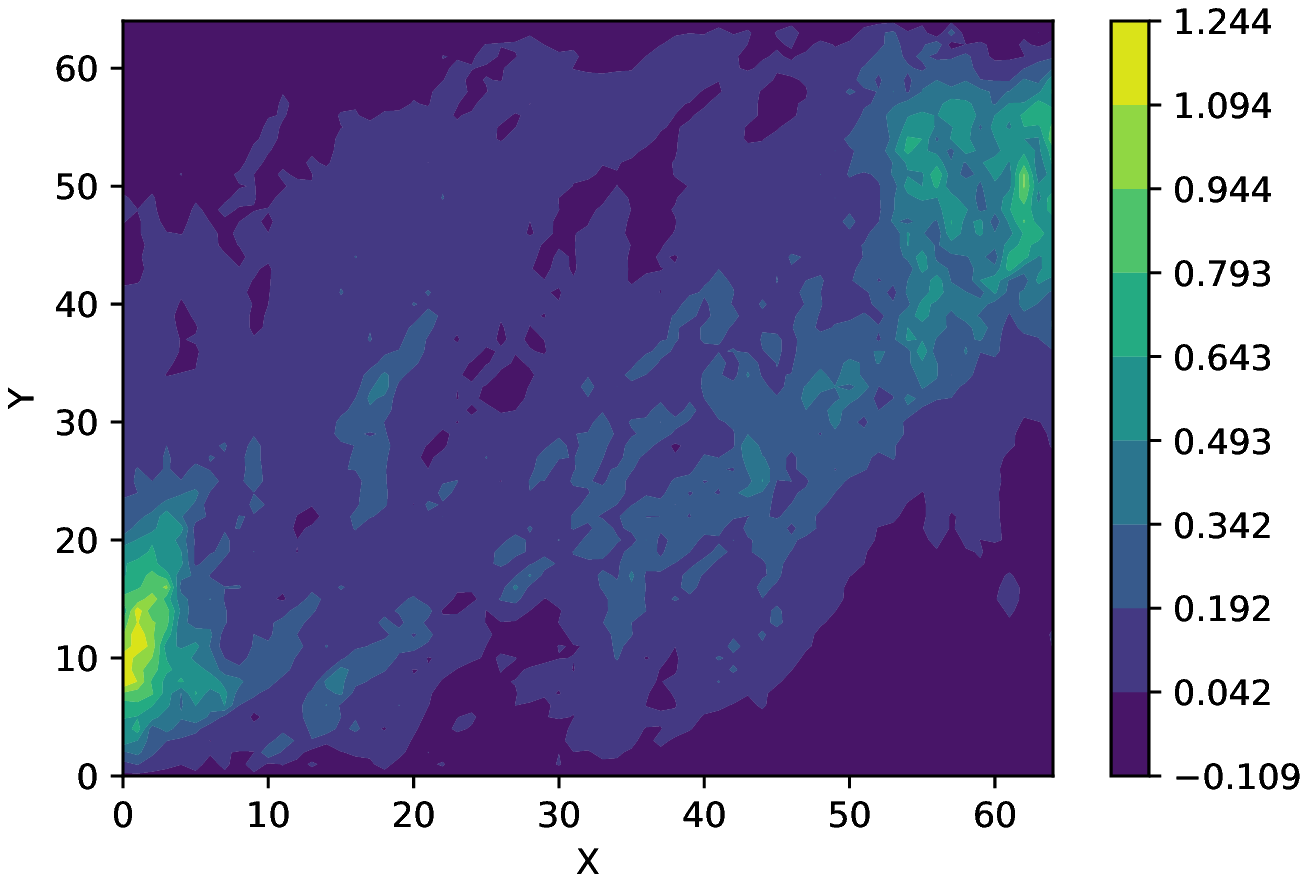}}}%
    \hspace{0.01cm}
    \subfloat[\centering KLE4225 Velocity(Y)]{{\includegraphics[width=4.1cm,height=4.1cm]{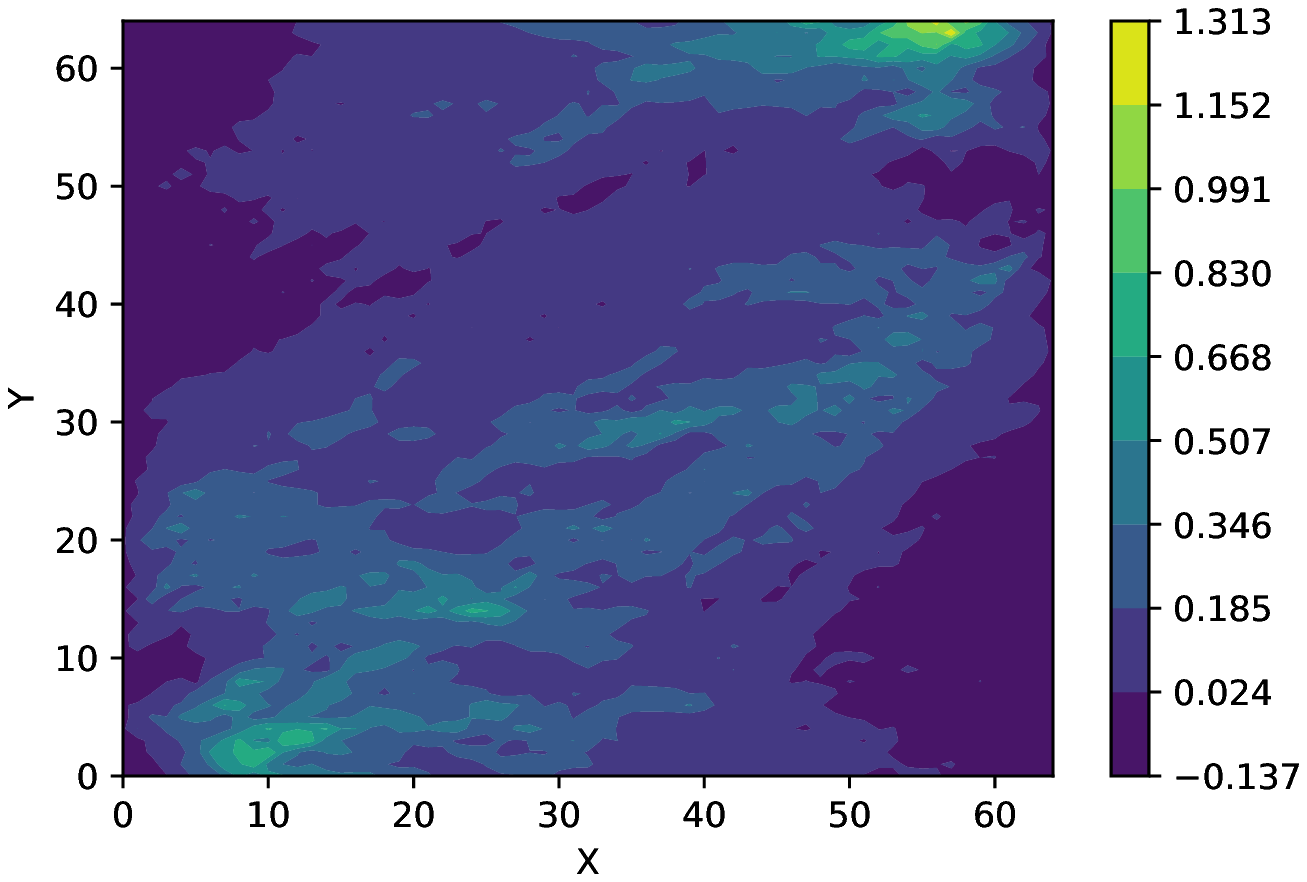}}}%
    \caption{Sample realization of the input and the output fields.
    The first, second, and third row corresponds to KLE50, KLE500, and KLE4225 respectively.}%
    \label{fig:fig1}%
\end{figure}

In order to control the effective dimensionality of the input data, we use Karhunen–Loève expansion (KLE). Following \cite{zhu2018bayesian}, three cases corresponding to KLE50, KLE500, and KLE4225 have been considered. Realizations of the random log-permeability field generated using KLE are shown in Fig. \ref{fig:fig1}. It is evident that as the dimensionality of KLE increases, the input field becomes more and more complex. Accordingly, the complexity in the three response variables also increases. 

\section{Proposed framework}\label{sec:pf}

We introduce GLU-Net, a network that can tackle high dimensional input even with lesser data and can achieve good generalization accuracy, and produces the solution to the collective image to image regression problem posed by our generalized PDE system. The model primarily has two components, one being the Gaussian Gated Linear Network (GLN) and the other one is the well-known U-Net architecture. We first discuss the fundamentals of U-net and GLN, and then proceed to discussing how the two are coupled to develop the proposed GLU-net.

\subsection{U-net}

\begin{figure}[ht!]
	\centering
	\includegraphics[width = 1\textwidth]{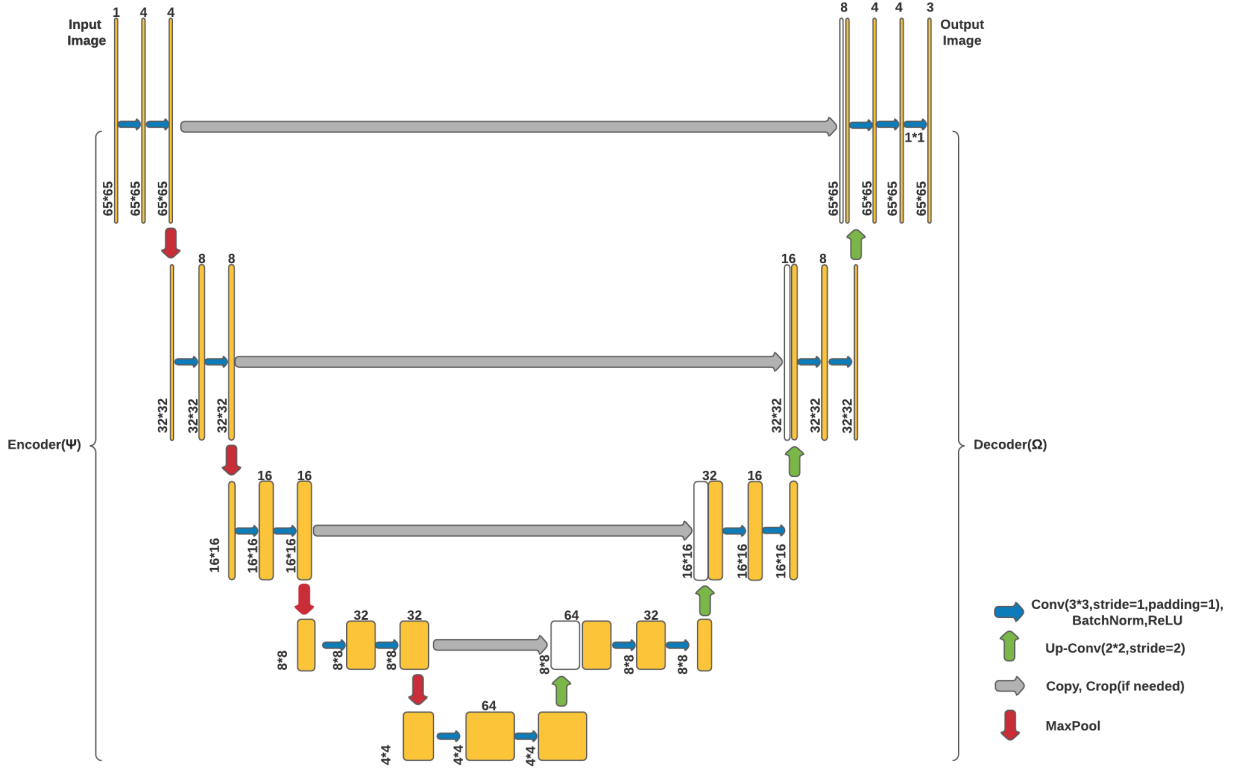}
	\caption{Schematic of the variant of the U-Net Architecture used in GLU-Net. The Encoder($\psi$)'s outputs at each layer are sent into that corresponding layer of the decoder($\Omega$). The current dimension of the data at every layer is shown at the bottom left of every layer. The number of channels in each layer is depicted on the top of that corresponding layer. Each arrow's descriptions are added in the bottom right corner.}
	\label{unet}
\end{figure}

U-Net \cite{ronneberger2015u} is an encoder-decoder styled convolutional neural network (CNN), with a skip connection between a layer in the encoder and its corresponding counterpart in the decoder. The encoder encodes key parts of the information into lesser dimensional space, and the decoder decodes that information and learns the required PDE. The input of every layer in the decoder is the concatenation of the output of the corresponding encoder layer and the output of the previous decoder layer. This input goes through a series of convolution layers, where a learnable filter gets convolved over the input feature maps and eventually sent to the next layer of the decoder using an up convolution operation (can be thought of as a reverse operation of convolution). Rectified linear unit \cite{nair2010rectified} is used as activation function. 
Compared to the original U-net, we make the following alterations:
\begin{itemize}
    \item We alter the number of filters in the original architecture by halving the number of filters at every layer.
    \item We also slightly alter the convolutional layers by applying a batch normalization layer for the weights of the convolutional layer.
\end{itemize}

The basic premise of any Encoder-Decoder architecture is to compress the input dimensions into a latent dimension keeping the large and small distance patterns intact. The decoder architecture builds and learns from the latent dimension. A schematic representation of the U-net architecture is shown in Fig. \ref{unet}.


\subsection{GLN}

In Gated Linear Network(GLN) \cite{veness2019gated}, we allow every layer to predict the target. The input to the network is referred to as context, and its intuition is explained later. Two variants of GLN exist in the literature, Bernoulli GLN \cite{veness2019gated}, and Gaussian GLN \cite{budden2020gaussian}. As the name suggests, Bernoulli GLN predicts the target as a Bernoulli distribution, and Gaussian GLN predicts the output as a Gaussian distribution. In our case, the target variables are continuous, and hence, we limit ourselves to Gaussian GLN only. In Gaussian GLN, every neuron receives every previous layer neurons' output normal distributions in a fully connected fashion and performs \textit{Gaussian mixing} as discussed below.

Let us consider $m$ univariate Gaussian distributions and assume they are expressed as  $f_i(y) = \mathcal N (\mu_i,\sigma_i^2)$ where  $ i\in {1,2,...m}$,
\begin{equation}\label{eq:eq9}
    {f_i}(y) = \frac{1}{{\sqrt {2\pi {\sigma ^2}} }}\exp \left\{ { - \frac{1}{2}{{\left( {\frac{{x - \mu }}{\sigma }} \right)}^2}} \right\}.
\end{equation}
Now given a weight vector $w = \left( {{w_1},{w_2},...,{w_m}} \right) \subset R_ + ^m$, a weighted Product of Gaussians (PoG) is defined as follows, 
\begin{equation}\label{eq:eq10}
    Po{G_w}(y;{f_1}(.),...{f_m}(.)): = \frac{{\prod\limits_{i = 1}^m {{{[{f_i}(y)]}^{{w_i}}}} }}{{\int {\prod\limits_{i = 1}^m {{{[{f_i}(y)]}^{{w_i}}}dy} } }}
\end{equation}
It is trivial to proof that the above definition outputs a Gaussian distribution with mean $\mu_{PoG}(w)$ and variance $\sigma_{PoG}^2(w)$,
\begin{equation}\label{eq:eq12}
    {\mu _{PoG}}(w): = \sigma _{PoG}^2(w){\left[ {\sum\limits_{i = 1}^{m} {\frac{{{w_i}{\mu _i}}}{{\sigma _i^2}}} } \right]},
\end{equation}
\begin{equation}\label{eq:eq11}
     \sigma _{PoG}^2(w): = {\left[ {\sum\limits_{i = 1}^{m} {\frac{{{w_i}}}{{\sigma _i^2}}} } \right]^{ - 1}}.    
\end{equation}
Each of the weight $w_i$ is constrained to be in the set $W$, where 
\begin{equation}\label{eq:eq13a}
        W: = \{ w \in {[0,b]^m}:||w|{|_1} \ge \varepsilon \},    
\end{equation}
with $0 < \varepsilon  < 1$ and $b \ge 1$. We define vectors $\eta$ and $\mu$
\begin{equation}
    {\eta _i}: = {w_i}/\sigma _i^2,\eta = ({\eta_1},...{\eta_m}),
\end{equation}
\begin{equation}
    {\mu}: = ({\mu _1},...{\mu _m}).
\end{equation}
Now knowing the fact that weights are positive as defined above, the PoG terms can be expressed as, 
\begin{equation}
    \sigma _{PoG}^2 = ||\eta ||_1^{ - 1}
\end{equation}
and
\begin{equation}
    {\mu _{PoG}} = {\eta ^T}\mu /||\eta |{|_1},
\end{equation}
where $||w|{|_1}$ represents the L1-Norm of the vector $w$.
Given the fact that the goal during training is to push the mean $\mu_{PoG}\left(\cdot\right)$ towards the target $y$ and minimize the variance, the loss-function in this case is defined as \cite{veness2019gated,budden2020gaussian}
\begin{equation}\label{eq:eq13}
        l(y;w): = \log \sigma _{PoG}^2(w) + \frac{{{{(y - {\mu _{PoG}})}^2}}}{{\sigma _{PoG}^2(w)}}.    
\end{equation}
Rewriting the expression in the form of vectors $\eta$ and $\mu$, we obtain
\begin{equation}\label{eq:eq14}
     l(y;\eta ) =  - \log ||\eta |{|_1} + {(x - {\eta ^T}\mu /||\eta |{|_1})^2}||\eta |{|_1}.
 \end{equation}
We note that the loss function in Eq. \eqref{eq:eq14} is convex under the weight constraint and differentiable (\ref{app:app2}). In our implementation, we use the automatic differentiation library of PyTorch \cite{paszke2019pytorch}. 

The weight selection mechanism for the above Gaussian mixing process is based on the half space  gating  procedure. Given a vector $\bm v \subset {R^n}$ and a constant $b \subset R$, let us consider a hyperplane  $ H \{ z \in {R^n}:z.v \ge b\}$, the half space context function is 
\begin{equation}
    c(z):= \left\{\begin{array}{ll}
        1 & \text{  if } z \subset H\\
        0 & \text{  elsewhere }
    \end{array}\right.
\end{equation}
This is equivalent to dividing the entire ${R^n}$ subspace into two halves, hence the name, half space gating. In a similar fashion, we can have $m$ context function, and can combine them into a higher order context function given by, $\bm c(z) = ({c_1}(z),\ldots, {c_m}(z))$. Here $m$ is defined as the context dimension.

\begin{figure}[ht!]
	\centering
	\includegraphics[width = 12cm,height=9cm]{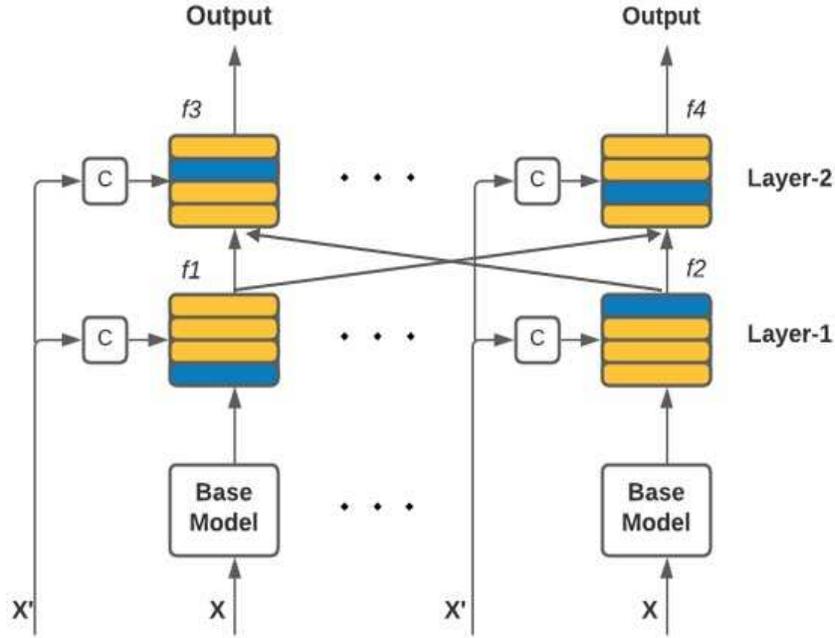}
	\caption{Schematic of how a Gated Linear Network works. Each $C$ module is a Half space gating module, used to select the weights for the Gaussian Mixing Process. $X'$ and $X$ being the normalized input($\frac{{X - \mu }}{\sigma }$) to the model. Each of the modules ${f_1}(\cdot)$,..${f_2}(\cdot)$ and ${f_3}(\cdot)$,..${f_4}(\cdot)$ tries to predict the target $y$, with the weights chosen. Every neuron's output in layer-1 is sent to layer-2.}
	\label{gln}
\end{figure}
A schematic representation of the GLN model is shown in Fig. \ref{gln}. The context for each of the neurons in every layer is the input of the network. For the first layer, the mean and the variance is chosen according to a base model. We have chosen our base model to be identified in the sense that the mean of the first layer would be the input itself, and the variance is a constant of 0.01. This gets fed to the next layer, which mixes these according to weights generated by the context function, which in turn takes the input to the network as the input. Each neuron in the Gaussian GLN, has a set of weights corresponding to each of the higher context state $\bm c(z) = ({c_1}(z),\ldots,{c_m}(z))$. So, we have $m$ bits in the $\bm c(z)$ context state, leading to ${2^m}$ such states. It is now evident that $m$ being the context dimension generates a weight matrix of dimension ${2^m}\times k$, where $k$ is the number of neurons in the previous layer (number of input  Gaussian distributions for the current layer). Initially, weights are sampled from a Gaussian distribution $\mathcal N (0,1)$. The $z,b$ value for each of the hyperplanes are chosen from Gaussian distribution $\mathcal N(0,1)$ and is such that, if the input context for each of the neuron is a distribution of a mean of 0 and variance of 1, we get a required partition as per the hyperplane, otherwise, the selected plane may always be fit into a single partition, and only that weight may be selected again and again. 

Though we are set to solve an image to image regression problem, GLNs are useful for tabular data modelling problems. Similar to Gaussian GLN, GLNs are also defined for Bernoulli distributions as in \cite{veness2019gated} which would be helpful in classification tasks. The property of weight selection on the basis of the input subspace and the property of fully connected neurons helps GLNs attain superior performance.

\subsection{GLU-Net}
\begin{figure}[ht!]
	\centering
	\includegraphics[width = 1\textwidth]{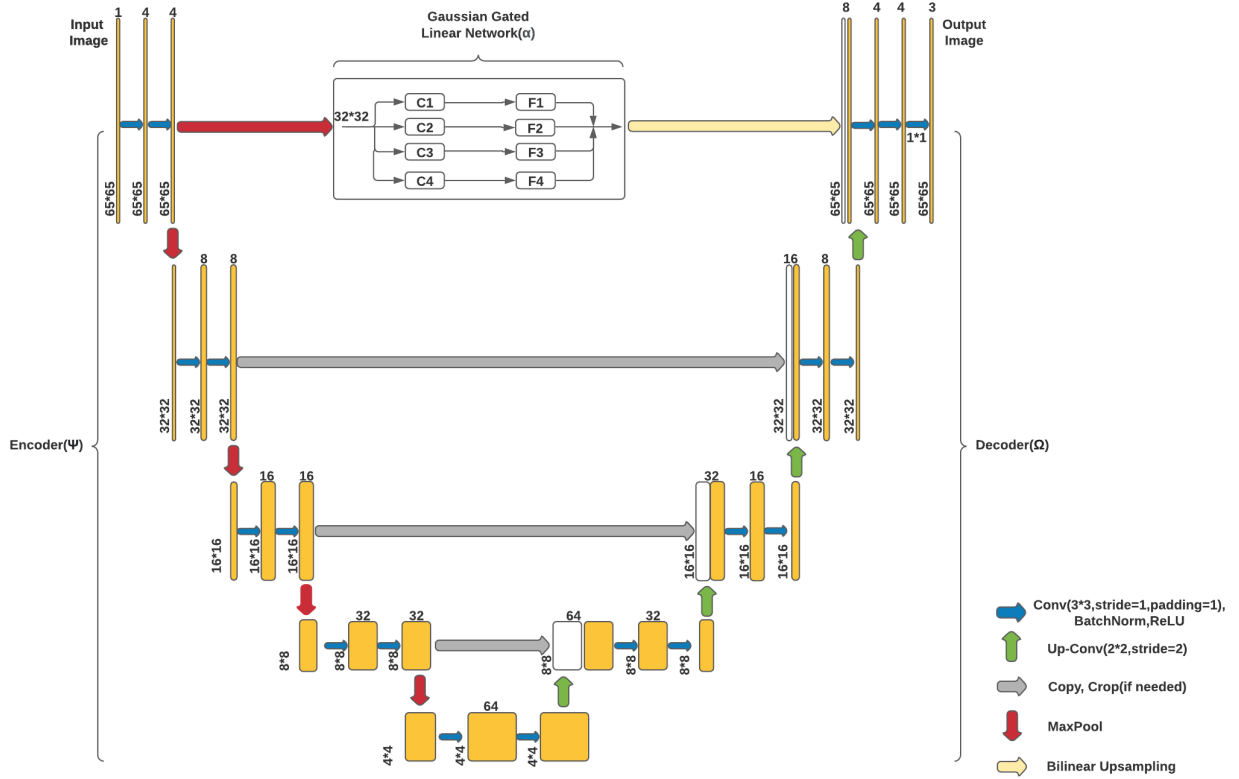}
	\caption{GLU-Net with UQ Module. The uncertainty module, $\alpha$ is inserted into the modified U-Net architecture. As described, the G-GLN Module($\alpha$) is inscribed into the architecture and it's outputs are added to the last layer of the decoder($\Omega$)}
	\label{Figure 5}
\end{figure}

We develop a novel neural network architecture by combining U-net with the Gaussian GLN - the basic premise here is to enhance the U-net and enable it with a UQ module. In specific, we add a single Gaussian GLN module at the skip connections of the U-Net architecture. We choose the G-GLN module with only a single layer, with 4 neurons, and use its output to model uncertainty. Our context map size for the G-GLN is also 4, leading to 16, weight pairs for each neuron. The mean of the output of the last layer of the G-GLN is sent across to the last layer of the decoder. The number of parameters in each module is presented in Table \ref{tab:Table 1}. The bottleneck layer is the layer that connects the last layer of the encoder and the first layer of the decoder (the bottom-most position shown in Fig. \ref{Figure 5}). We have a total of 159,351 parameters in the GLU-Net Model - this is $44\%$ less as compared to the state-of-the-art architecture DenseED-c16 \cite{zhu2018bayesian}.

For uncertainty propagation, we draw samples using the mean and variance obtained using GLN and propagate it through the decoder's last layer. Our proposed architecture not only predicts each output pixel's PDF distribution accurately but also produces a confidence interval for the PDF at every value that pixel could take. To strike a balance between the need for model parameters and the usefulness of uncertainty estimation, we introduce a max-pooling layer at the start of the skip connection and a bilinear upsample phase that  reverts the G-GLN's output to its original shape. We use two fully connected layers, one to reshape our output of the maxpool filter after flattening, the other one to reshape the output of GLN to send it to the bilinear upsampling phase. 

\begin{table}[]
    \centering
    \caption{Number of parameters in each module of GLU-Net}
    \label{tab:Table 1}
    \begin{tabular}{lll}
    \hline
    \textbf{Sl No.}     &\textbf{Layer} & \textbf{Parameters}  \\ \hline
    01 & Encoder & 18,564\\
    02 & UQ Module & 37,320 \\
    03 & Bottleneck Layer &  55,552\\
    04 & Decoder & 47,915  \\ \hline 
    \end{tabular}
\end{table}

As described in the above sections, we have made specific changes to the original U-Net architecture, and the G-GLN summarized here. For the U-Net architecture, we use Batch Normalization as in \cite{zhou2019normalization}, and decrease the number of filters in each layer as presented above. In the GLN module, we ignore the bias model, which is a constant Gaussian distribution and is concatenated to the output of every layer before forwarding it to the next. Although we implement the automatic differentiable gradient, since G-GLN does not predict the target distribution directly, we cannot use that; instead, we use the autograd module of PyTorch to achieve our goal.


\subsection{Training}
The proposed GLU-net has been implemented in PyTorch \cite{paszke2019pytorch}. We use the well-known mean-squared loss function for training. During training, the gradient of the loss function is computed using the automatic differentiation module available within PyTorch. For training, the well-known Adam optimizer is used. For the first 100 epochs, a learning rate of 0.001, and beyond that, a learning rate of 0.0005 is used. The optimizer is run for 150 epochs. For regularizing the solution, we use a weight-decay of 0.00001. A batch size of 8 is used.
For ease of readers, a pseudo-code for the training phase is shown in Algorithm \ref{alg:train}. Further details on the training phase are provided in the \ref{app:1} 


\begin{algorithm}[h]
\caption{Training GLU-Net}\label{alg:train}
Provide training batch $\mathcal D= \{ \bm X_i, \bm Y_i \}_{i=1}^{N_s}$. \\
Get $\{ \bm Z_i \}_{i=1}^{N_s}$ using GLU-net Encoder's first layer comprising double conv layers.
Pass $\{ \bm Z_i \}_{i=1}^{N_s}$ to the remaining layers in encoder-decoder and to the G-GLN module to get  $\bm Y_i, \forall i$\\
Update parameters of $\hat{\mathcal M}_{\bm \theta}$, using $Y$ and $X$ using Adam Optimizer \\
\textbf{Output: } Samples of response $\bm Y_i, i=1,\ldots,N_{sim}$. 
\end{algorithm}


\subsection{UQ using GLU-Net}\label{sec:uqusinggln}

Having discussed the basic network architecture of the proposed GLU-net, we proceed towards the primary objective -- uncertainty quantification of physical systems having high-dimensional stochastic inputs. The uncertainty in our case stems from uncertainty in the model parameters (aleatoric uncertainty). Additionally, the fact that the proposed GLU-net is trained from limited data results in epistemic uncertainty. We are interested in quantifying both types of uncertainties by using the proposed GLU-net. 

For quantifying the aleatoric uncertainty, we use the proposed GLU-net as a surrogate to the actual PDE solver. Using a generic notation, we consider $\bm \Xi$ the input parameters and $\bm Y$ the output responses. In the context of the Darcy flow problem considered in this paper, $\bm \Xi$ represents the permeability field $K(\bm x)$ and $\bm Y$ represent the pressure field $P(\bm x)$ and the velocity fields $u(\bm x), v(\bm x)$. Also consider the true mapping between the input and the output is represented by $\mathcal M$,
\[ \mathcal M: \bm \Xi \mapsto \bm Y. \]
In practice, the actual mapping $\mathcal M$ is either unknown or extremely expensive to evaluate. As a consequence, it is extremely challenging to propagate the uncertainty in $\bm Xi$ to the responses $\bm Y$. 
Considering $\mathcal D = \{\bm \Xi_i, \bm Y_i \}_{i=1}^{N_s}$ to be training data, we train the proposed GLU net model to learn an approximate mapping $\hat{\mathcal M}_{\bm \theta}: \bm \Xi \mapsto \bm Y$ parameterized by the network parameters $\bm \theta$. The mapping $\hat{\mathcal M}_{\bm \theta}$ acts as a surrogate to the true mapping $\mathcal M$. We use $\hat{\mathcal M}_{\bm \theta}$ for propagating the uncertainty from the input to the output. The details on the same can be found in Algorithm \ref{alg:aleatoric_propagation}.
\begin{algorithm}[h]
\caption{Propagation of aleatoric uncertainty}\label{alg:aleatoric_propagation}
\textbf{Initialize:} Provide training data $\mathcal D= \{\bm \Xi_i, \bm Y_i \}_{i=1}^{N_s}$. \\
Learn $\hat{\mathcal M}_{\bm \theta}: \bm \Xi \mapsto \bm Y$ using GLU-net \Comment*[r]{Algorithm \ref{alg:train}}
Generate samples from the stochastic input parameters
\[\bm \xi_i \sim p\left( \bm \Xi \right),\;\; \text{for }i=1,2,\ldots,N_{sim} \]
where $ P\left( \bm \Xi \right)$ represents probability density function of the stochastic input, $\bm \Xi$. \\
Compute $\bm y_i, \forall i$ by using the trained surrogate model $\hat{\mathcal M}_{\bm \theta}$. \\
\textbf{Output: } Samples of response $\bm Y_i, i=1,\ldots,N_{sim}$. 
\end{algorithm}

As for quantifying epistemic uncertainty, the proposed GLU-net has a UQ module embedded within it. To be specific, we draw samples from the output distribution of the G-GLN, pass it to the decoder(append it to the decoder's last layer), and get sample outputs for a single input. Overall, the proposed GLU-net provides a comprehensive framework to quantify both aleatoric and epistemic uncertainty in a stochastic system.

\section{Implementation and Results}\label{sec:iandr}
We illustrate the performance of the proposed approach in solving the Darcy flow problem defined in Eq. \eqref{eq:eq3}. The required training samples are generated by using the open-source finite element solver \texttt{FeNICS} \cite{alnaes2015fenics}. The problem domain is subdivided in $65 \times 65$ elements. 
Accordingly, $\bm \Xi \in \mathbb R^{65\times 65}$ and $\mathbf Y \in \mathbb R^{3\times 65 \times 65}$. In order to control the intrinsic dimensionality, we have used the Karhunen Loève expansion \cite{huang2001convergence},
\begin{equation}\label{eq:kle}
    \log K(\bm s) = \mu_K + \sum_{i=1}^q{\sqrt{\lambda_i}z_i\phi_i(\bm s)},
\end{equation}
where $\mu_K$ is the mean permeability, $\lambda_i$ is the $i-$th eigenvalue, and $\phi_i(\bm s)$ is the value of the eigenfunction at $\bm s$. The eigenvalue and the eigenfunctions are obtained by solving an eigenvalue problem with the covariance function of the log-permeability field. $z_i$ represents the stochastic variable. Finally, $q$ represents the intrinsic stochastic dimensionality. For illustrating scalability of the proposed approach, we have considered $q=50, 500$, and $4225$. Hereafter, the three cases are referred to as KLE50, KLE500, and KLE4225. 

\subsection{Metrics and Evaluation}
To evaluate the accuracy of the proposed approach, we use two error metrics, namely (a) mean-squared error and (b) coefficient of determination ($R^2$ score). Mean-squared error is a global measure is an absolute error metric and computed as
\begin{equation}\label{eq:eq15}
        L(y,{y'}) = \frac{1}{{3.{n_s}}}\left( {\sum\limits_{i = 1}^3 {\sum\limits_{j = {n_1}}^{{n_s}} {{{\left( {{y_{i,j}} - y'_{i,j}} \right)}^2}} } } \right),
\end{equation} 
where ${n_s}$ is the length of the set $S$ and $i$ being each of the output field for a single simulation, $y$ being the predicted output fields for a single simulation, and $y'$ being the true output fields. $R^2-$score on the other hand is a relative error measure and computed as
\begin{equation}\label{eq:eq16}
{R^2} = 1 - \frac{{\sum\limits_{k = 1}^N {\sum\limits_{i = 1}^3 {\sum\limits_{j = {n_1}}^{{n_s}} {{{\left( {y_{i,j}^k - y_{i,j}^{',k}} \right)}^2}} } } }}{{\sum\limits_{k = 1}^N {\sum\limits_{i = 1}^3 {\sum\limits_{j = {n_1}}^{{n_s}} {{{\left( {y_{i,j}^k - \mathop {{y_{i,j}}}\limits^ -  } \right)}^2}} } } }}    
\end{equation}
where $\{ {x^k},{y^k}\} _{k = 1}^N$, is the data set(training data set or the testing), N is the number of testing simulations, ${y'}$ being the output fields and $\mathop y\limits^ -$ being true output mean to evaluate our model. ${R^2}$ score gives us an estimate of how better is our prediction compared to predicting the mean at that spatial point. An ${R^2}$ score of 0, tells us that our estimate is no better than predicting mean of the sample, and an ${R^2}$ score of 1, tells us our estimate is a perfect prediction. An ${R^2}$ score of negative is a bad prediction. We also carry out visual inspection based qualitative assessment.

\subsection{Model validation}
\begin{figure}[htbp!]%
    \centering
    \subfloat[\centering Pressure]{{\includegraphics[width=0.3\textwidth]{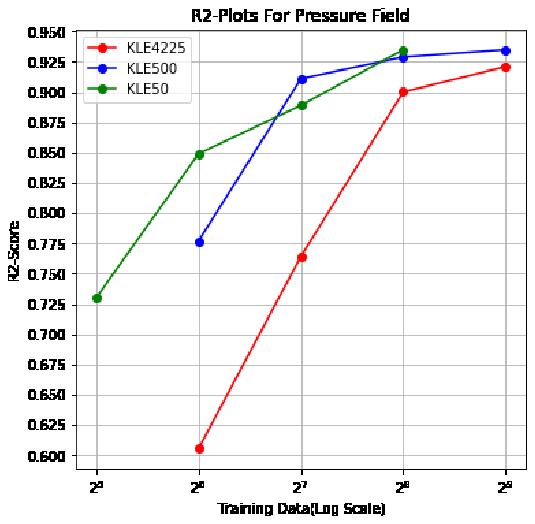}}}%
    \hspace{0.2cm}
    \subfloat[\centering Velocity(X)]{{\includegraphics[width=0.3\textwidth]{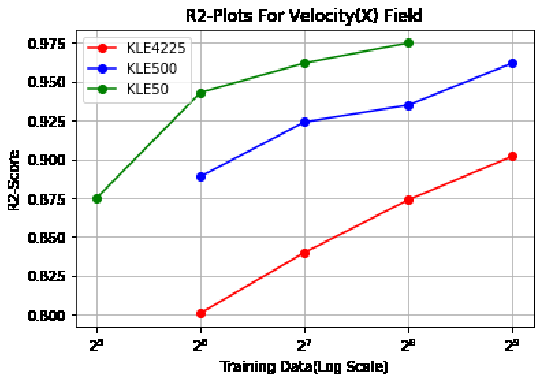}}}%
    \hspace{0.2cm}
    \subfloat[\centering Velocity(Y)]{{\includegraphics[width=0.3\textwidth]{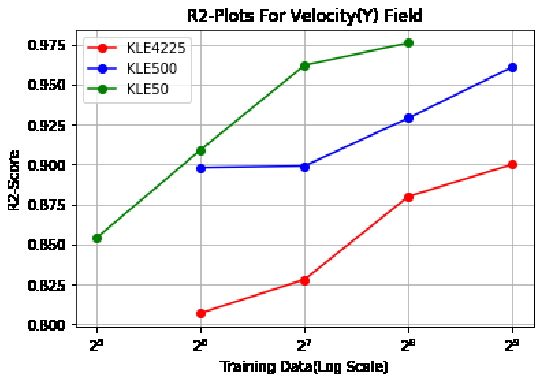}}}%
    \caption{In the above three figures, we display the ${R^2}$ scores on the unseen testing dataset, for models whose length of training data is shown in the X-Axis. We can easily observe that, the scores for the more complex KLE4225, the scores are usually lesser than the corresponding KLE500 and KLE50 trained on the same number of training samples. We also want to put forward that, the model performs well on the Velocity fields and struggles with Pressure field. It is evident that we require no more than 512 training samples, to attain a good result(${R^2}$ score of 0.9)}%
    \label{fig:r2-score}%
\end{figure}

\begin{figure}[ht!]
	\centering
	\includegraphics[width = 0.5\textwidth]{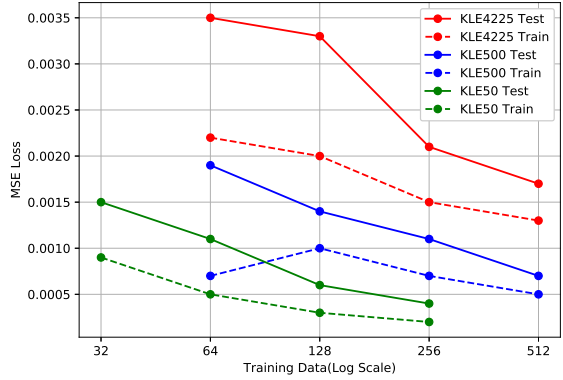}
	\caption{GLU-Net's Mean Squared Error Score for each of the model combinations, on both training set and the test set. The test set has a total of 500 unseen test set images. We ensure there is no overfitting on the models trained with lesser data.}
	\label{unet_gln}
\end{figure}

Before proceeding with the uncertainty quantification results, we validate the performance of the proposed GLU-net based on test data. To that end, we use 500 test data. Considering the fact that the proposed approach predicts $p(\bm y|\bm \Xi = \bm \xi)$, we compare the predictive mean $\mathbb E(\bm y|\bm \Xi = \bm \xi)$ with the true solution. For KLE50, we considered a training sample size of 32, 64, 128, and 256. For KLE500 and KLE4225, a training sample size of 64, 128, 256, and 512 were considered. Fig. \ref{fig:r2-score} shows the $R^2-$score for the pressure and velocity responses. As expected, the performance of the proposed approach is inversely proportional to the model complexity (as indicated by the intrinsic dimensionality). We note that the proposed GLU-net approximates the velocity fields better as compared to the pressure field. This is counter intuitive as the pressure field is simpler as compared to the velocity fields. One possible remedy to this is to use separate networks for the pressure and velocity fields. Nonetheless, for all the cases, we obtain $R^2-$score of above 0.90, indicating excellent performance of the proposed GLU-net. To understand the convergence of the proposed approach, training and test error corresponding to different training sample sizes are shown in Fig. \ref{unet_gln}. Mostly, as the number of training samples increases, the test and training error reduces. This indicates that the proposed GLU-net is trained properly with minimal to no overfitting.

\begin{figure}[htbp!]%
    \centering
    \subfloat[\centering True Pressure]{{\includegraphics[width=0.3\textwidth]{50_64_p_in.eps}}}%
    \hspace{0.2cm}
    \subfloat[\centering True Velocity(X)]{{\includegraphics[width=0.3\textwidth]{50_64_ux_in.eps}}}%
    \hspace{0.2cm}
    \subfloat[\centering True Velocity(Y)]{{\includegraphics[width=0.3\textwidth]{50_64_uy_in.eps}}}%
    \hspace{0.2cm}
    \subfloat[\centering Predicted Pressure (64 training points)]{{\includegraphics[width=0.3\textwidth]{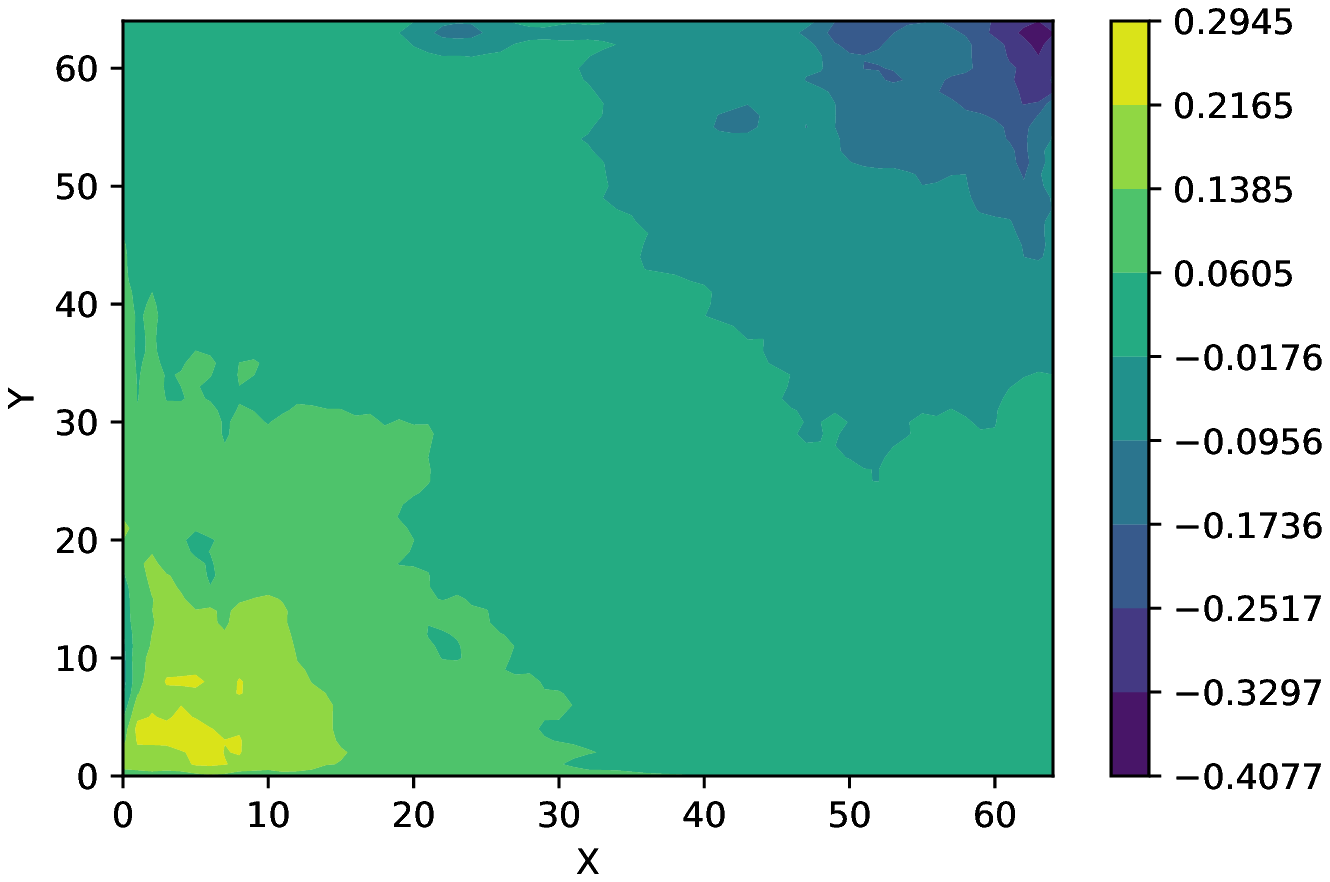}}}%
    \hspace{0.2cm}
    \subfloat[\centering X-component of Velocity (64 training samples)]{{\includegraphics[width=0.3\textwidth]{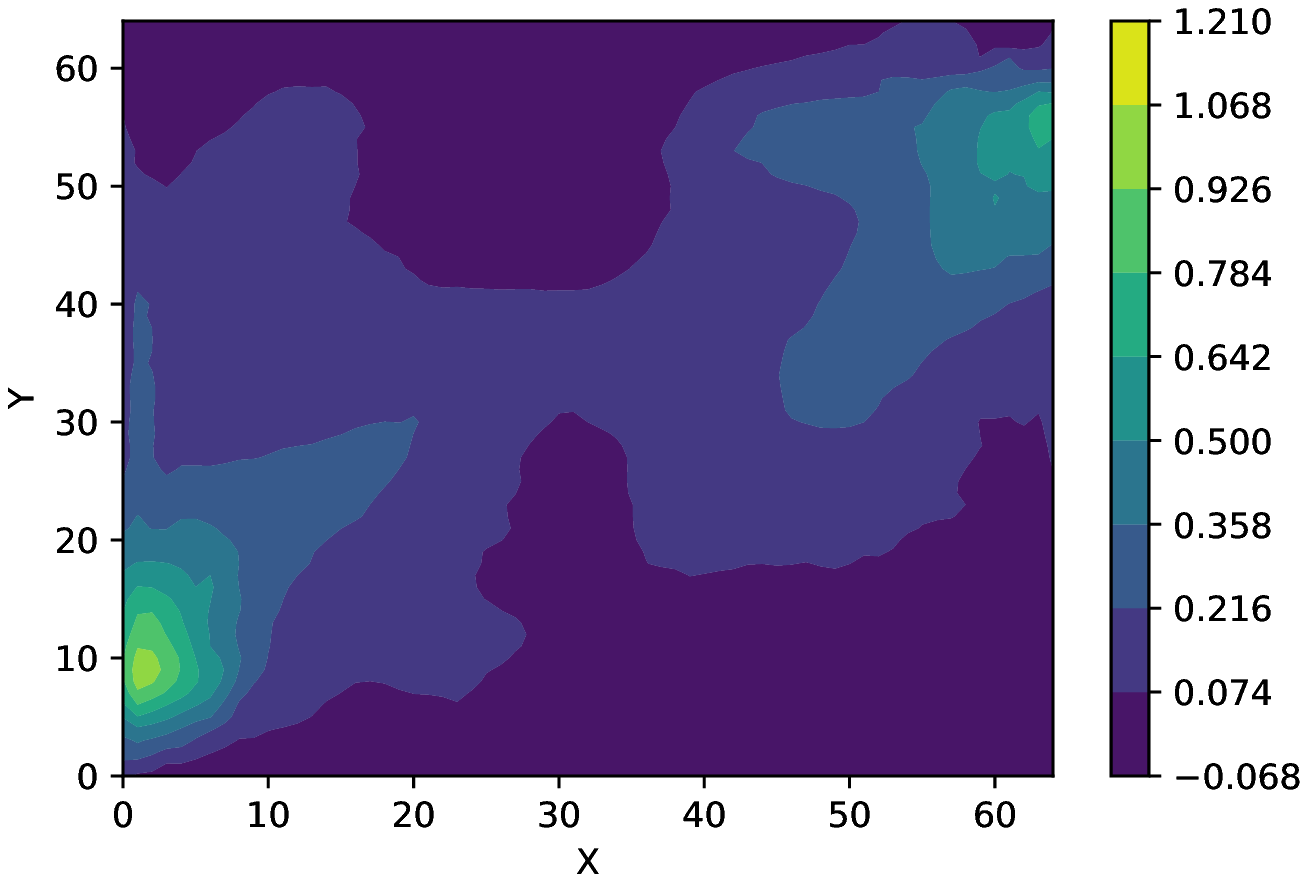}}}%
    \hspace{0.2cm}
    \subfloat[\centering Y-component of Velocity (64 training samples)]{{\includegraphics[width=0.3\textwidth]{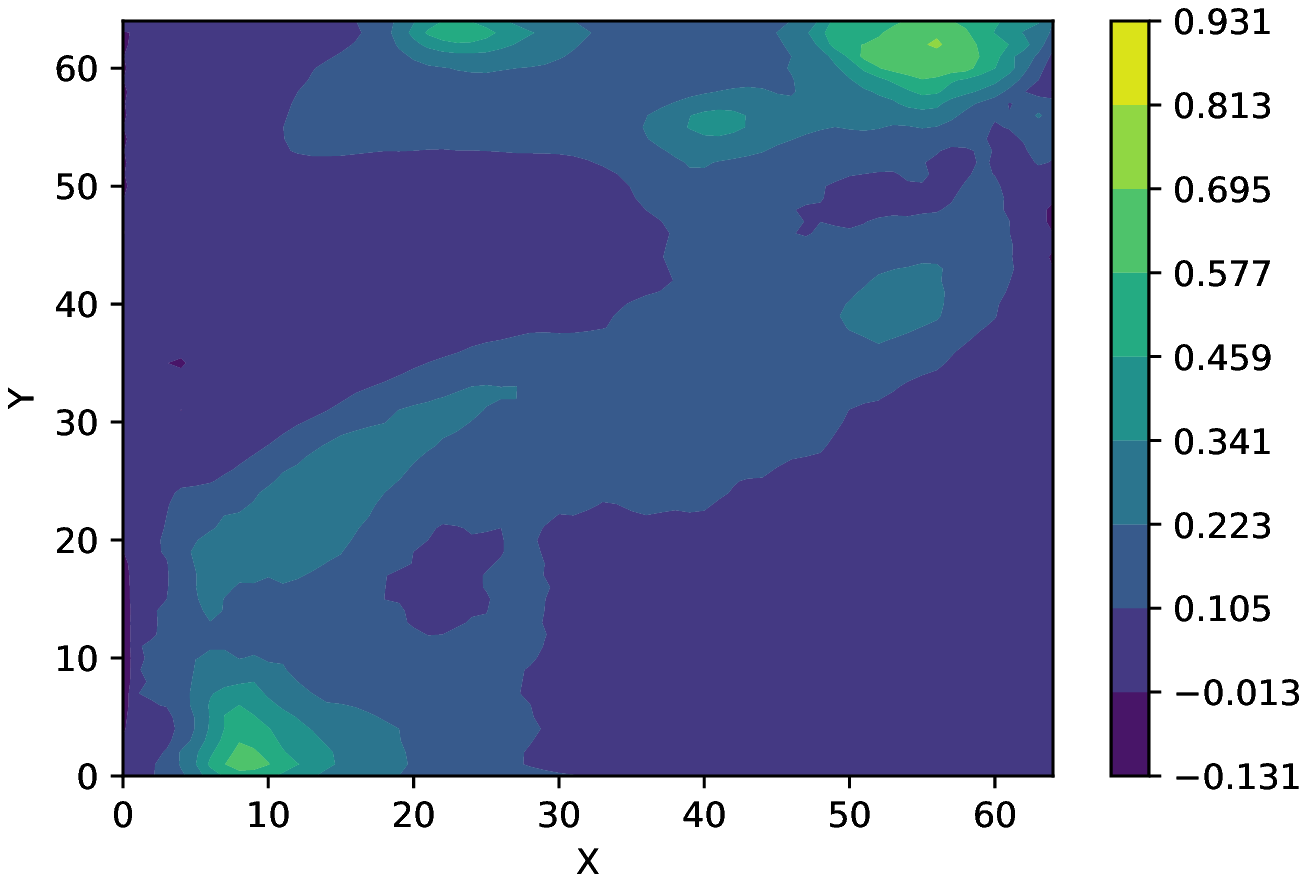}}}%
    \hspace{0.2cm}
    \subfloat[\centering Predicted Pressure (256 training points)]{{\includegraphics[width=0.3\textwidth]{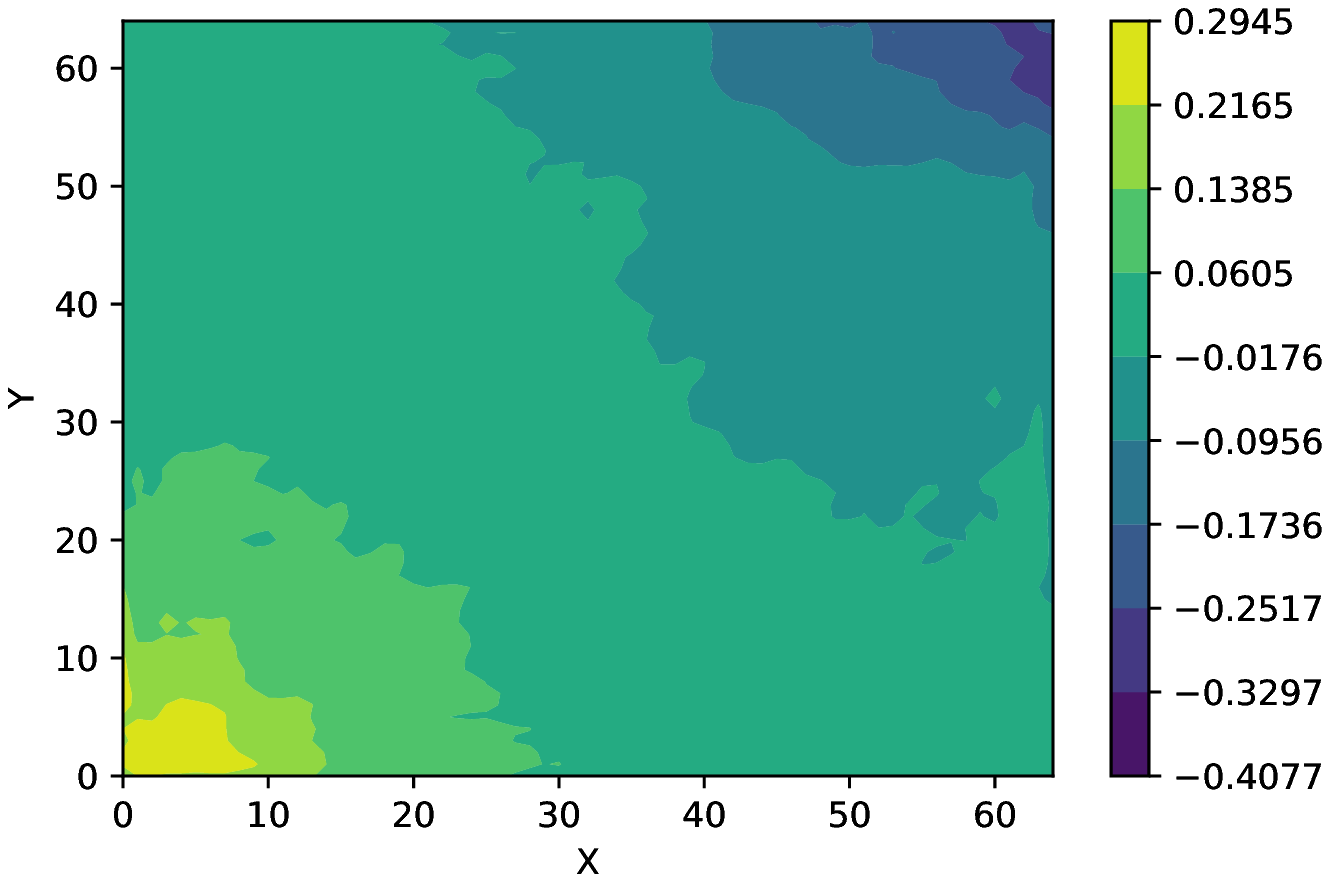}}}%
    \hspace{0.2cm}
    \subfloat[\centering X-component of Velocity (256 training samples)]{{\includegraphics[width=0.3\textwidth]{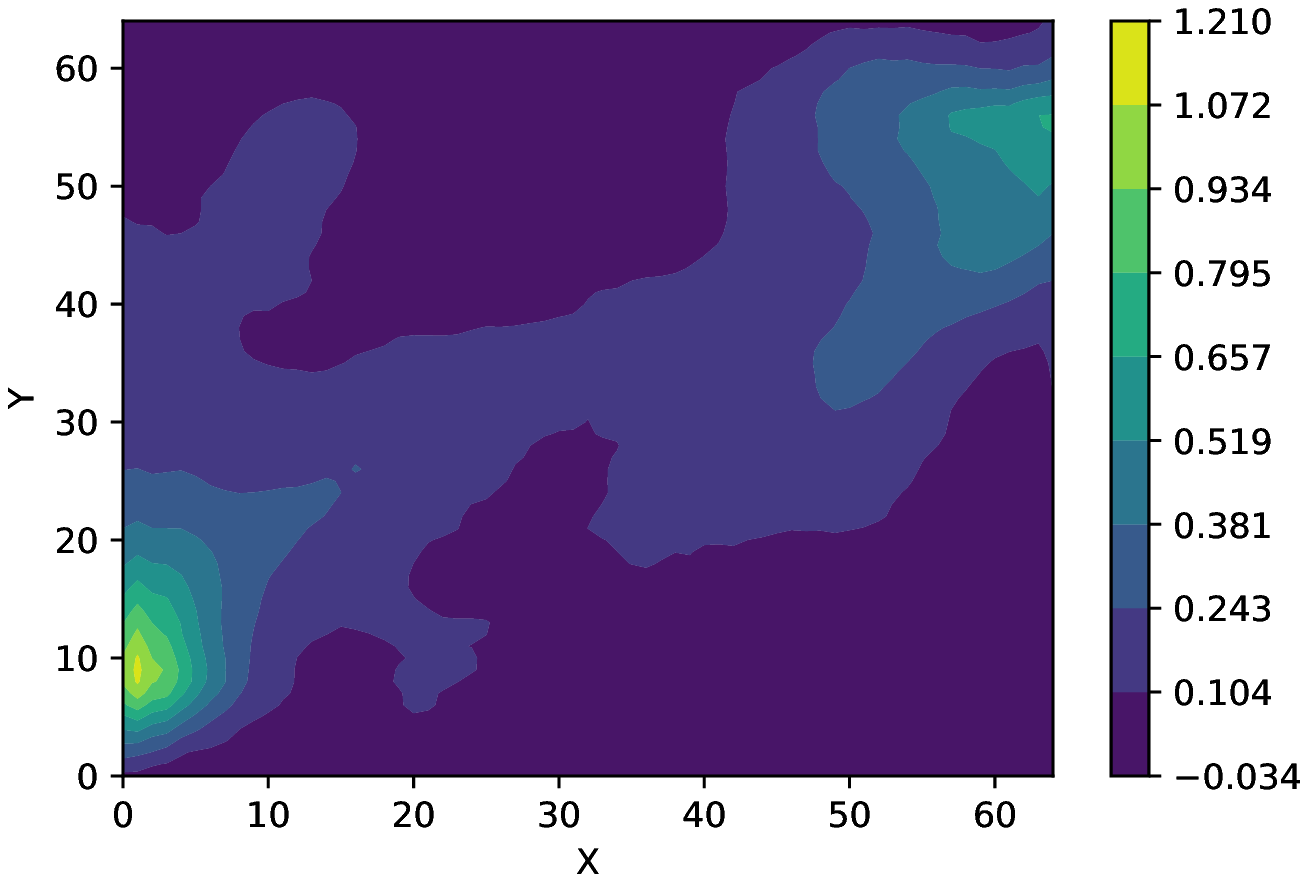}}}%
    \hspace{0.2cm}
    \subfloat[\centering Y-component of Velocity (256 training samples)]{{\includegraphics[width=0.3\textwidth]{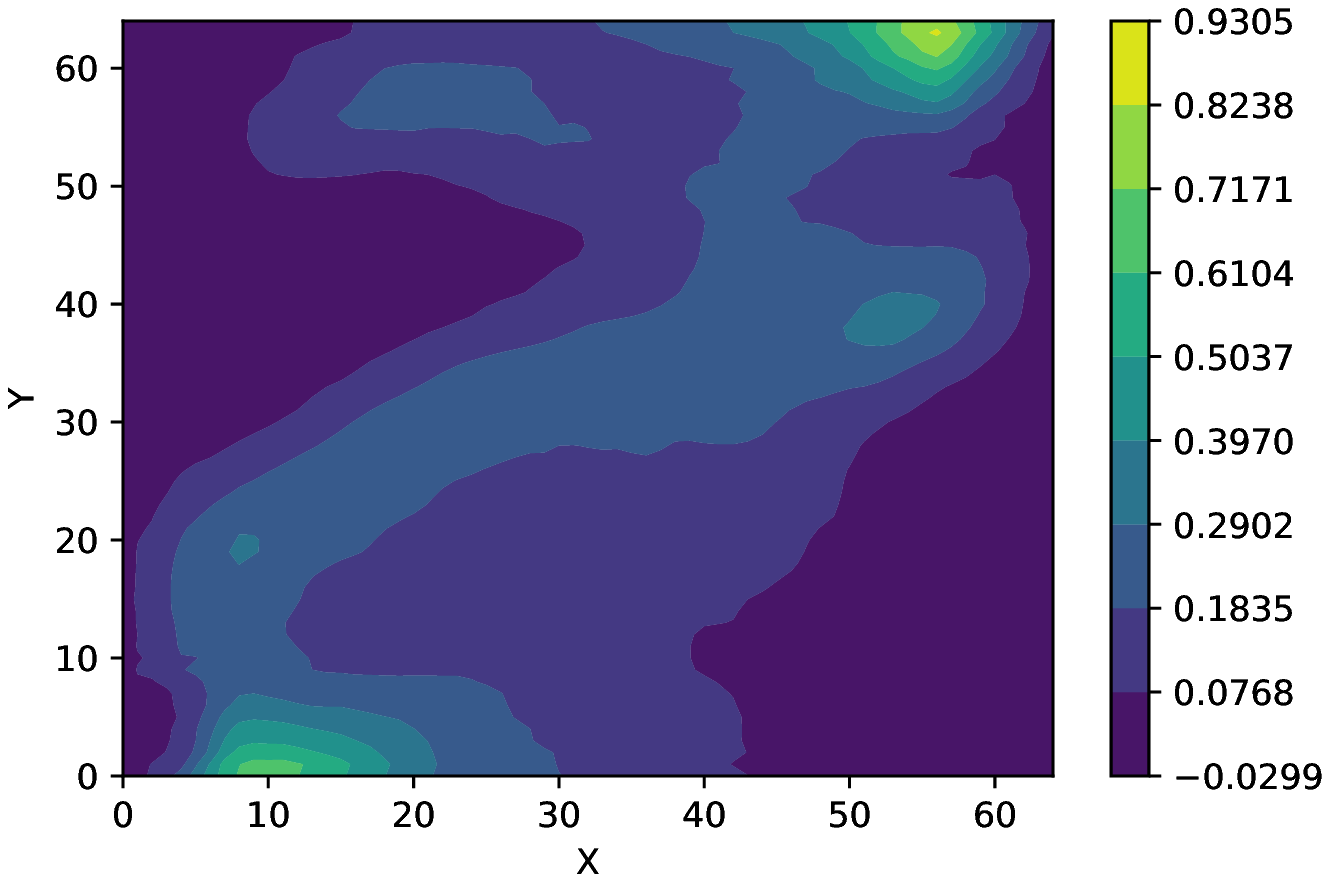}}}%
    \caption{We show our output prediction fields for the KLE50 input fields presented in figure Fig. \ref{fig:fig1}. The first row corresponds to the ground truth, the second and third rows being the output of the KLE50 GLU-Net trained with 64 and 256 training samples respectively.}%
    \label{fig:kle50}%
\end{figure}

\begin{figure}[htbp!]%
    \centering
    \subfloat[\centering True Pressure]{{\includegraphics[width=0.3\textwidth]{500_128_p_in.eps}}}%
    \hspace{0.2cm}
    \subfloat[\centering True Velocity(X)]{{\includegraphics[width=0.3\textwidth]{500_128_ux_in.eps}}}%
    \hspace{0.2cm}
    \subfloat[\centering True Velocity(Y)]{{\includegraphics[width=0.3\textwidth]{500_128_uy_in.eps}}}%
    \hspace{0.2cm}
    \subfloat[\centering Predicted Pressure (128 training points)]{{\includegraphics[width=0.3\textwidth]{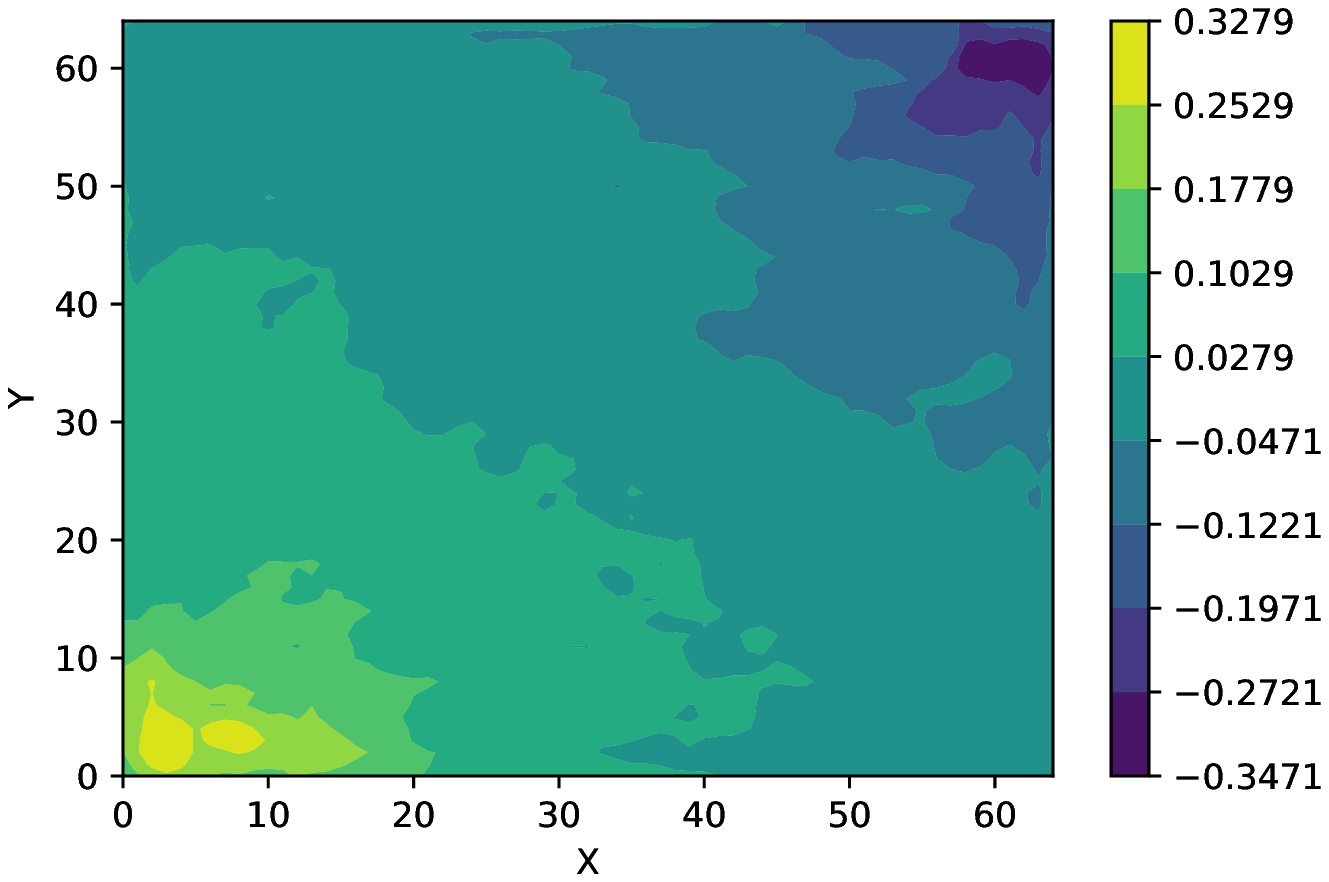}}}%
    \hspace{0.2cm}
    \subfloat[\centering X-component of Velocty (128 training samples)]{{\includegraphics[width=0.3\textwidth]{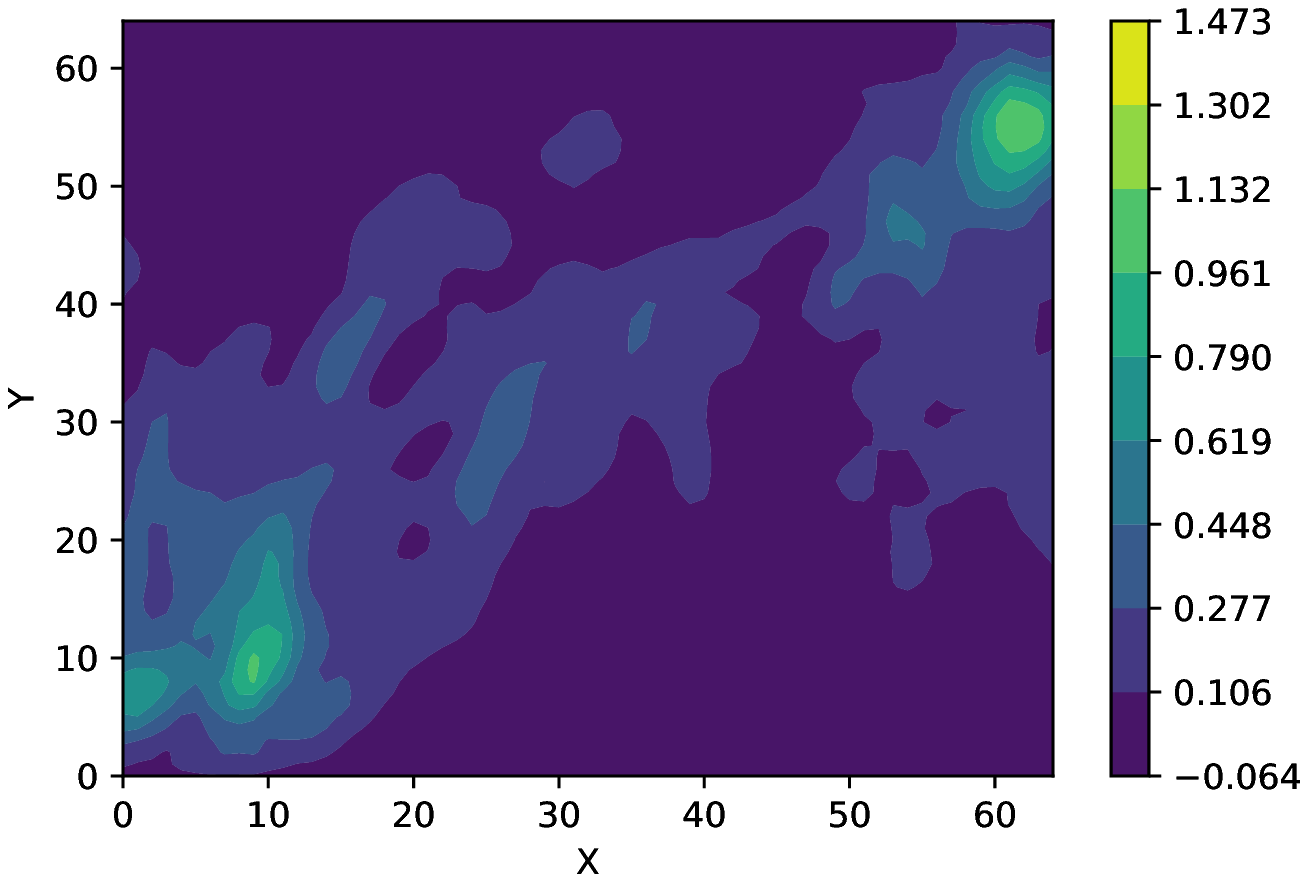}}}%
    \hspace{0.2cm}
    \subfloat[\centering Y-component of Velocty (128 training samples)]{{\includegraphics[width=0.3\textwidth]{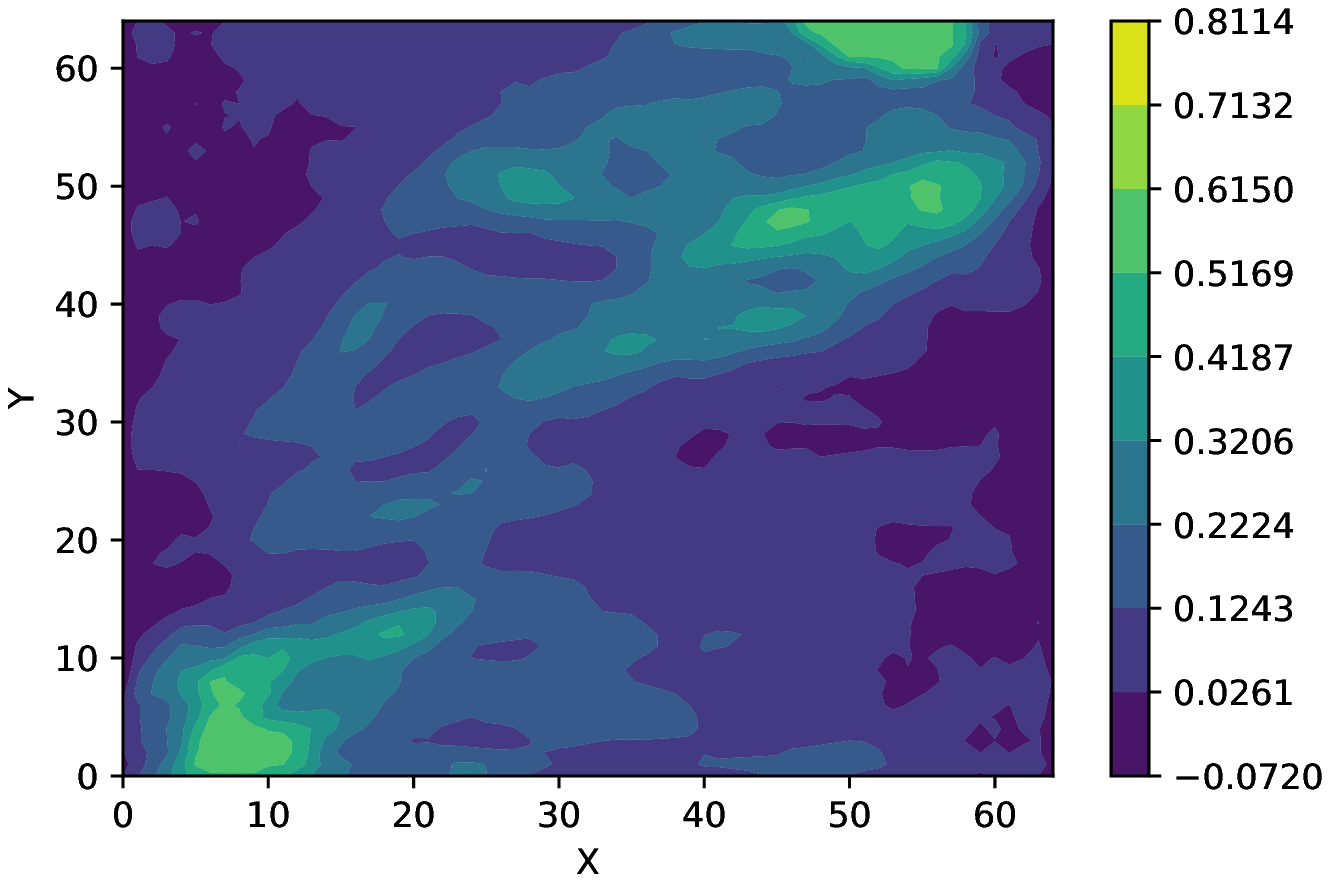}}}%
    \hspace{0.2cm}
    \subfloat[\centering Predicted Pressure (512 training points)]{{\includegraphics[width=0.3\textwidth]{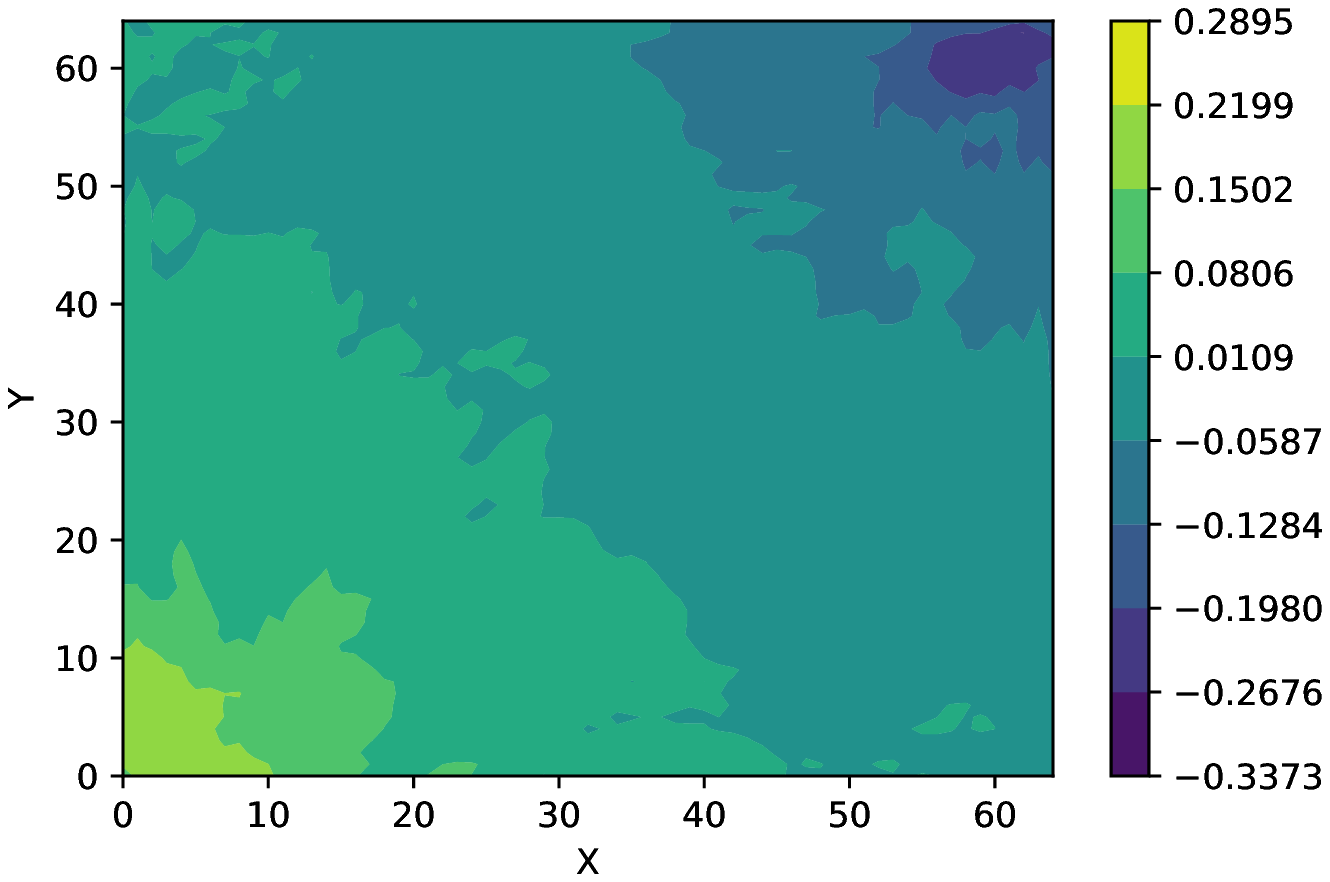}}}%
    \hspace{0.2cm}
    \subfloat[\centering X-component of Velocty (512 training samples)]{{\includegraphics[width=0.3\textwidth]{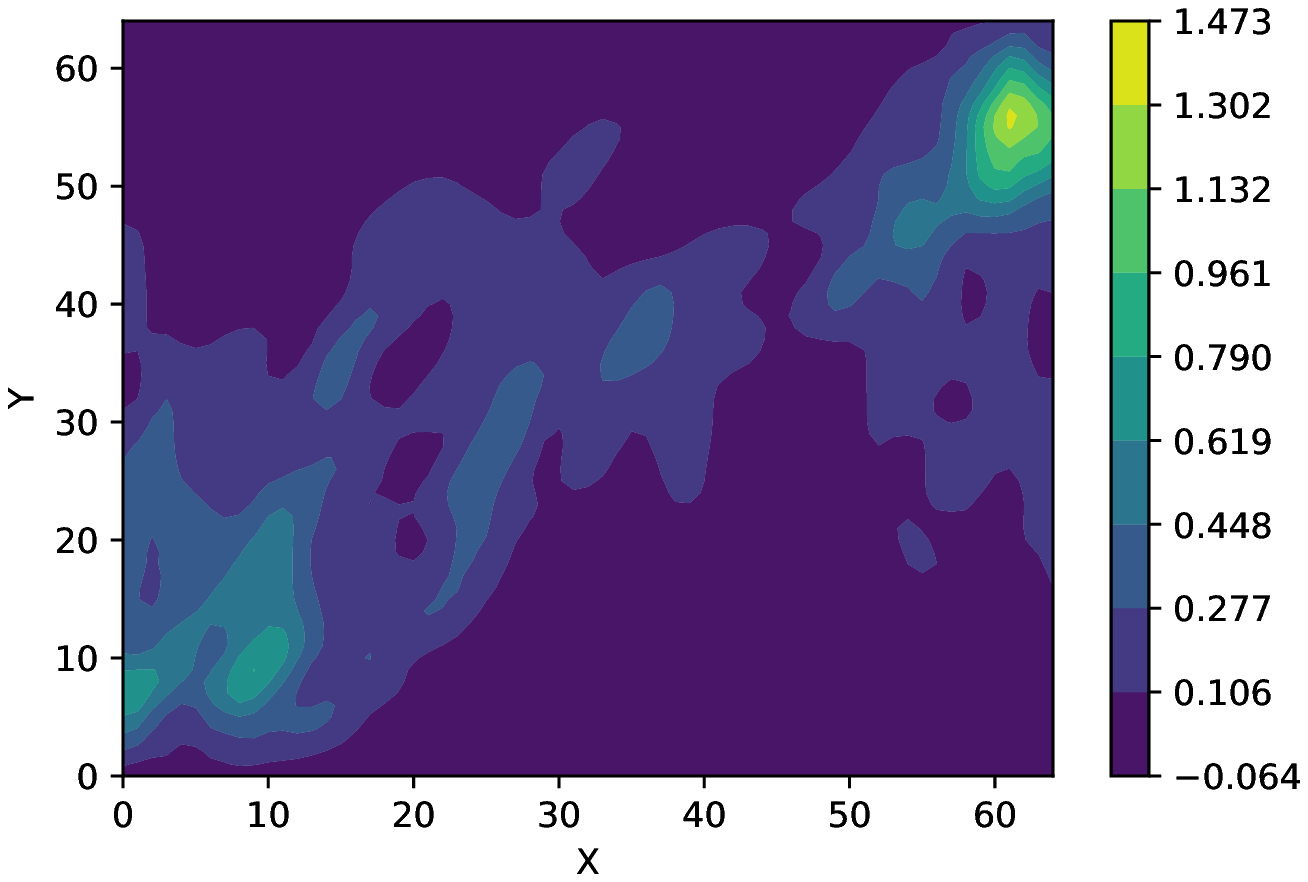}}}%
    \hspace{0.2cm}
    \subfloat[\centering Y-component of Velocty (512 training samples)]{{\includegraphics[width=0.3\textwidth]{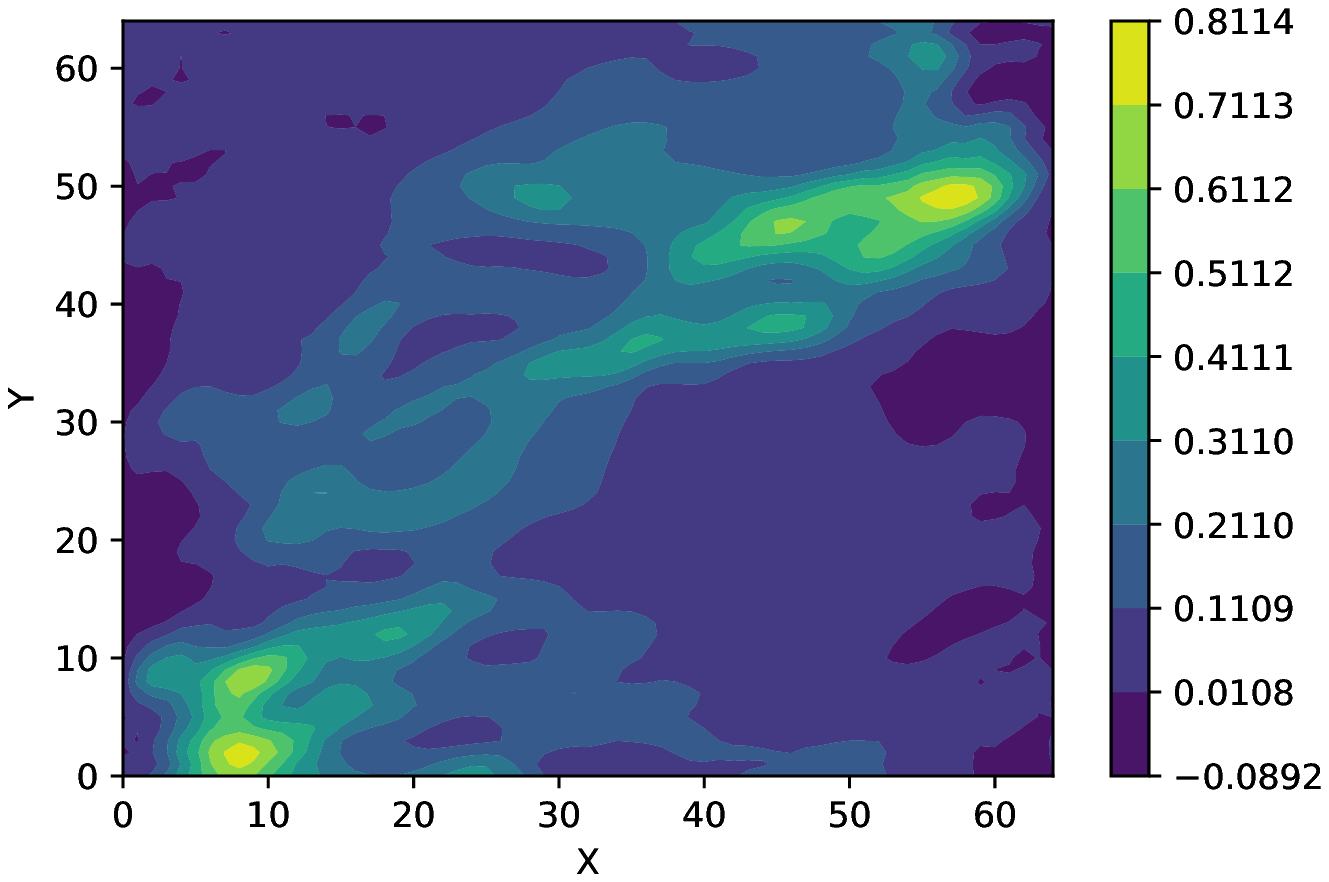}}}%
    \caption{We show our output prediction fields for the KLE500 input fields presented in Fig. \ref{fig:fig1}. The first row corresponds to the ground truth, the second and third row being the output of the KLE500 GLU-Net trained with 128 and 512 training samples respectively. Here we use more training samples than KLE50 as our bets model here, in KLE500.}%
    \label{fig:kle500}%
\end{figure}

\begin{figure}[htbp!]%
    \centering
    \subfloat[\centering True Pressure]{{\includegraphics[width=0.3\textwidth]{4225_128_p_in.eps}}}%
    \hspace{0.2cm}
    \subfloat[\centering True Velocity(X)]{{\includegraphics[width=0.3\textwidth]{4225_128_ux_in.eps}}}%
    \hspace{0.2cm}
    \subfloat[\centering True Velocity(Y)]{{\includegraphics[width=0.3\textwidth]{4225_128_uy_in.eps}}}%
    \hspace{0.2cm}
    \subfloat[\centering Predicted Pressure (128 training points)]{{\includegraphics[width=0.3\textwidth]{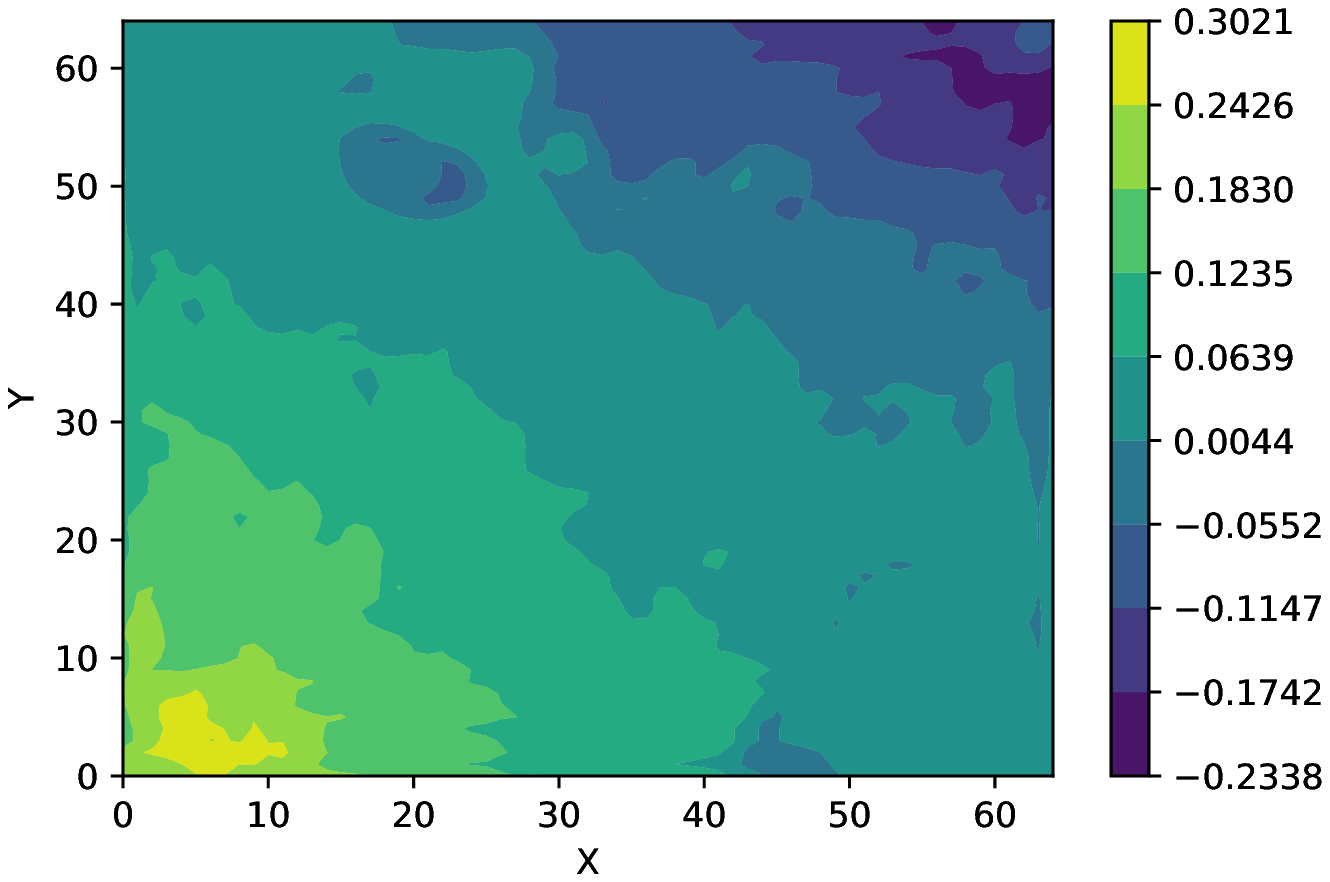}}}%
    \hspace{0.2cm}
    \subfloat[\centering X-component of Velocty (128 training samples)]{{\includegraphics[width=0.3\textwidth]{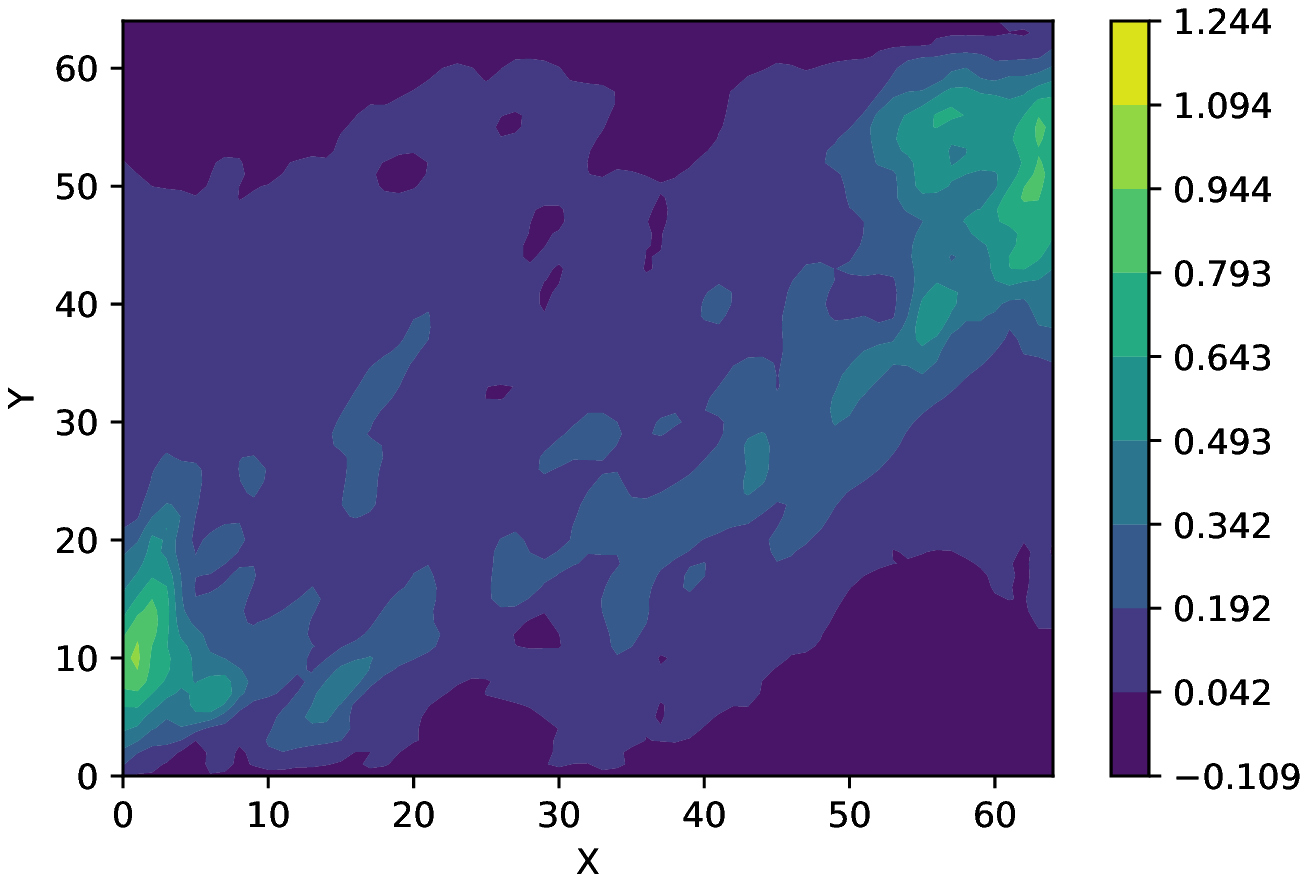}}}%
    \hspace{0.2cm}
    \subfloat[\centering Y-component of Velocty (128 training samples)]{{\includegraphics[width=0.3\textwidth]{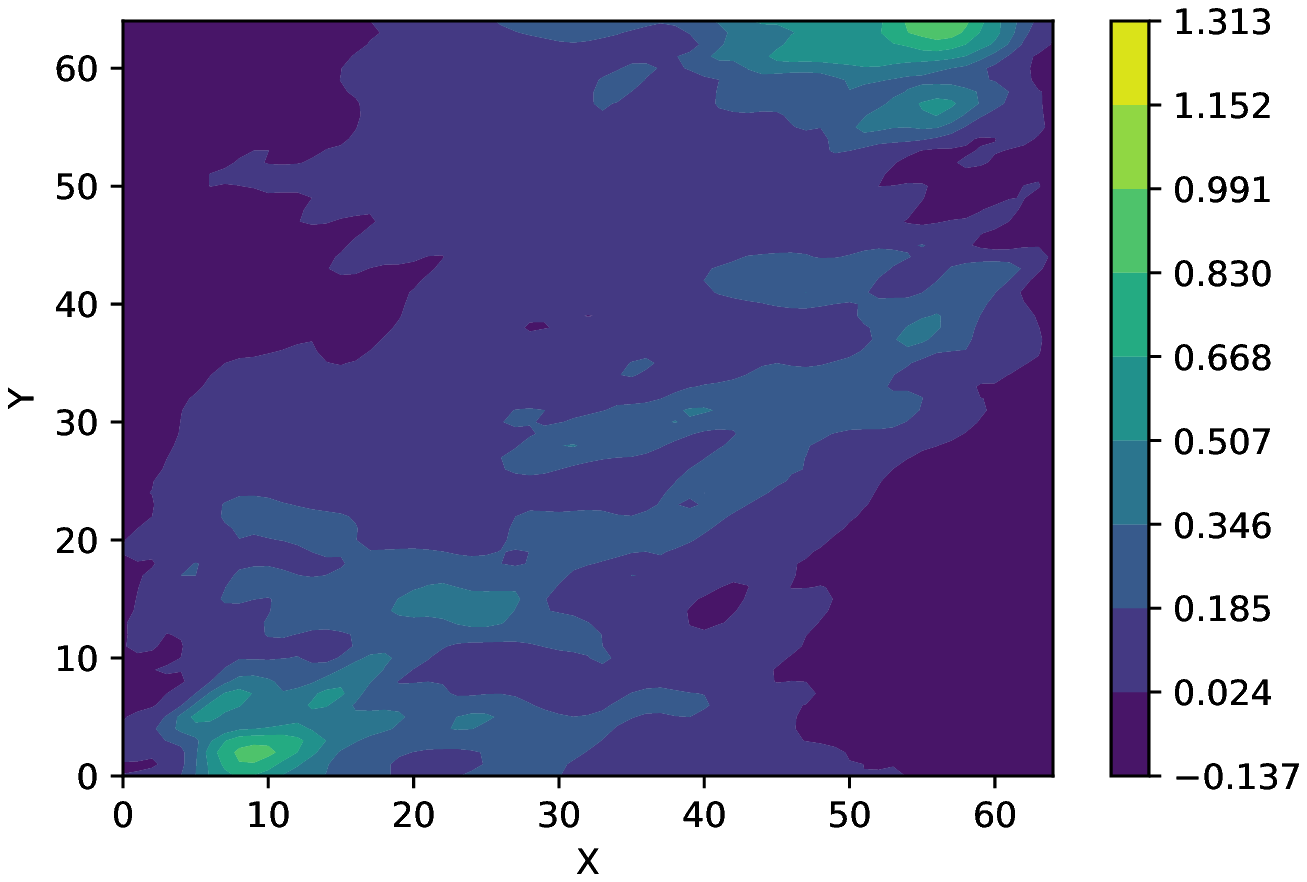}}}%
    \hspace{0.2cm}
    \subfloat[\centering Predicted Pressure (512 training points)]{{\includegraphics[width=0.3\textwidth]{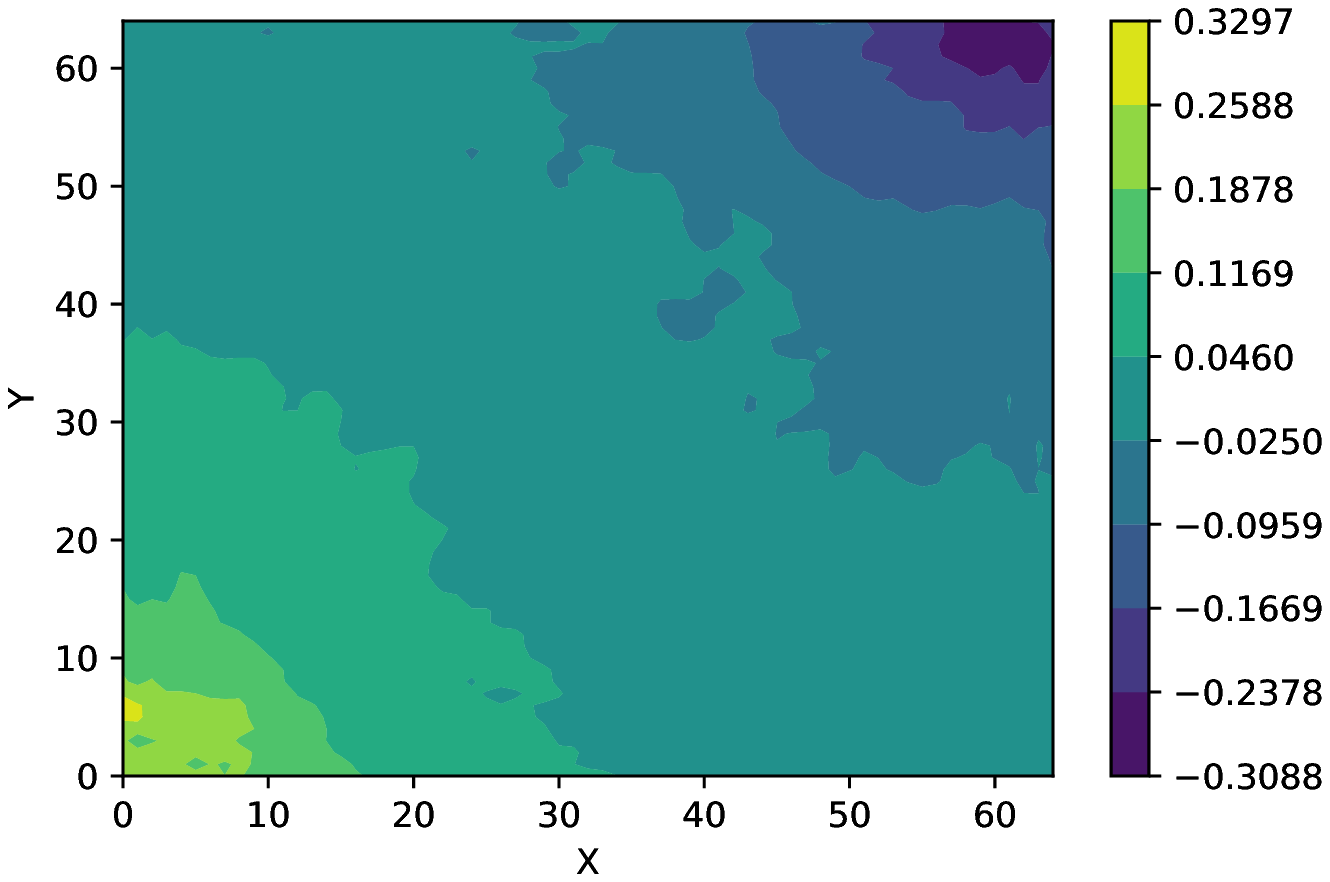}}}%
    \hspace{0.2cm}
    \subfloat[\centering X-component of Velocty (512 training samples)]{{\includegraphics[width=0.3\textwidth]{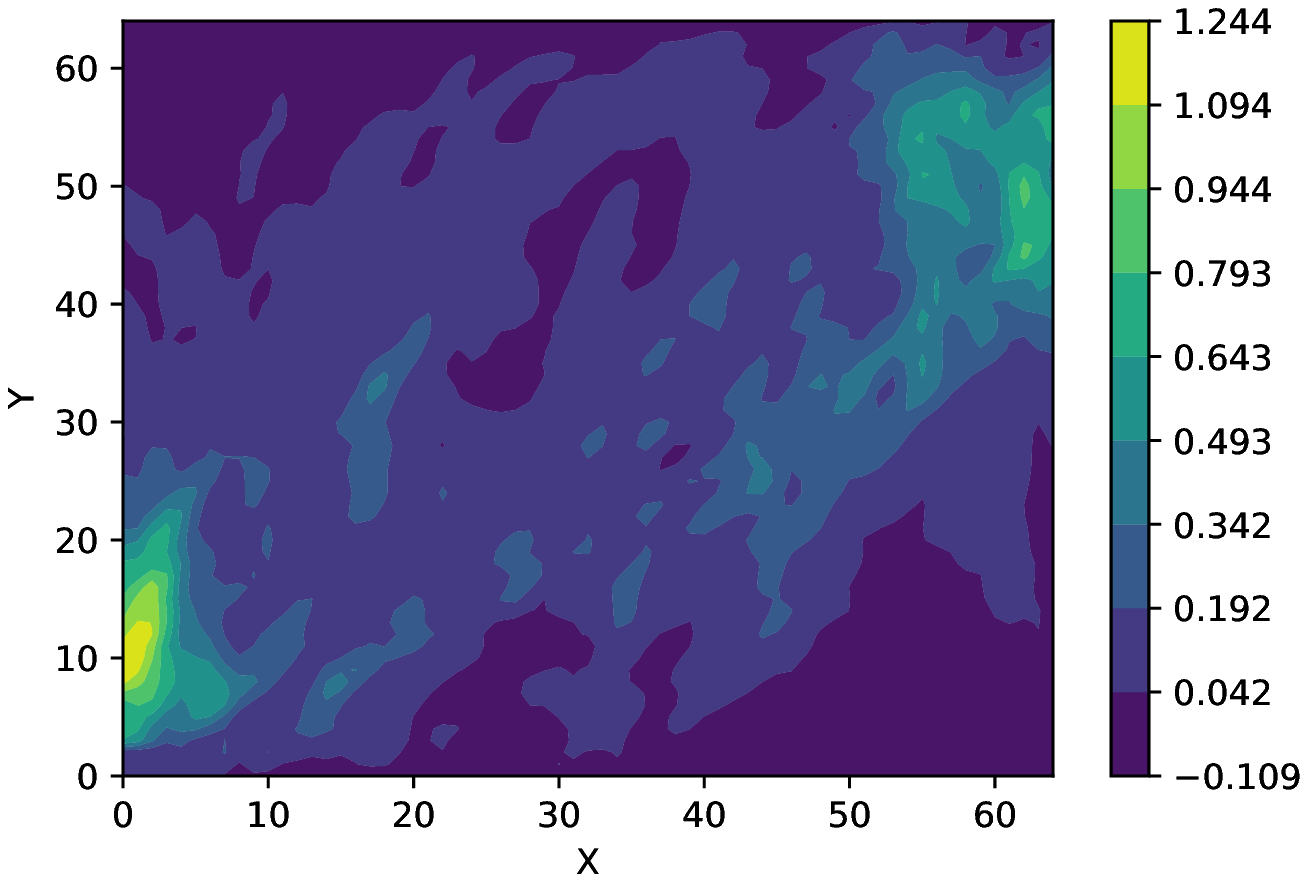}}}%
    \hspace{0.2cm}
    \subfloat[\centering Y-component of Velocty (512 training samples)]{{\includegraphics[width=0.3\textwidth]{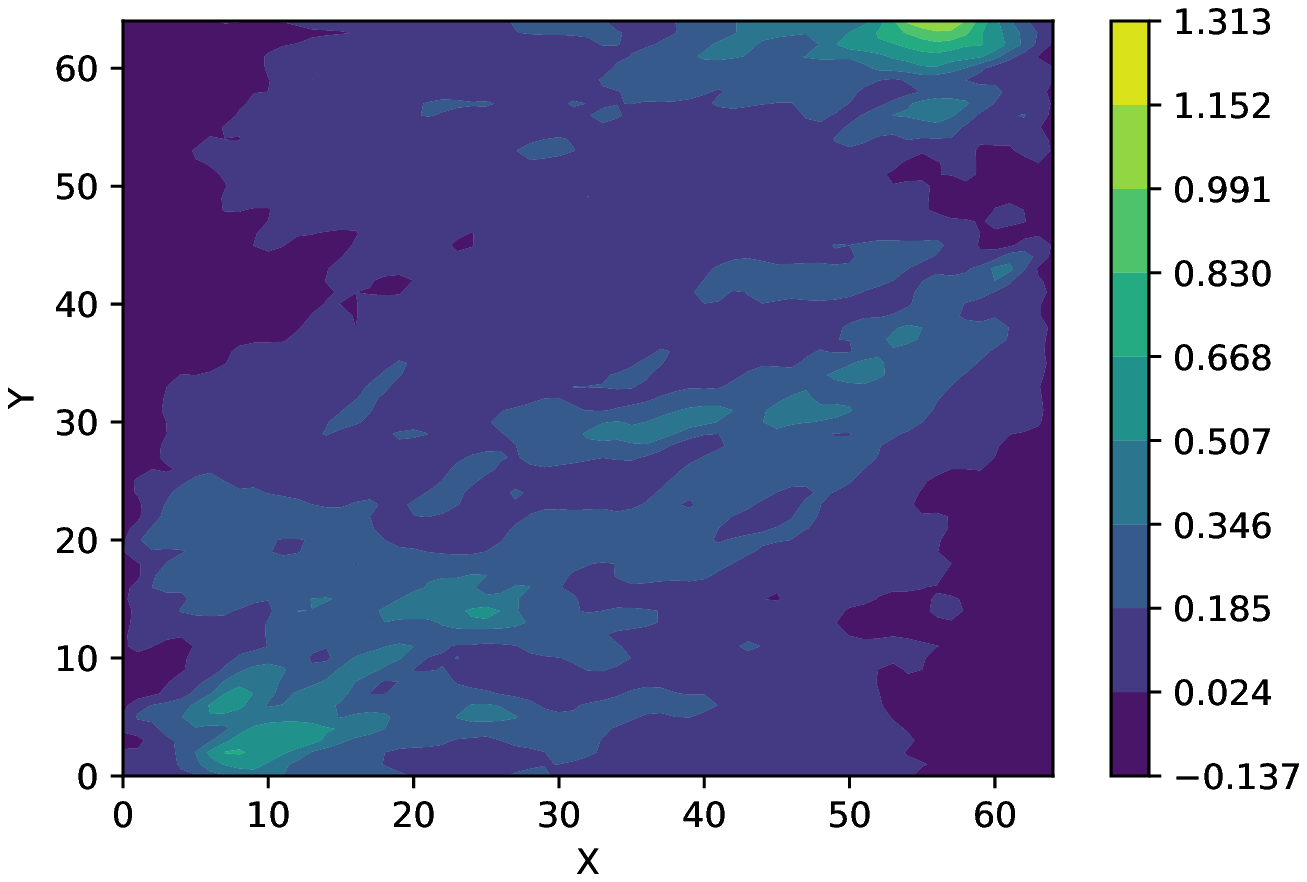}}}%
    \caption{We show our output prediction fields for the KLE4225 input fields presented in Fig. \ref{fig:fig1}. The second and third row being the output of the KLE500 GLU-Net trained with 128 and 512 training samples respectively. Here we use the same number of training samples than KLE500.}%
    \label{fig:kle4225}%
\end{figure}

Fig. \ref{fig:kle50} shows the predicted (mean) pressure and velocity field corresponding to the KLE50. The first row corresponds to the ground truth obtained using \texttt{FeNICS} while the second and the third rows correspond to results obtained using GLU-net trained with 64 and 256 training samples. We observe that the proposed approach with 256 training samples is able to reasonably capture the pressure and velocity fields. Figs. \ref{fig:kle500} and \ref{fig:kle4225} shows similar results for KLE500 and KLE4225. We observe that as the intrinsic dimensionality of the permeability increases, smaller features are observed. Although GLU-net for both KLE500 and KLE4225 yields reasonable estimates of the pressure and velocity fields, smaller flow features are better captured with 512 training samples.

\subsection{Results and discussion}
Having validated the proposed GLU-net, we proceed towards the paper's main objective - quantifying uncertainty in the response variables. We are primarily interested in quantifying the first two moments of the response fields and the probability density function (PDF). To that end, we carry out a Monte Carlo simulation (MCS) using the trained GLU-net model. Note that the proposed approach being probabilistic in nature yields predictive distribution $p\left(\bm y|\bm \Xi = \bm \xi^* \right)$ at each MCS points $\bm \xi^*$. For obtaining the contour of the mean and variance of the pressure and velocities, we have used $\mathbb E \left[\bm y | \bm \xi^* \right]$. In other words, the contour plots only represent the aleatoric uncertainty propagated from the input permeability field to the output responses.

\begin{figure}[t!]%
    \centering
    \subfloat[\centering True Mean Pressure]{{\includegraphics[width=0.3\textwidth]{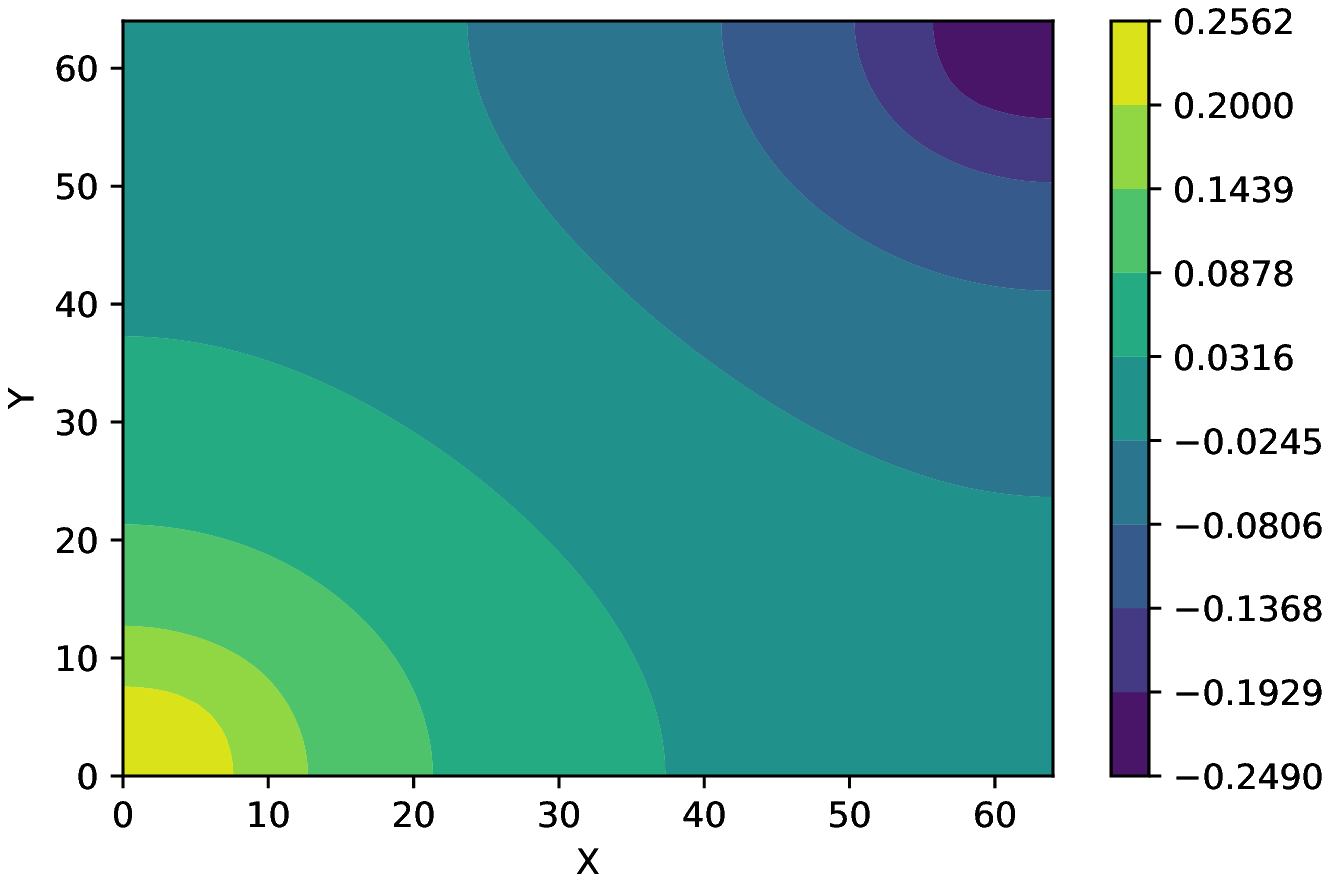}}}%
    \hspace{0.2cm}
    \subfloat[\centering True Mean Velocity(X)]{{\includegraphics[width=0.3\textwidth]{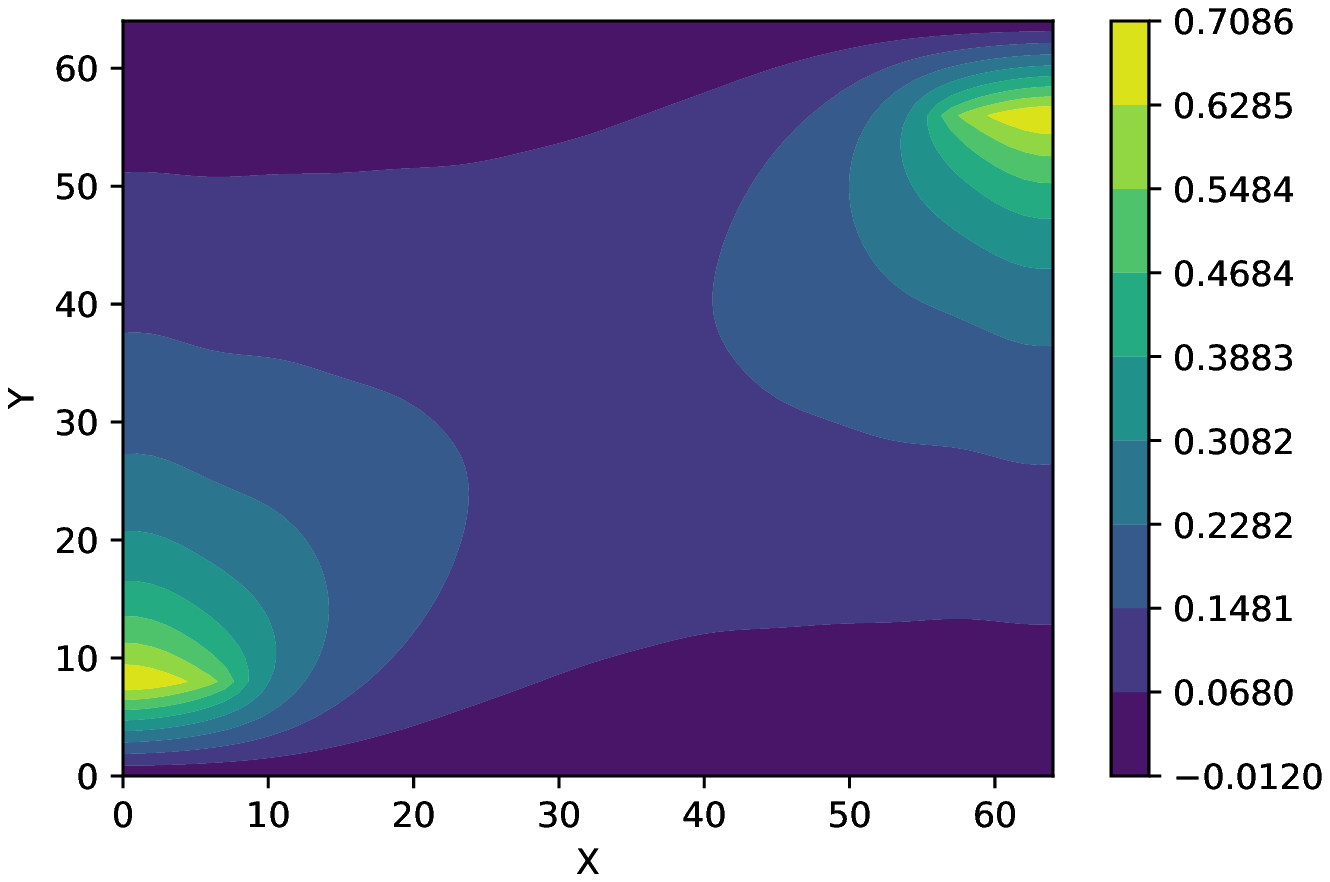}}}%
    \hspace{0.2cm}
    \subfloat[\centering True Mean Velocity(Y)]{{\includegraphics[width=0.3\textwidth]{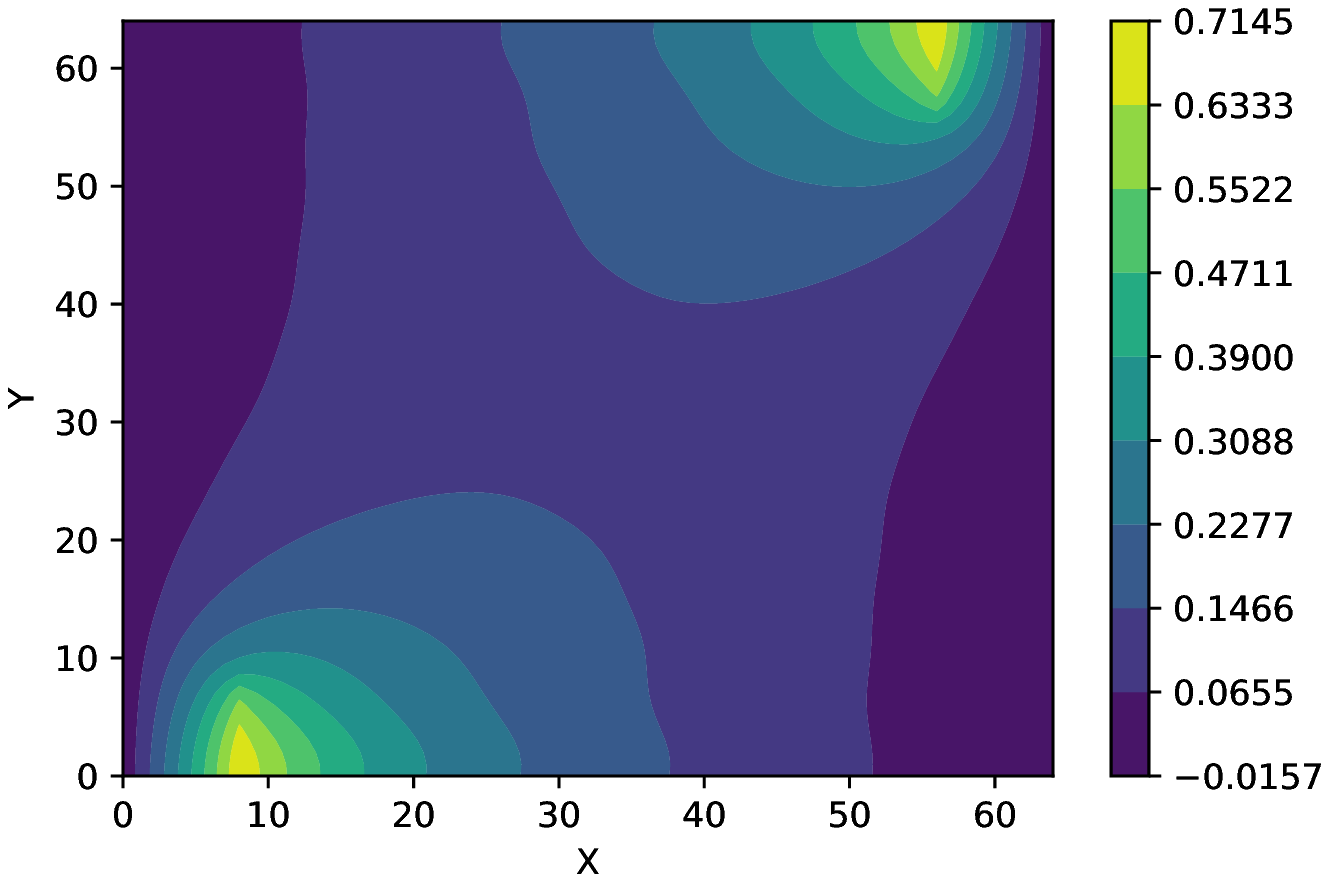}}}%
    \hspace{0.2cm}
    \subfloat[\centering Predicted Mean Pressure]{{\includegraphics[width=0.3\textwidth]{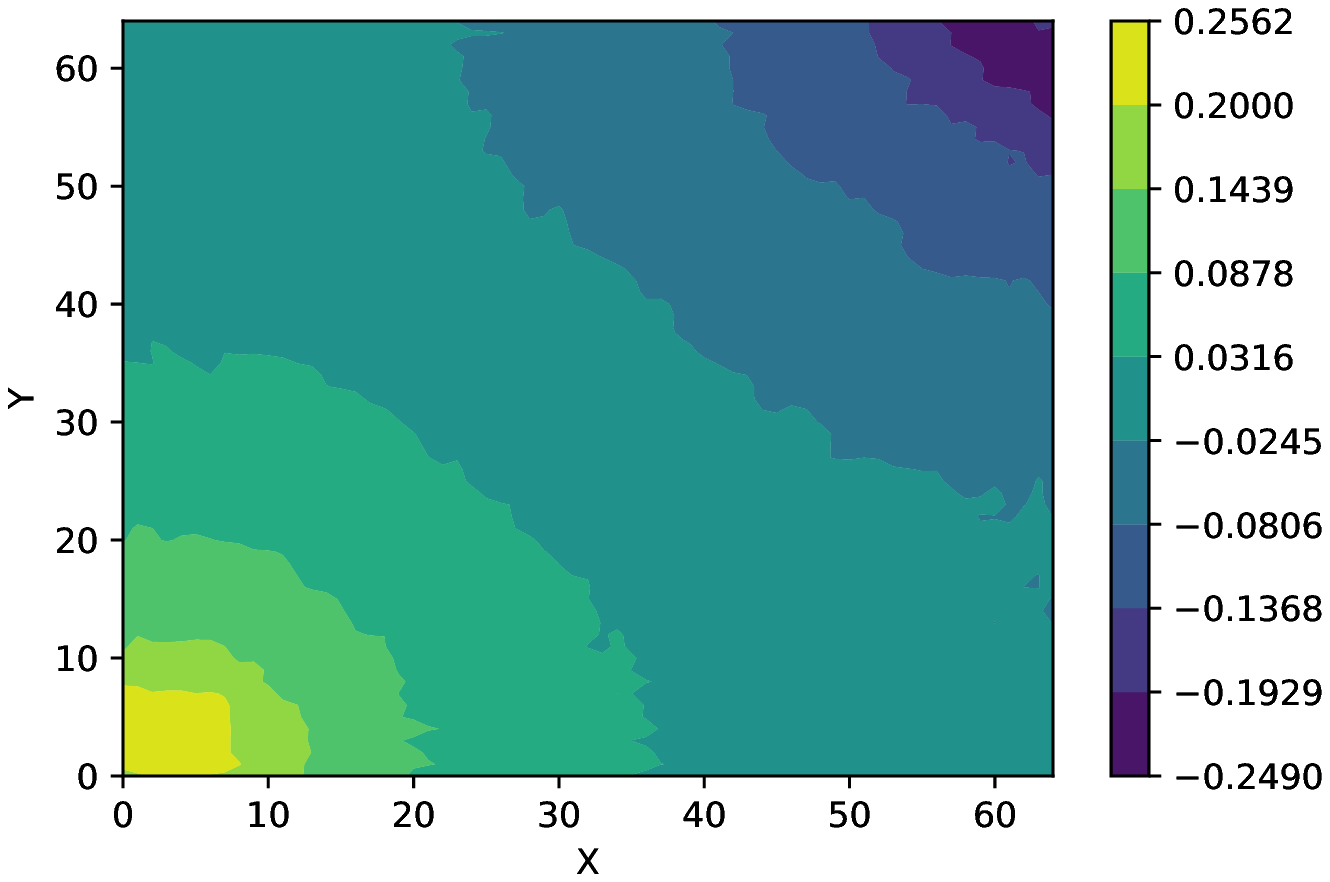}}}%
    \hspace{0.2cm}
    \subfloat[\centering Predicted Mean Velocity(X)]{{\includegraphics[width=0.3\textwidth]{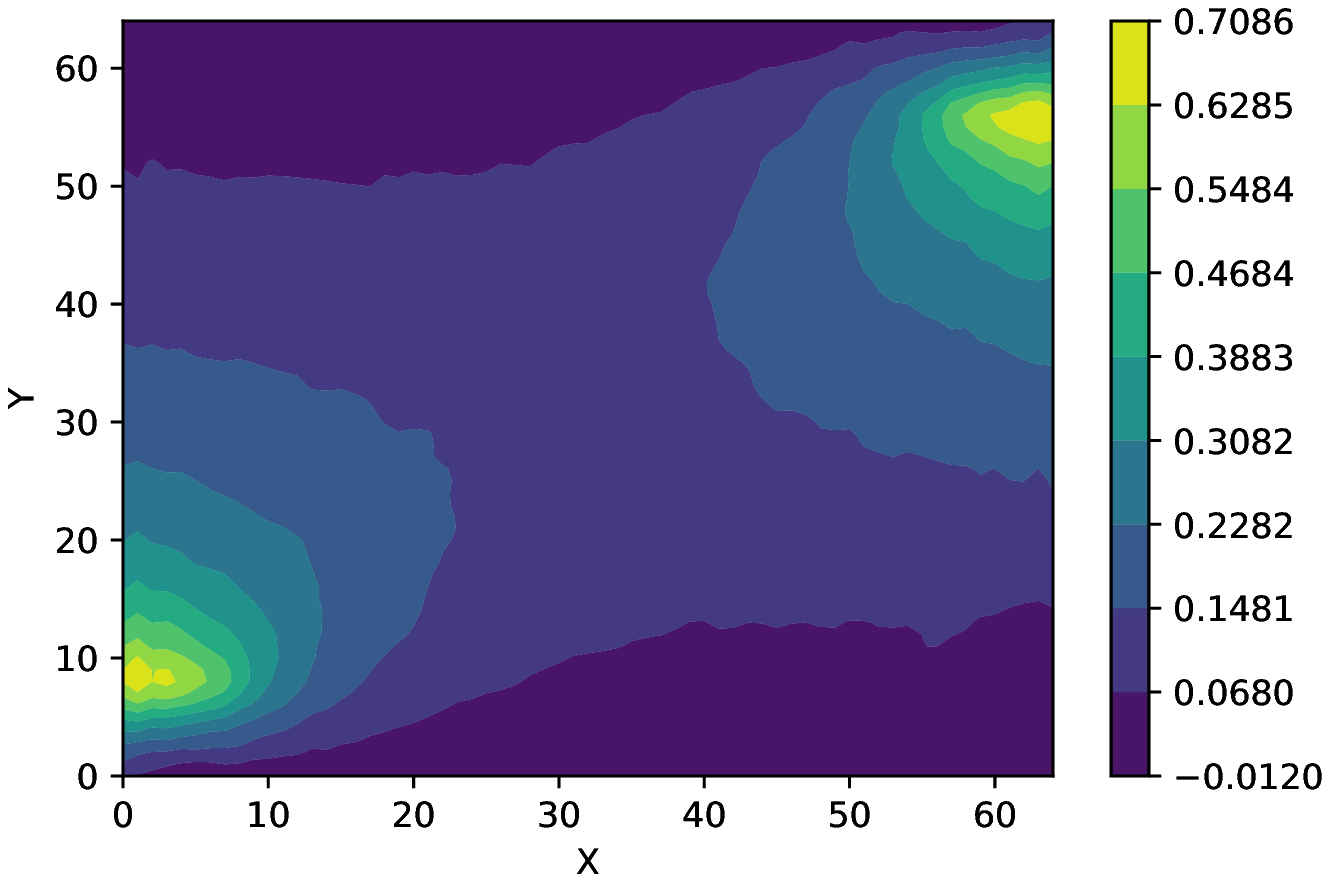}}}%
    \hspace{0.2cm}
    \subfloat[\centering Predicted Mean Velocity(Y)]{{\includegraphics[width=0.3\textwidth]{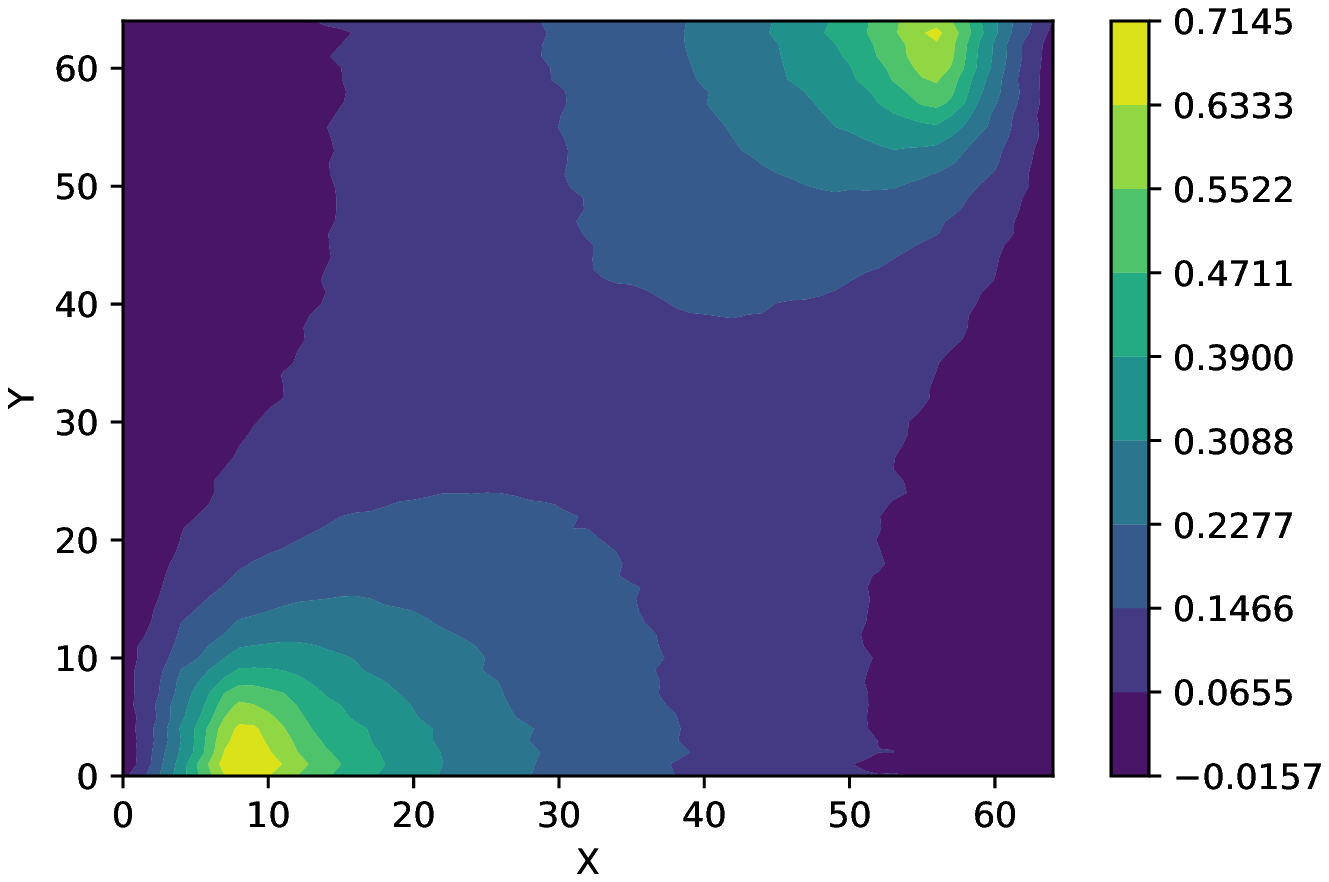}}}%
    \hspace{0.2cm}
    \subfloat[\centering True Variance Pressure]{{\includegraphics[width=0.3\textwidth]{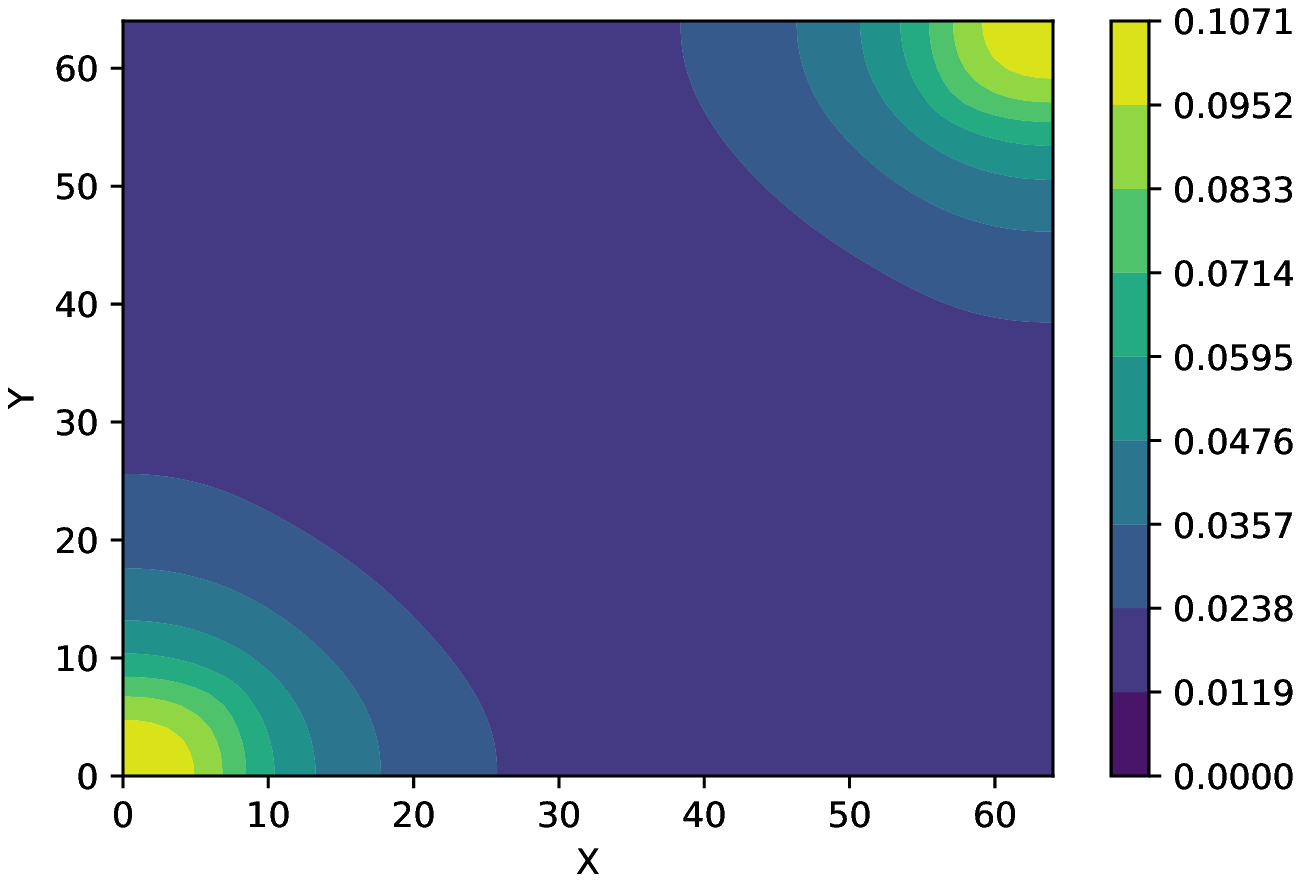}}}%
    \hspace{0.2cm}
    \subfloat[\centering True Variance Velocity(X)]{{\includegraphics[width=0.3\textwidth]{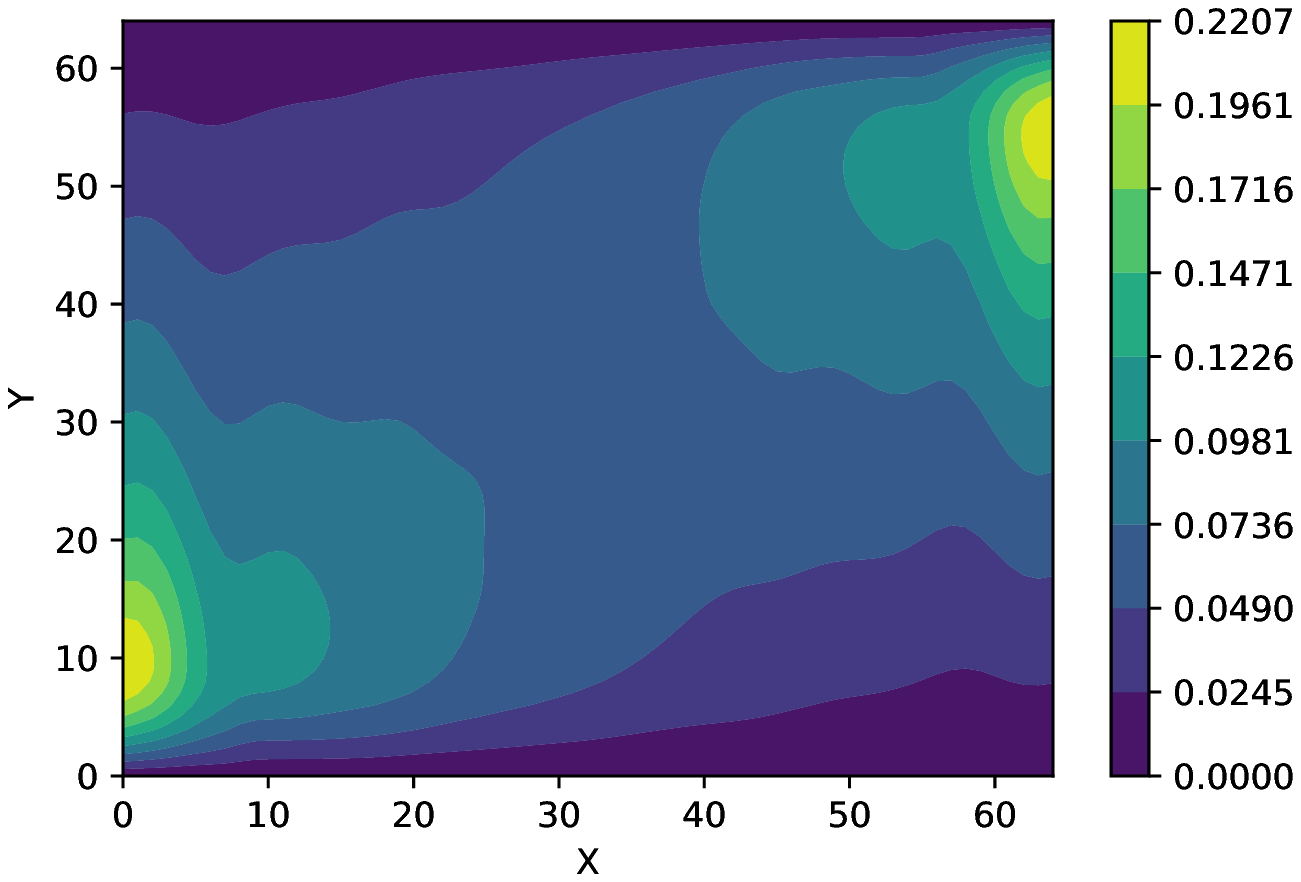}}}%
    \hspace{0.2cm}
    \subfloat[\centering True Variance Velocity(Y)]{{\includegraphics[width=0.3\textwidth]{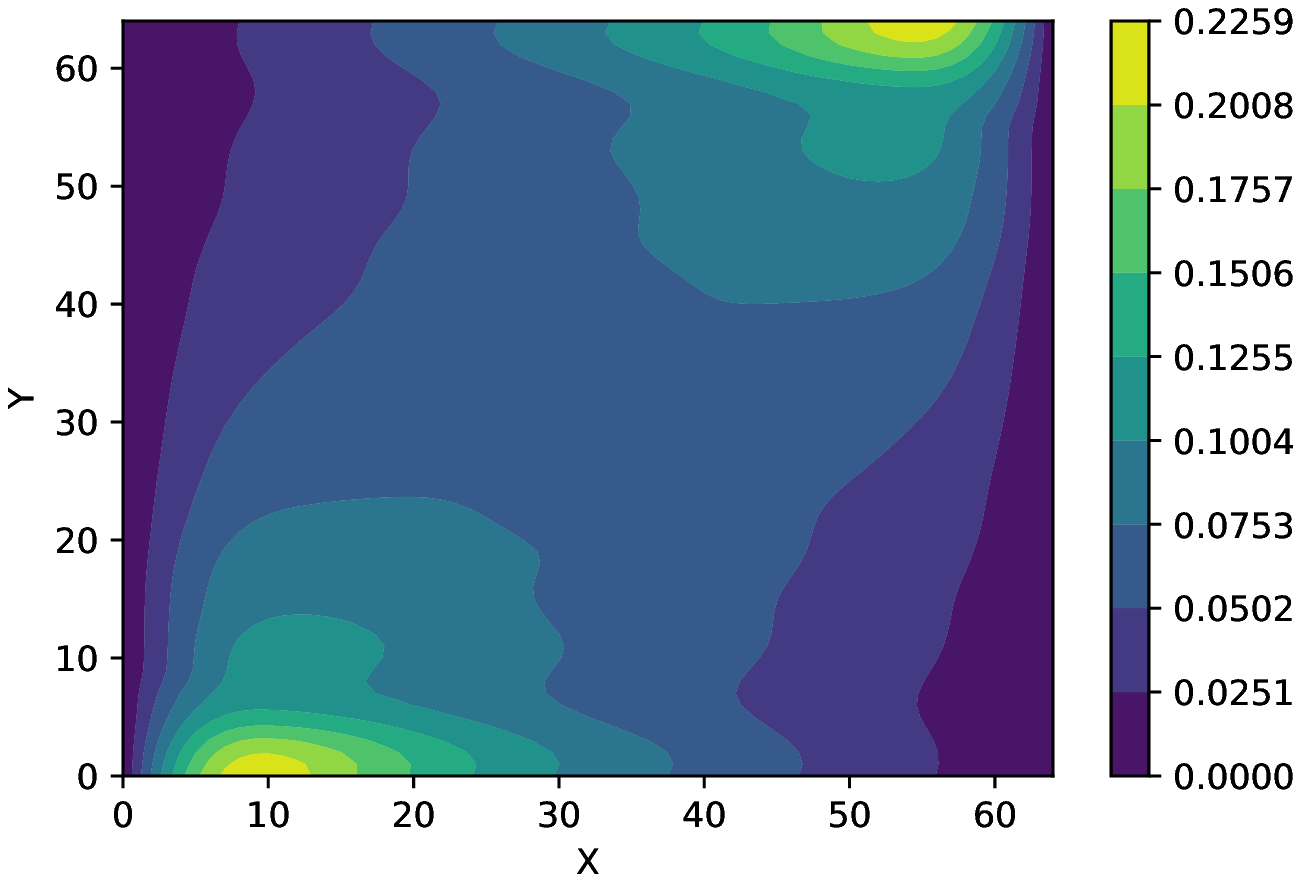}}}%
    \hspace{0.2cm}
    \subfloat[\centering Predicted Variance Pressure]{{\includegraphics[width=0.3\textwidth]{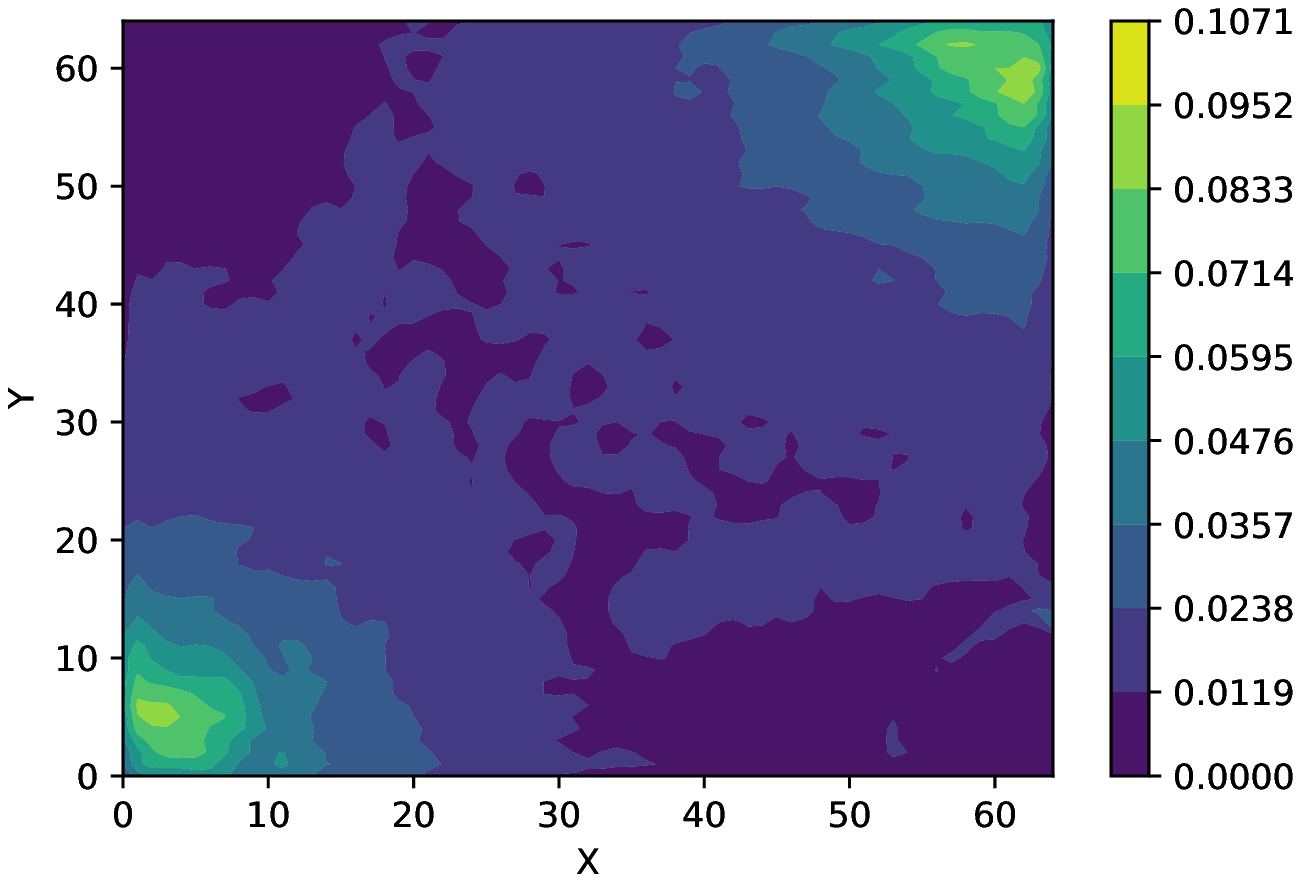}}}%
    \hspace{0.2cm}
    \subfloat[\centering Predicted Variance Velocity(X)]{{\includegraphics[width=0.3\textwidth]{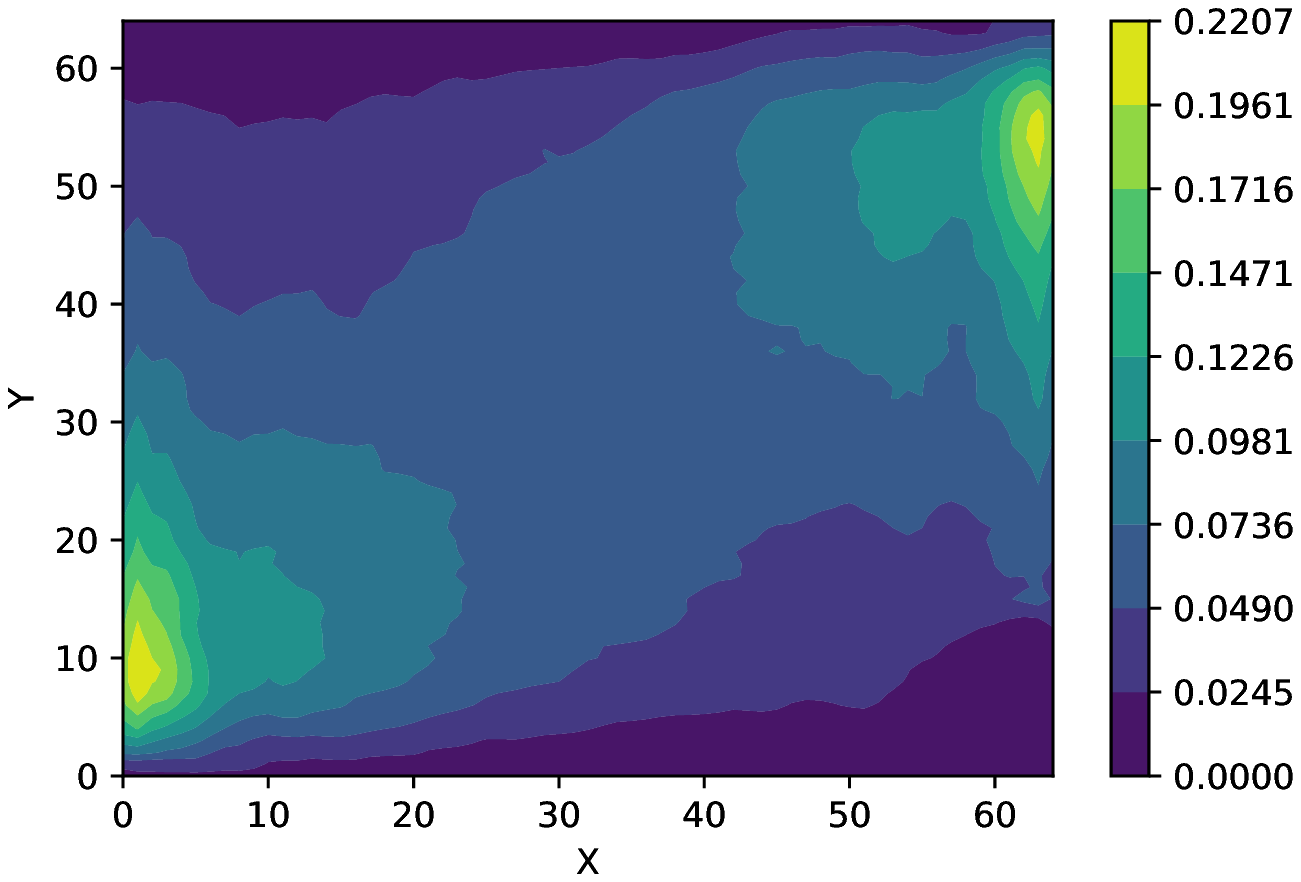}}}%
    \hspace{0.2cm}
    \subfloat[\centering Predicted Variance Velocity(Y)]{{\includegraphics[width=0.3\textwidth]{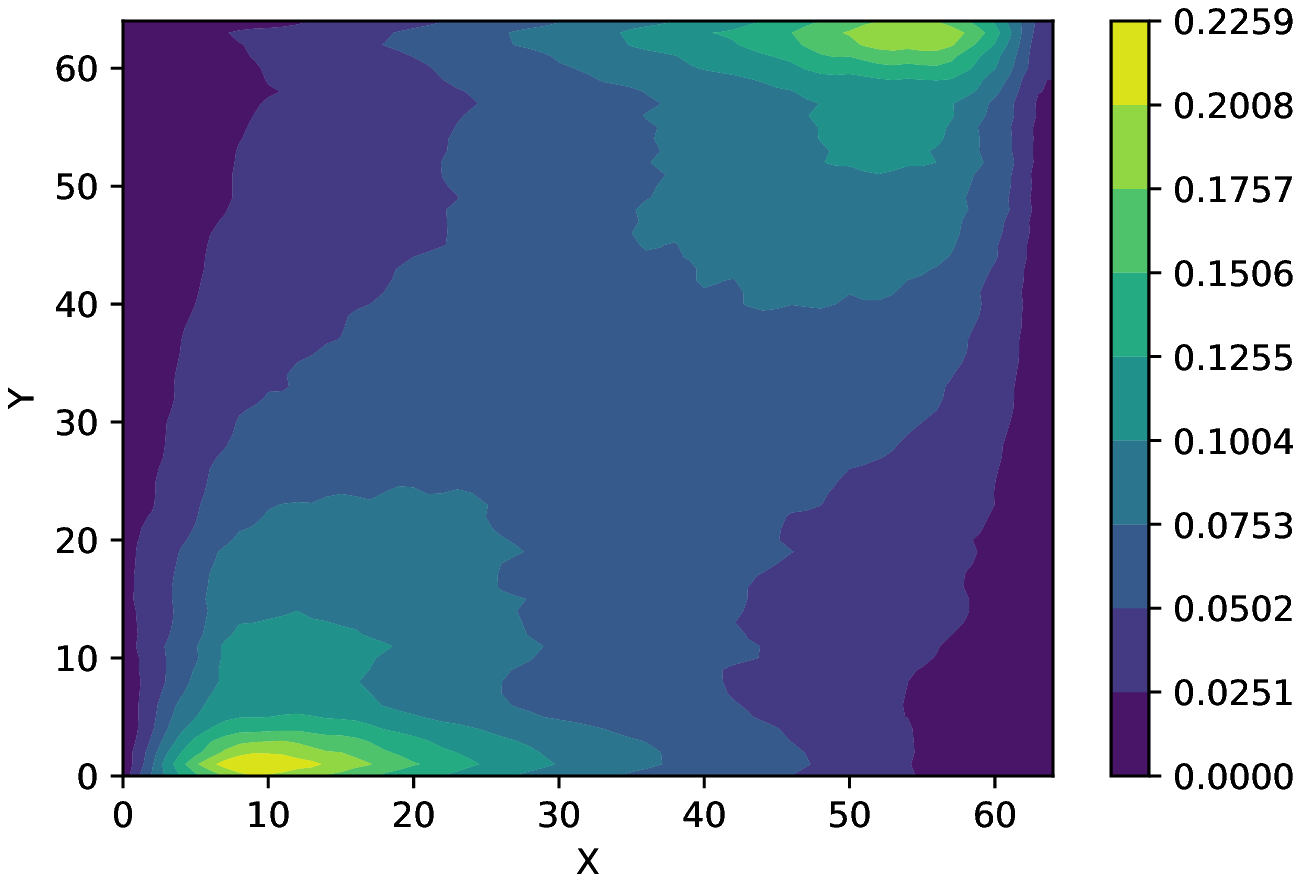}}}%
    \caption{Aleatoric uncertainty propagation using GLU-Net for KLE50 trained on 256 samples. The first and the third row correspond to the contour plot of the ground truth mean and the ground truth standard deviation. Second and fourth row corresponds to the results obtained using GLU-net.}%
    \label{fig:alea1}%
\end{figure}

\begin{figure}[t!]%
    \centering
    \subfloat[\centering Actual Mean Pressure]{{\includegraphics[width=0.3\textwidth]{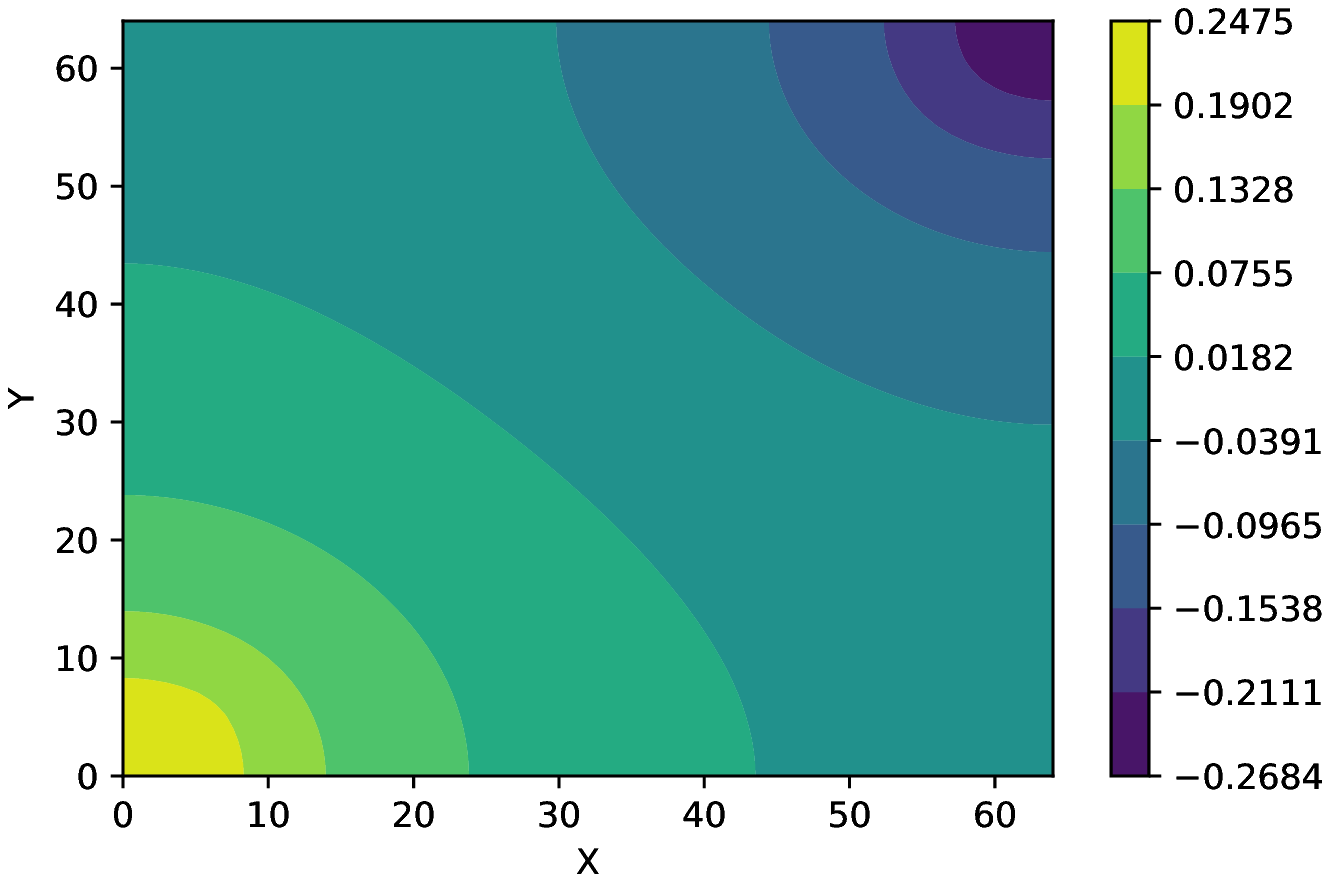}}}%
    \hspace{0.2cm}
    \subfloat[\centering Actual Mean Velocity(X)]{{\includegraphics[width=0.3\textwidth]{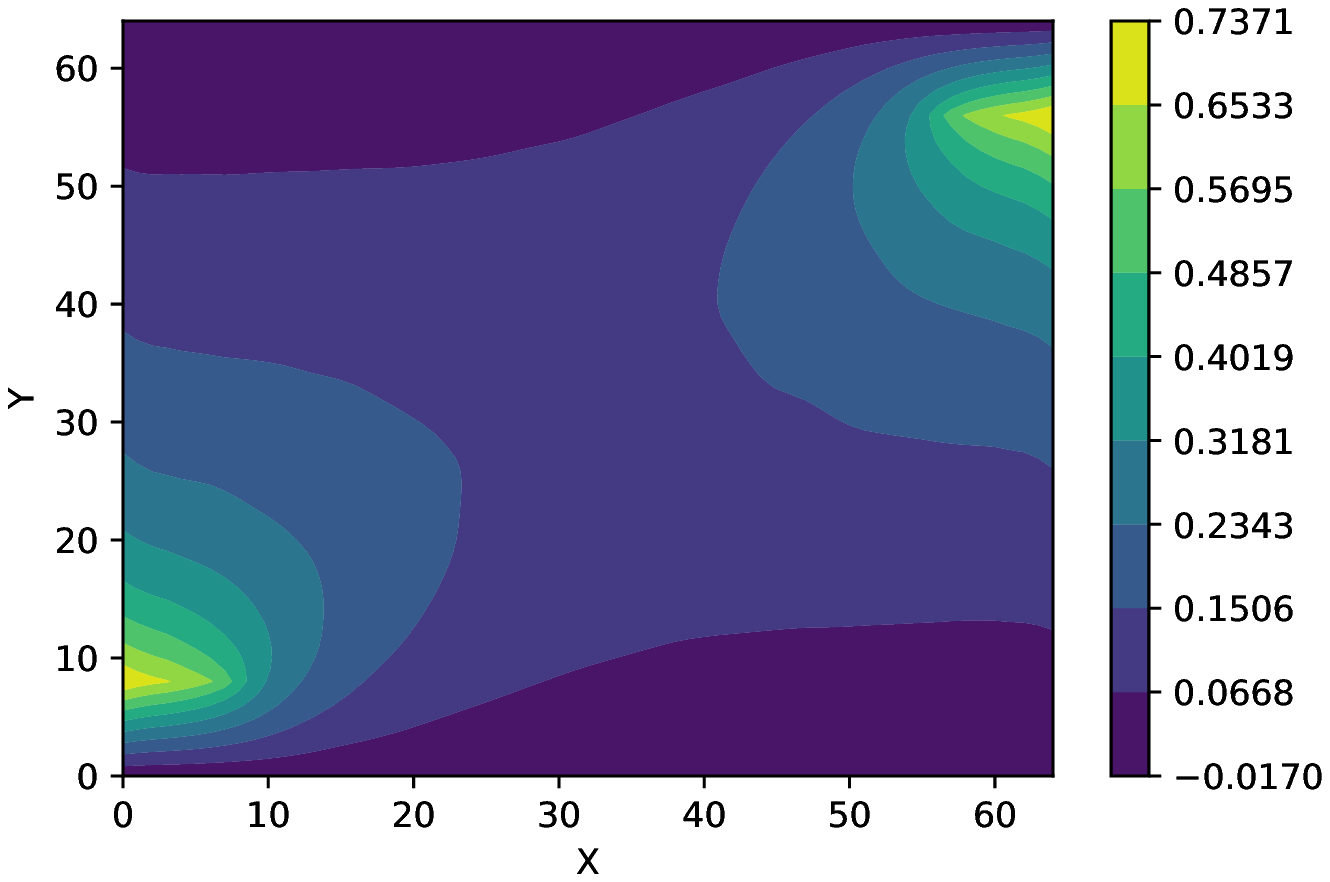}}}%
    \hspace{0.2cm}
    \subfloat[\centering Actual Mean Velocity(Y)]{{\includegraphics[width=0.3\textwidth]{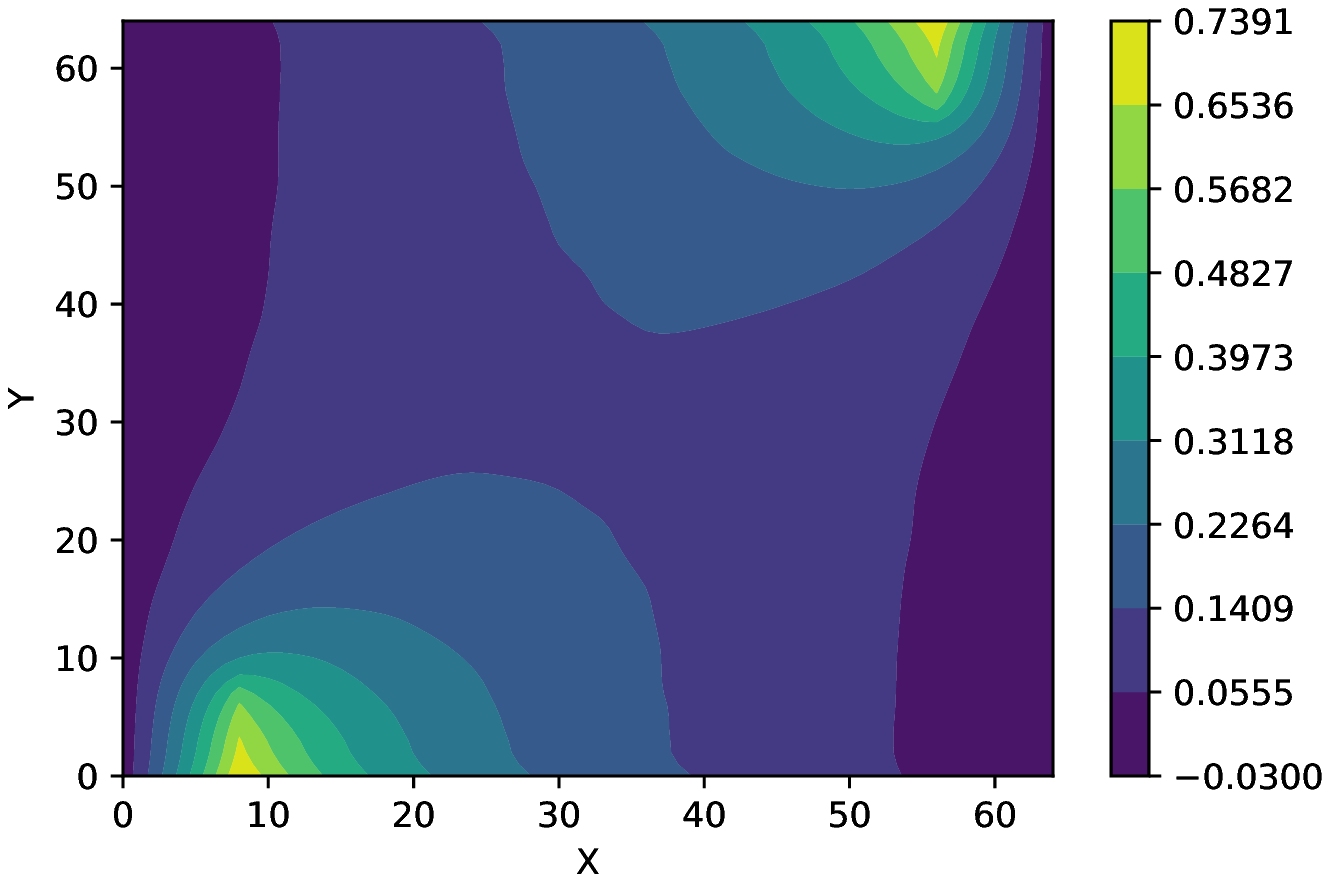}}}%
    \hspace{0.2cm}
    \subfloat[\centering Predicted Mean Pressure]{{\includegraphics[width=0.3\textwidth]{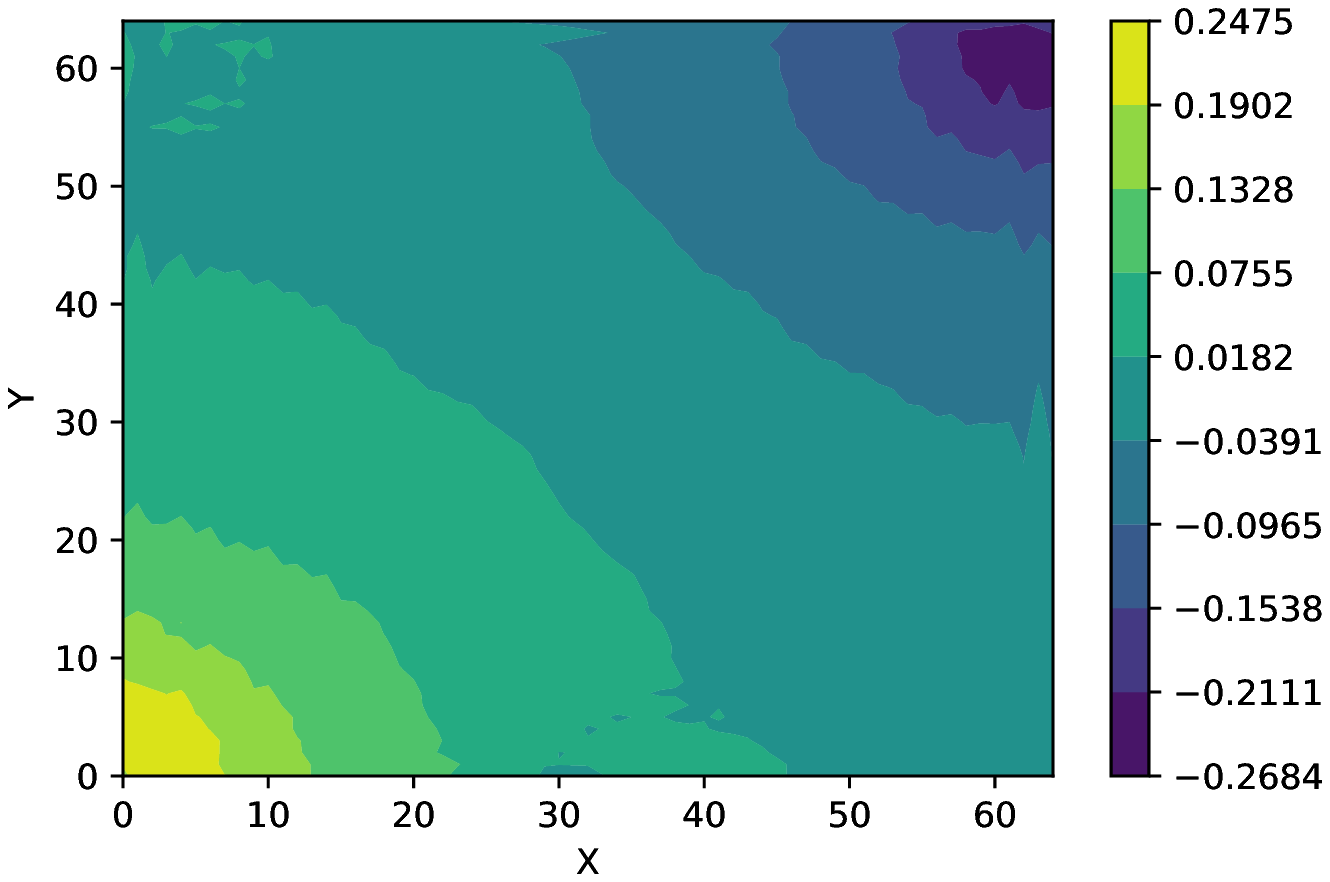}}}%
    \hspace{0.2cm}
    \subfloat[\centering Predicted Mean Velocity(X)]{{\includegraphics[width=0.3\textwidth]{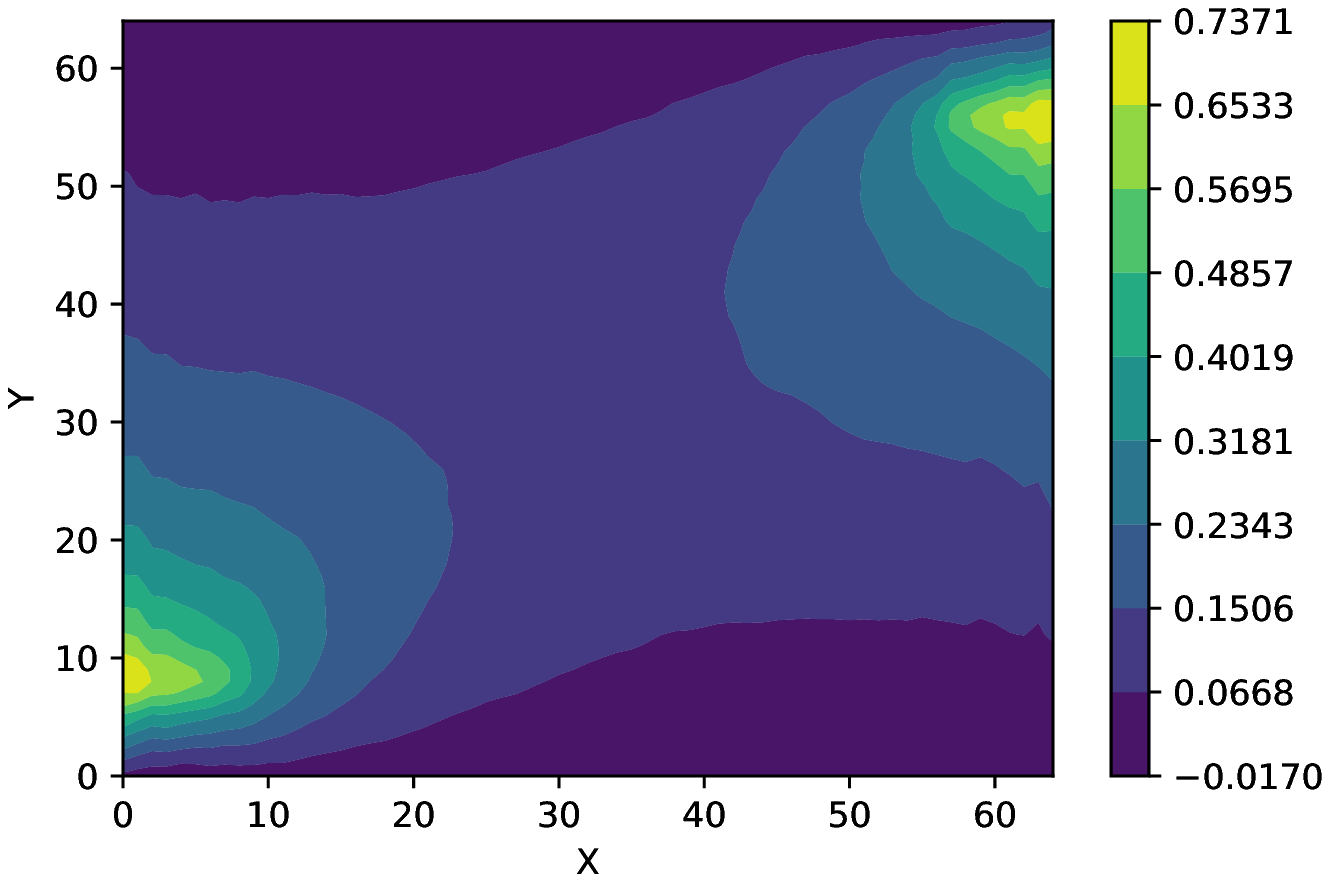}}}%
    \hspace{0.2cm}
    \subfloat[\centering Predicted Mean Velocity(Y)]{{\includegraphics[width=0.3\textwidth]{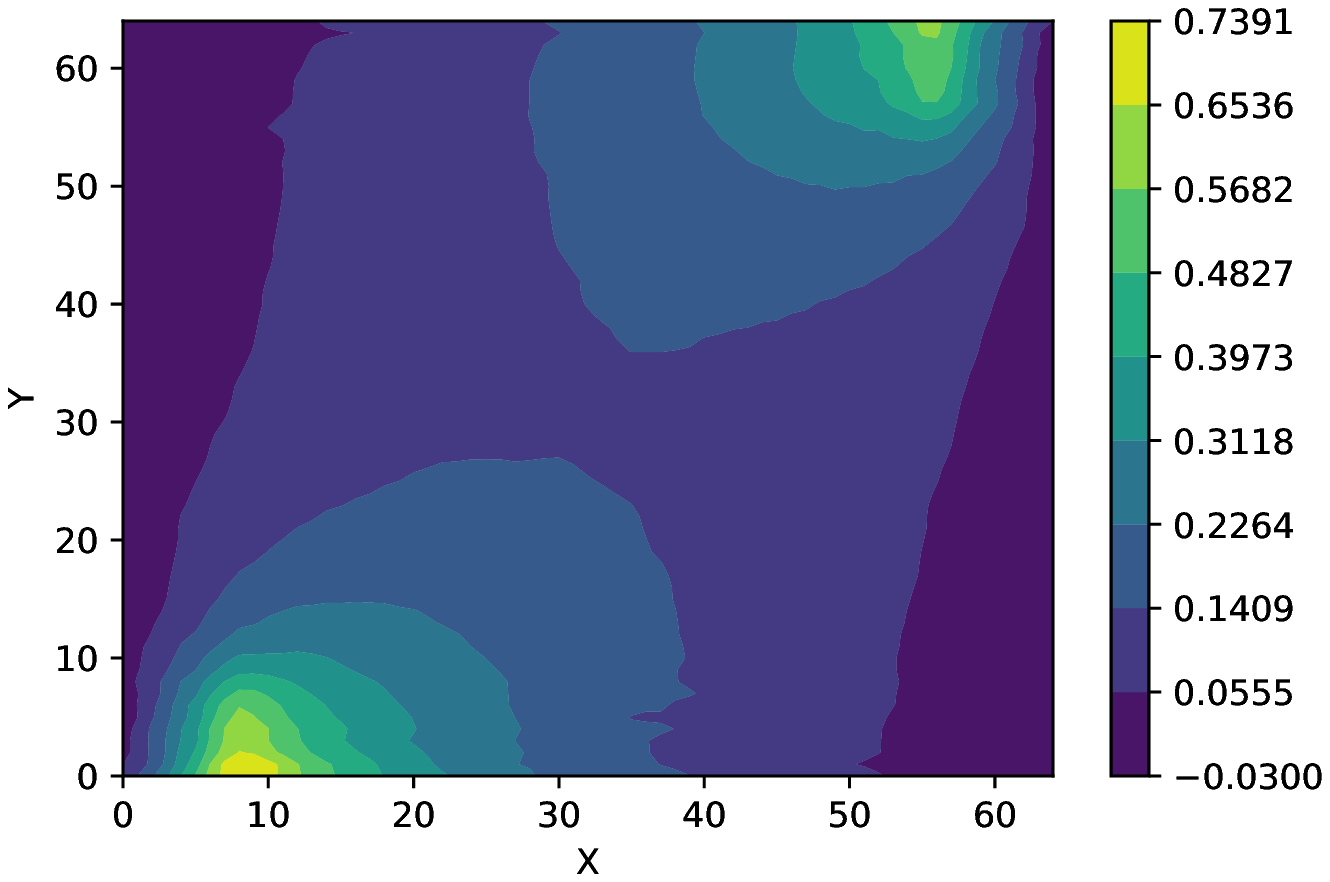}}}%
    \hspace{0.2cm}
    \subfloat[\centering Actual Variance Pressure]{{\includegraphics[width=0.3\textwidth]{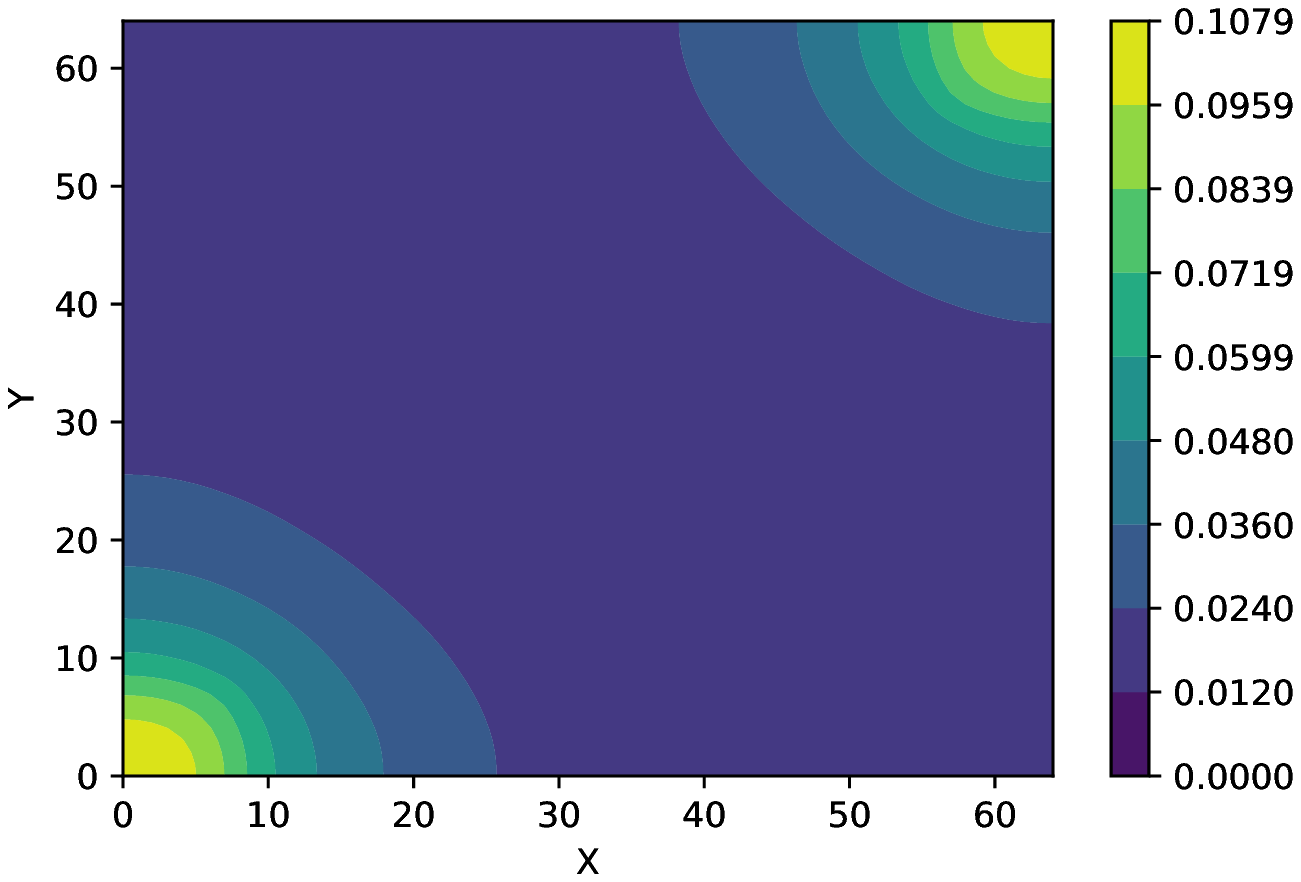}}}%
    \hspace{0.2cm}
    \subfloat[\centering Actual Variance Velocity(X)]{{\includegraphics[width=0.3\textwidth]{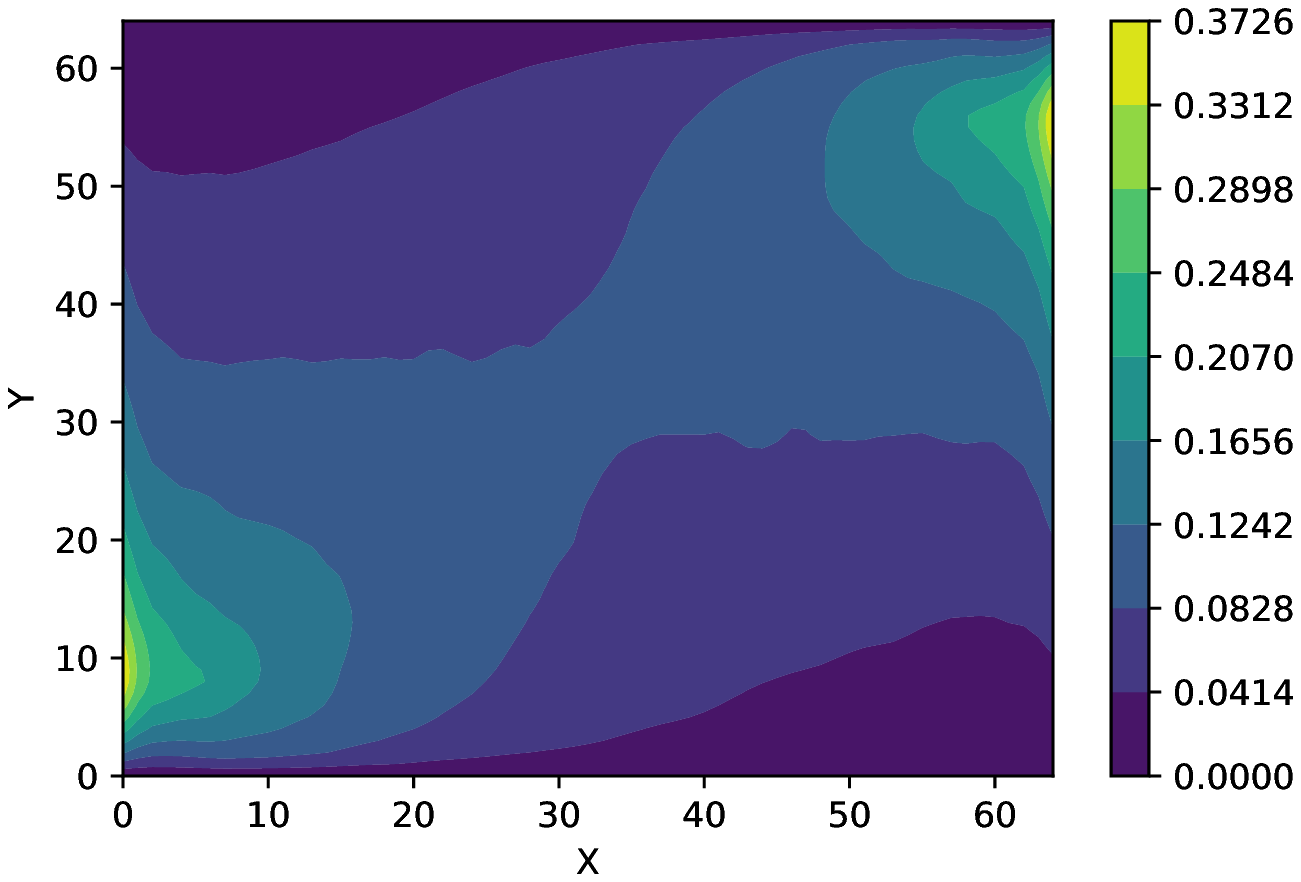}}}%
    \hspace{0.2cm}
    \subfloat[\centering Actual Variance Velocity(Y)]{{\includegraphics[width=0.3\textwidth]{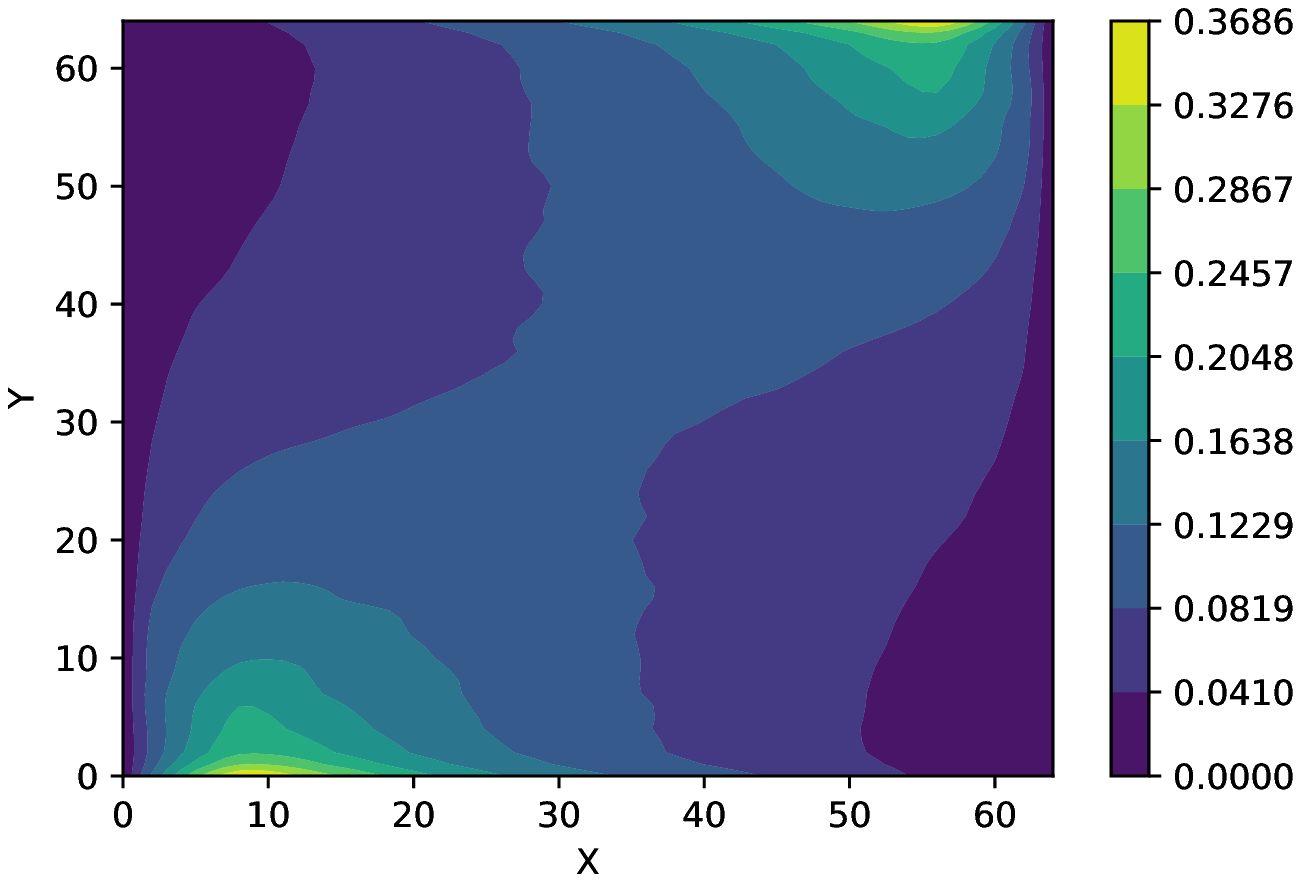}}}%
    \hspace{0.2cm}
    \subfloat[\centering Predicted Variance Pressure]{{\includegraphics[width=0.3\textwidth]{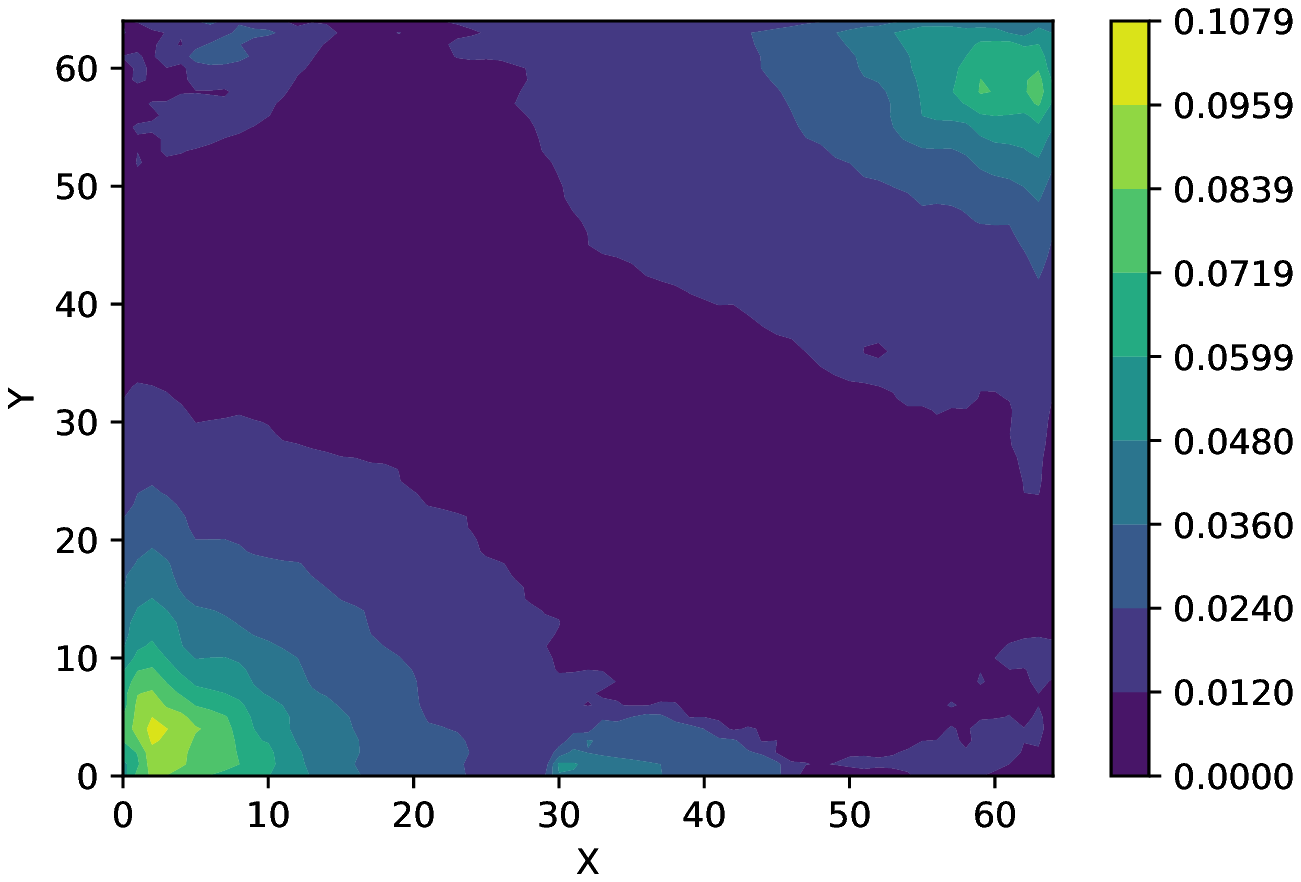}}}%
    \hspace{0.2cm}
    \subfloat[\centering Predicted Variance Velocity(X)]{{\includegraphics[width=0.3\textwidth]{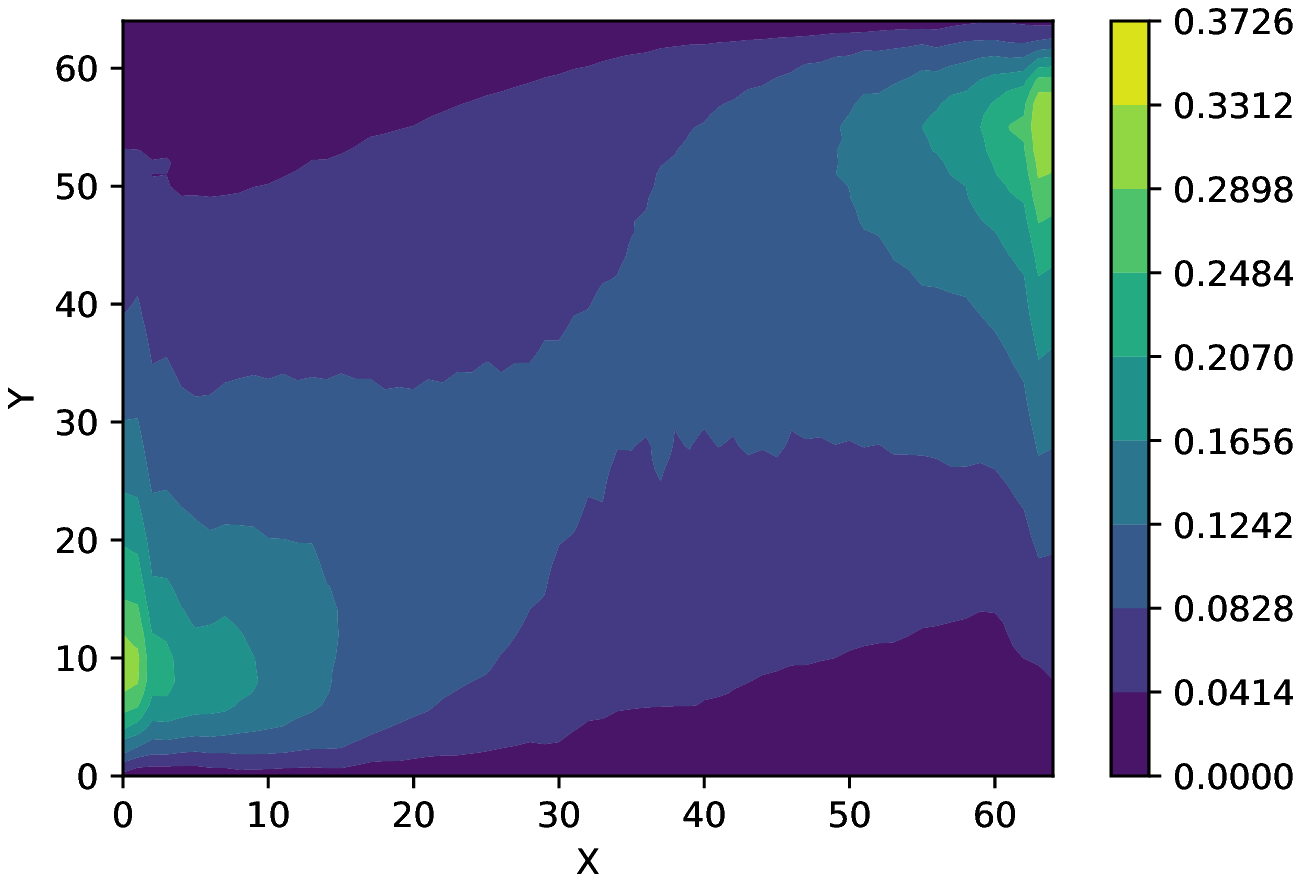}}}%
    \hspace{0.2cm}
    \subfloat[\centering Predicted Variance Velocity(Y)]{{\includegraphics[width=0.3\textwidth]{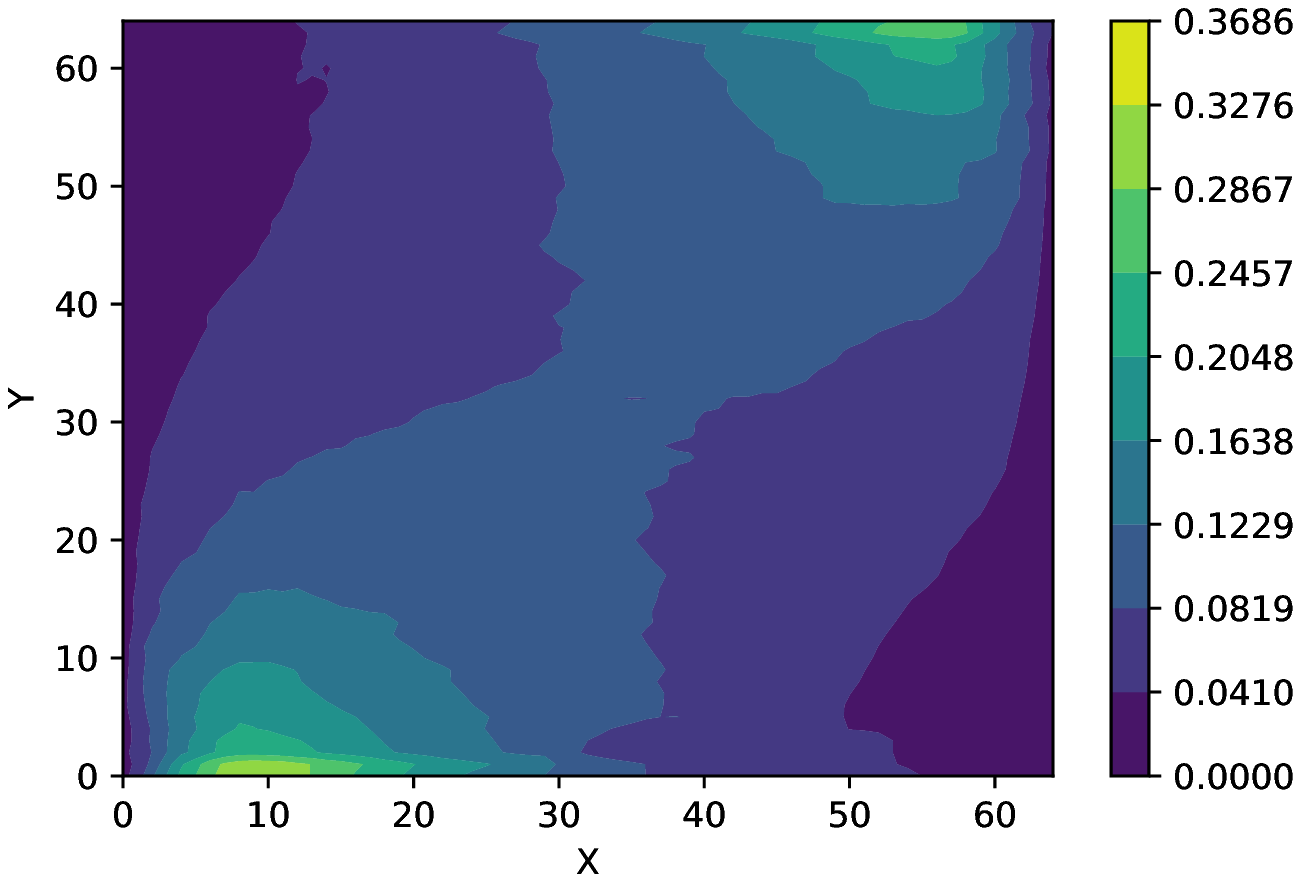}}}%
    \caption{Aleatoric uncertainty propagation of GLU-Net for KLE500 trained on 512 training samples. For each row descriptions, refer to Fig. \ref{fig:alea1}.}%
    \label{fig:alea2}%
\end{figure}

\begin{figure}[t!]%
    \centering
    \subfloat[\centering Actual Mean Pressure]{{\includegraphics[width=0.3\textwidth]{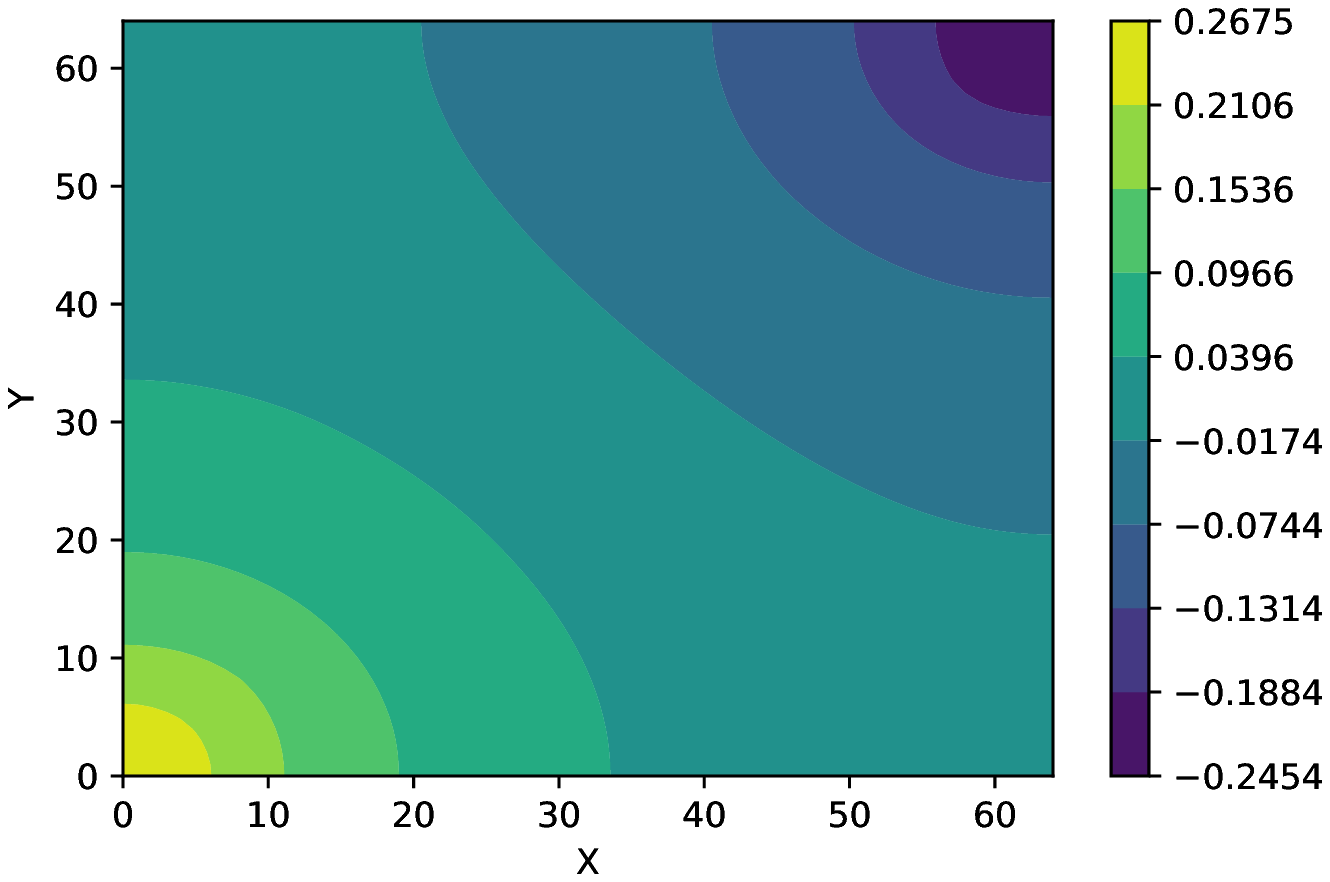}}}%
    \hspace{0.2cm}
    \subfloat[\centering Actual Mean Velocity(X)]{{\includegraphics[width=0.3\textwidth]{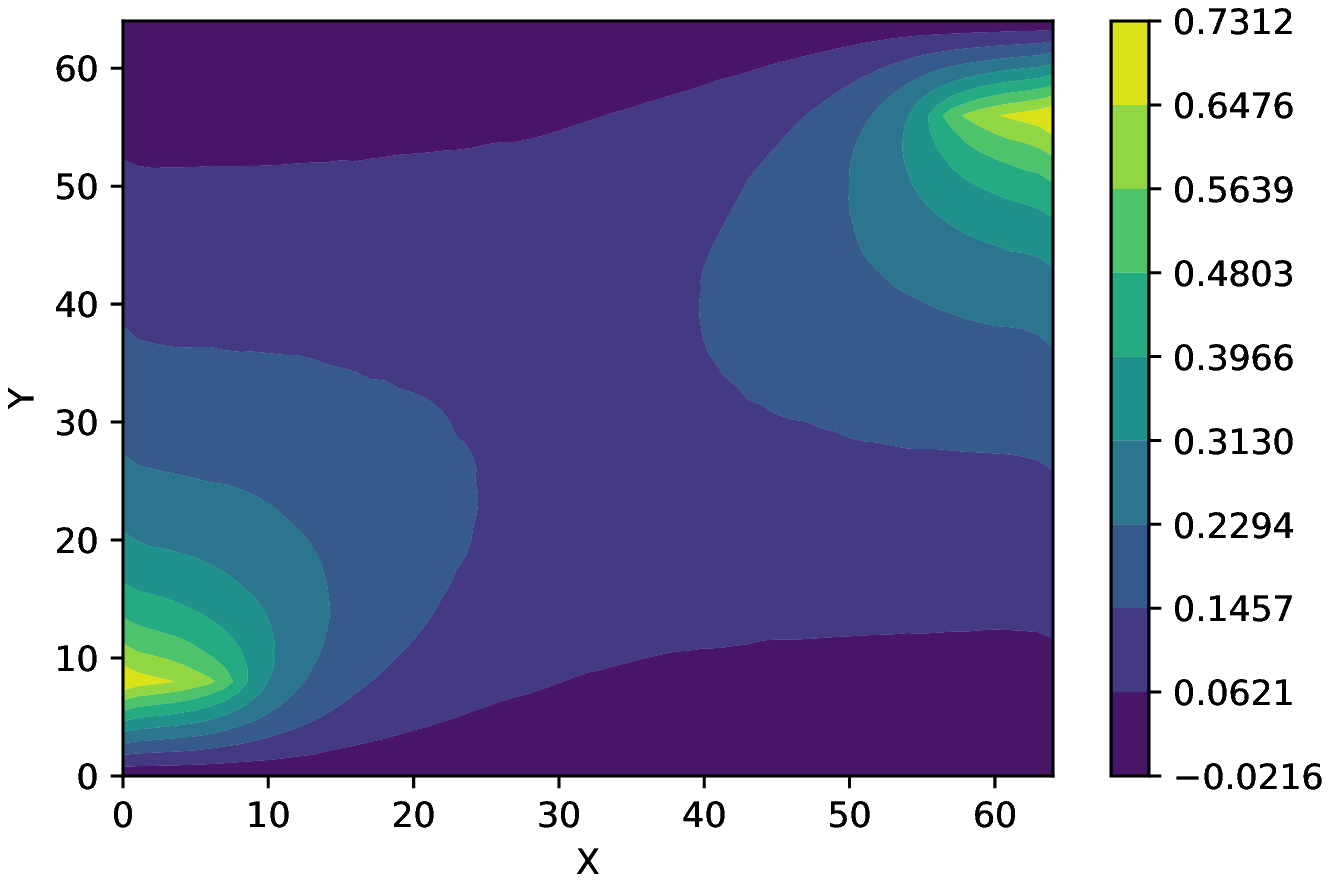}}}%
    \hspace{0.2cm}
    \subfloat[\centering Actual Mean Velocity(Y)]{{\includegraphics[width=0.3\textwidth]{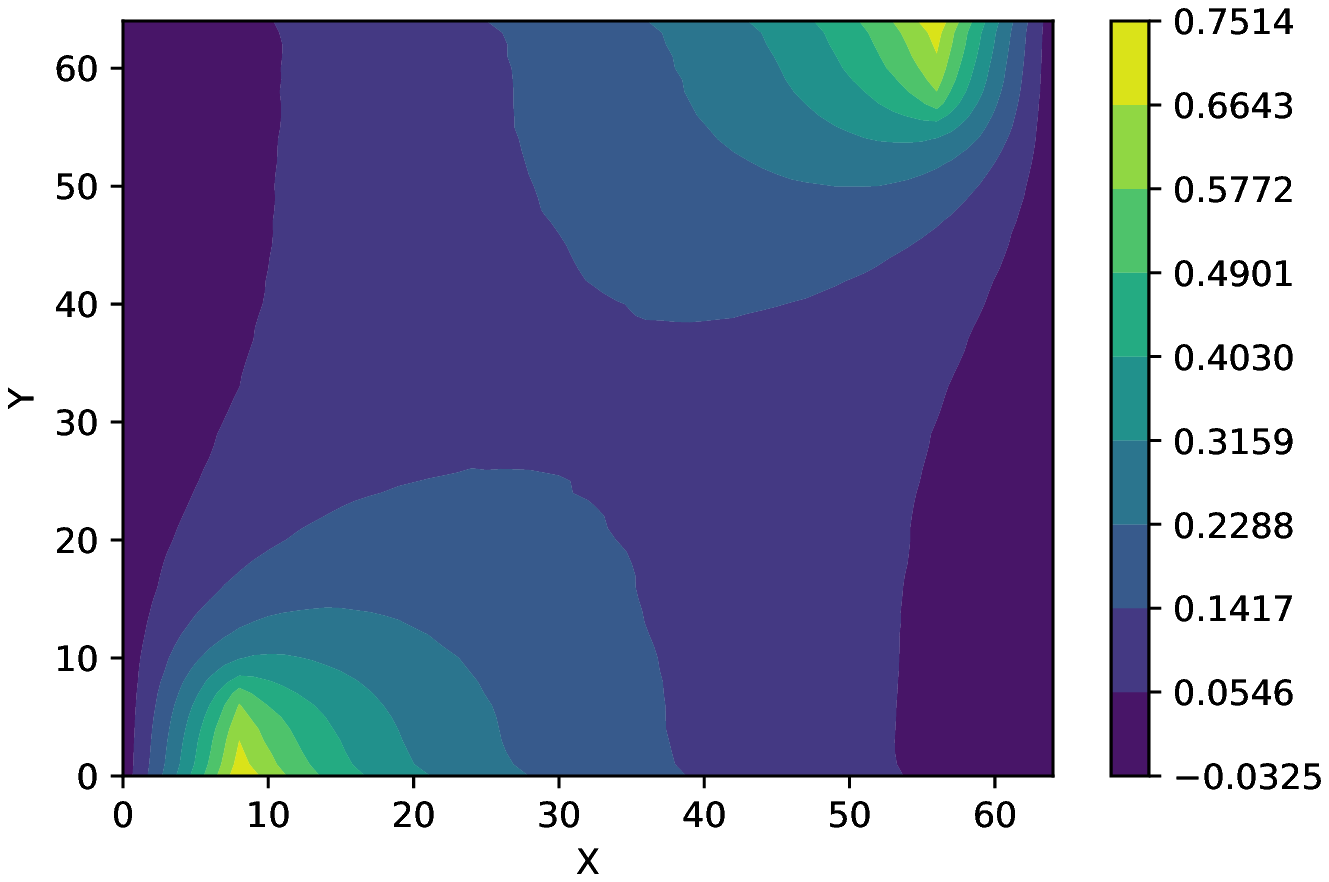}}}%
    \hspace{0.2cm}
    \subfloat[\centering Predicted Mean Pressure]{{\includegraphics[width=0.3\textwidth]{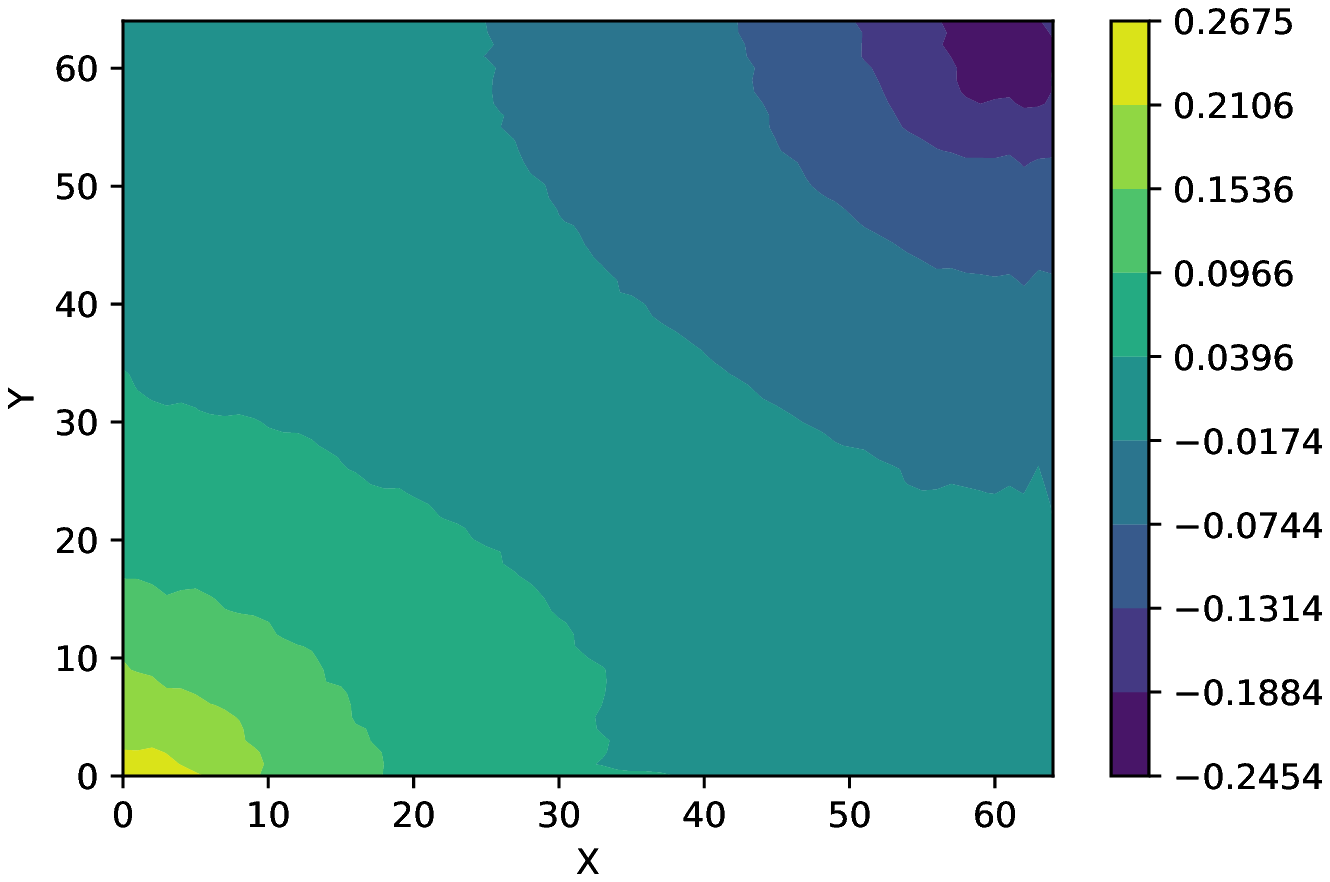}}}%
    \hspace{0.2cm}
    \subfloat[\centering Predicted Mean Velocity(X)]{{\includegraphics[width=0.3\textwidth]{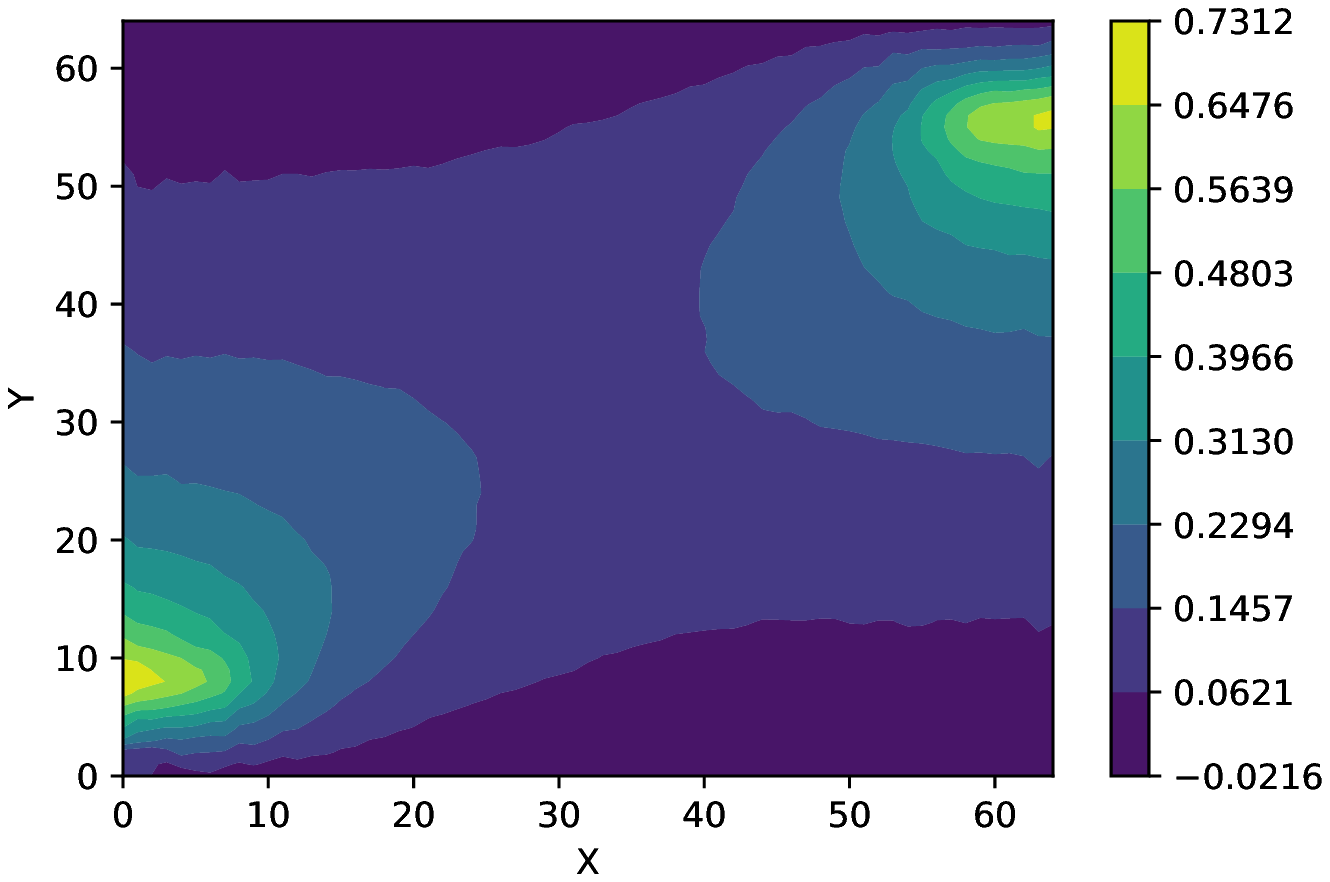}}}%
    \hspace{0.2cm}
    \subfloat[\centering Predicted Mean Velocity(Y)]{{\includegraphics[width=0.3\textwidth]{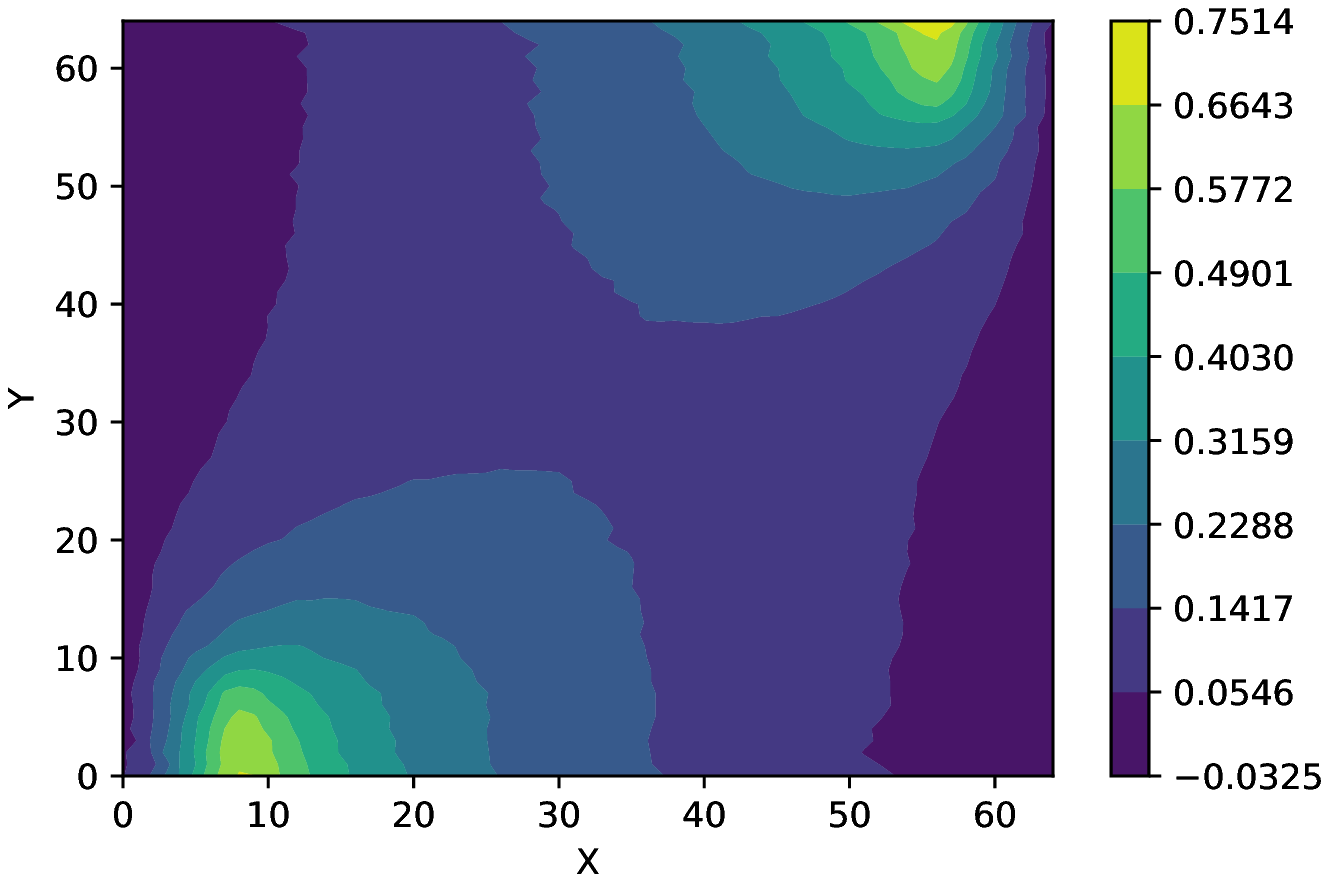}}}%
    \hspace{0.2cm}
    \subfloat[\centering Actual Variance Pressure]{{\includegraphics[width=0.3\textwidth]{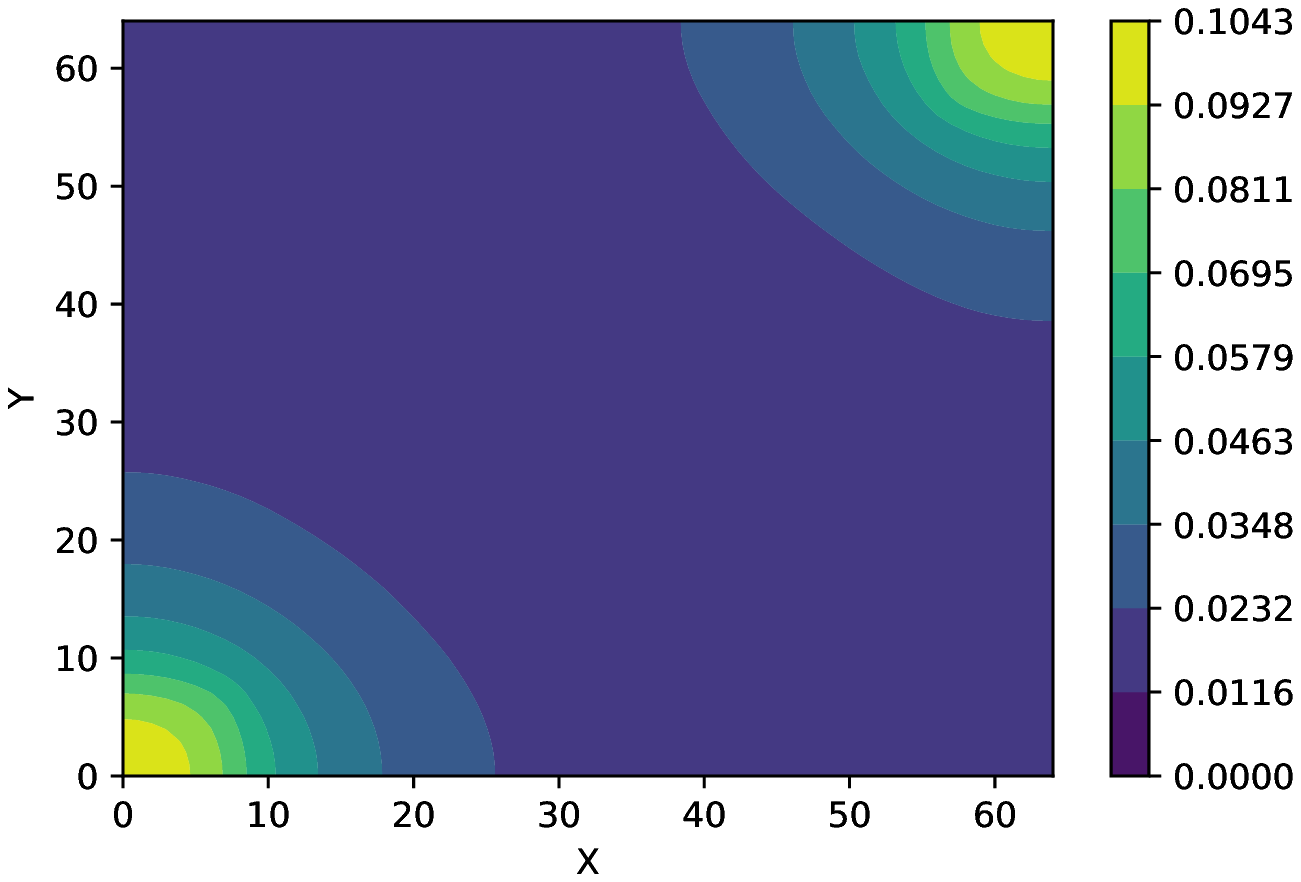}}}%
    \hspace{0.2cm}
    \subfloat[\centering Actual Variance Velocity(X)]{{\includegraphics[width=0.3\textwidth]{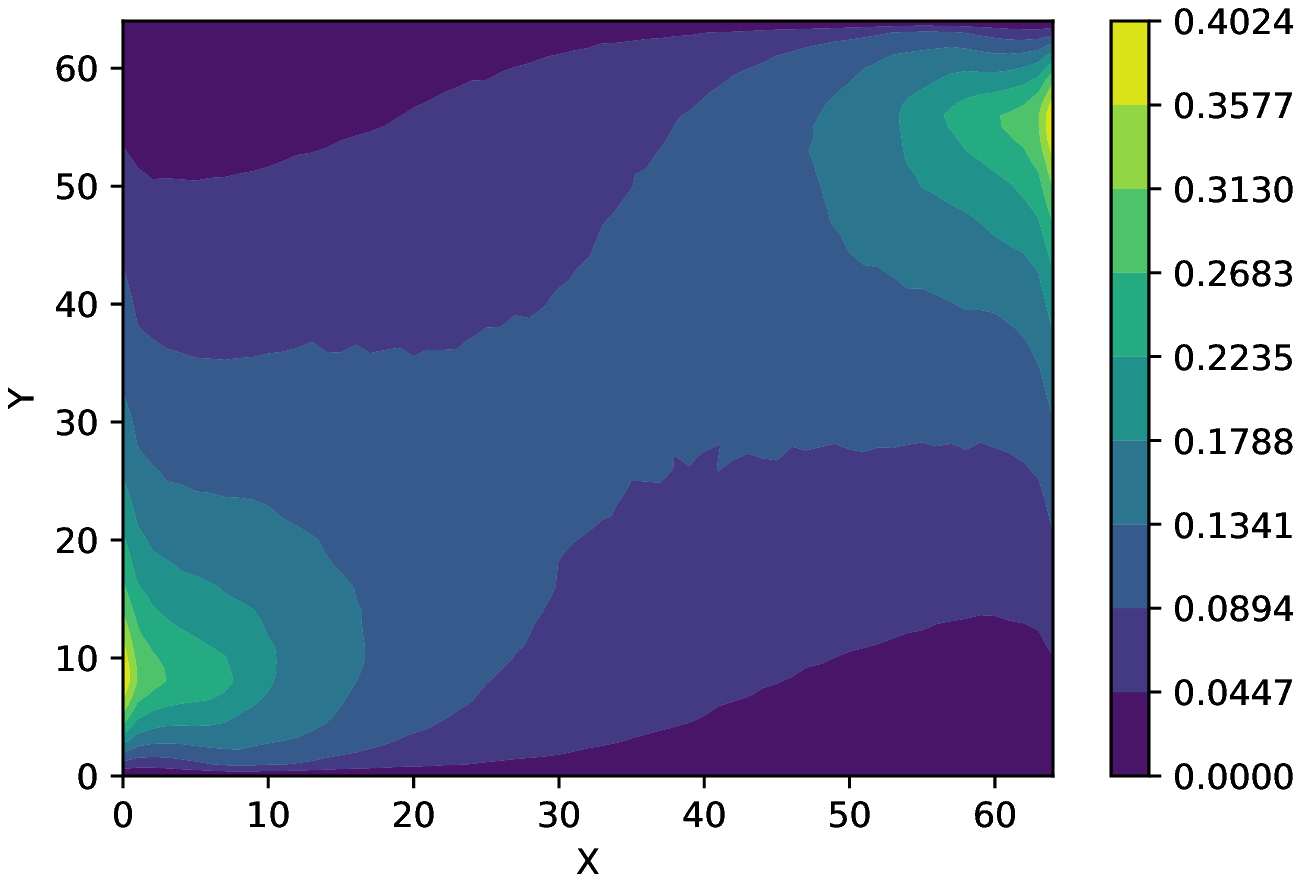}}}%
    \hspace{0.2cm}
    \subfloat[\centering Actual Variance Velocity(Y)]{{\includegraphics[width=0.3\textwidth]{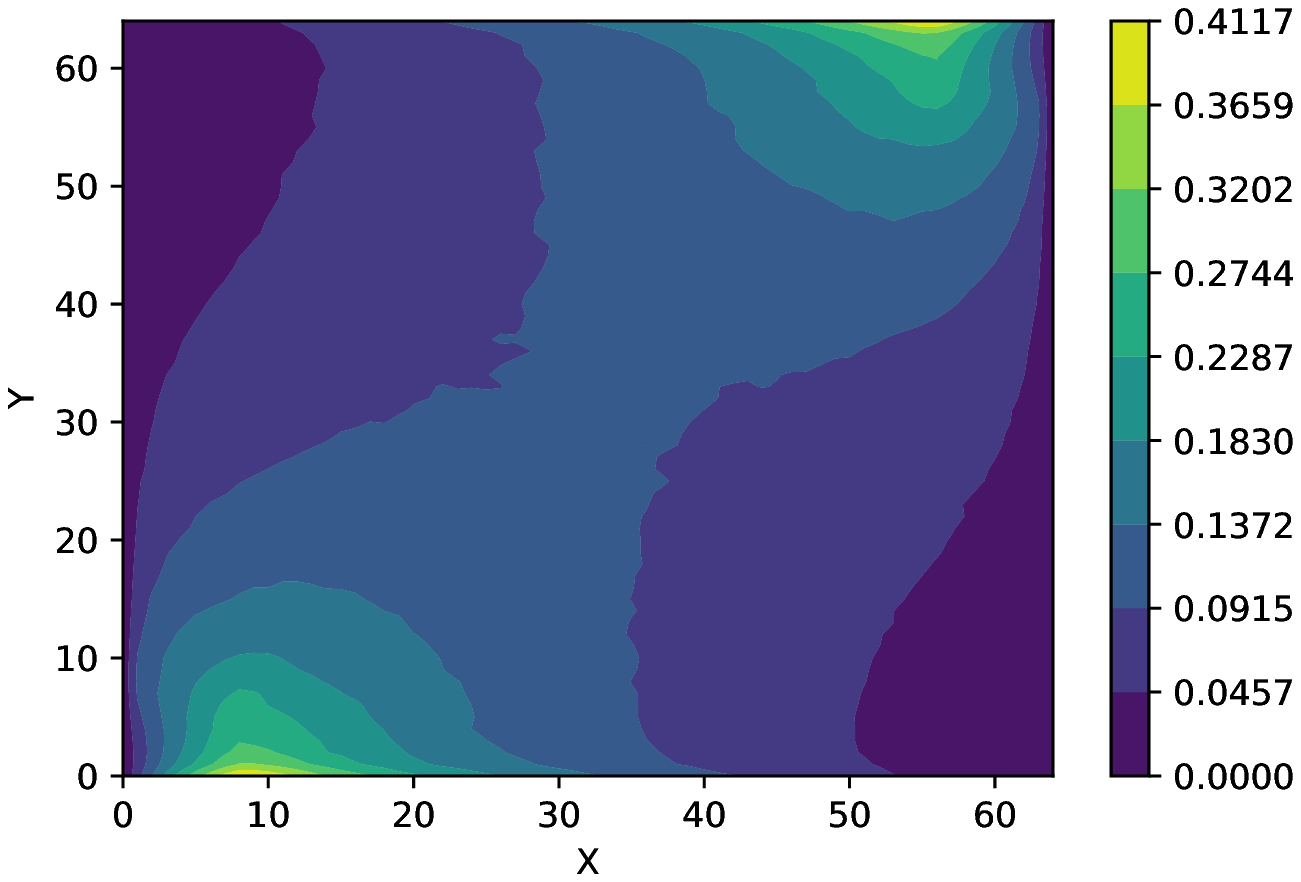}}}%
    \hspace{0.2cm}
    \subfloat[\centering Predicted Variance Pressure]{{\includegraphics[width=0.3\textwidth]{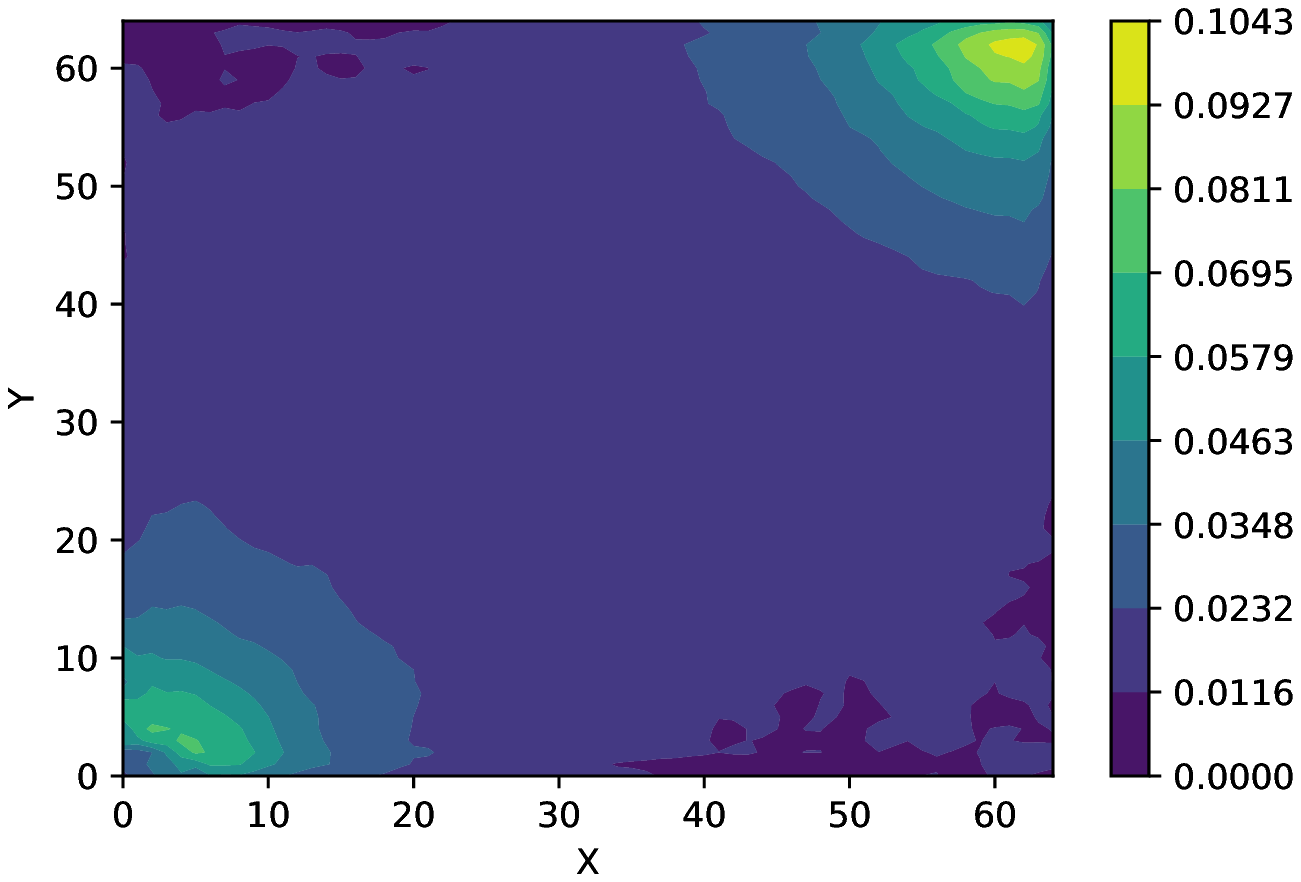}}}%
    \hspace{0.2cm}
    \subfloat[\centering Predicted Variance Velocity(X)]{{\includegraphics[width=0.3\textwidth]{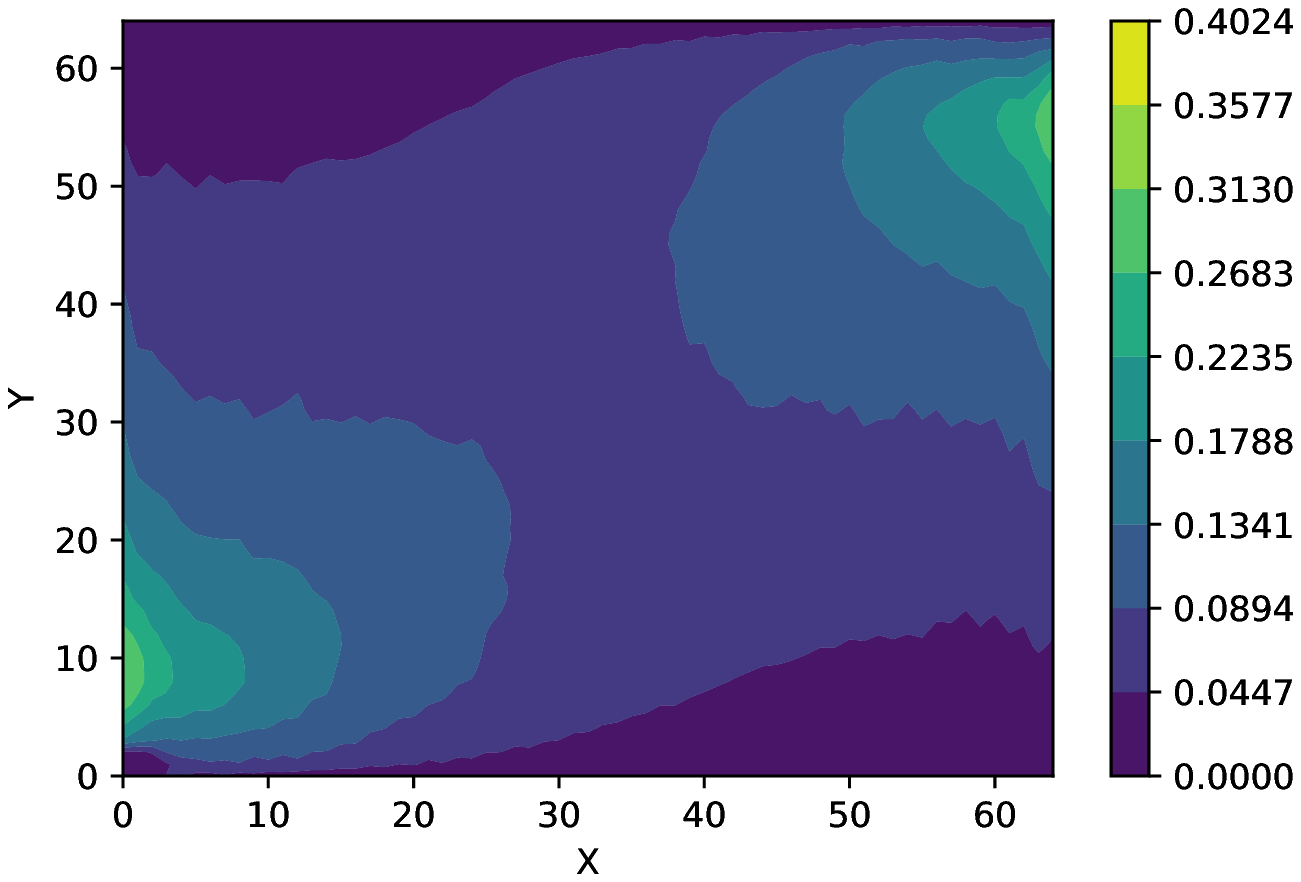}}}%
    \hspace{0.2cm}
    \subfloat[\centering Predicted Variance Velocity(Y)]{{\includegraphics[width=0.3\textwidth]{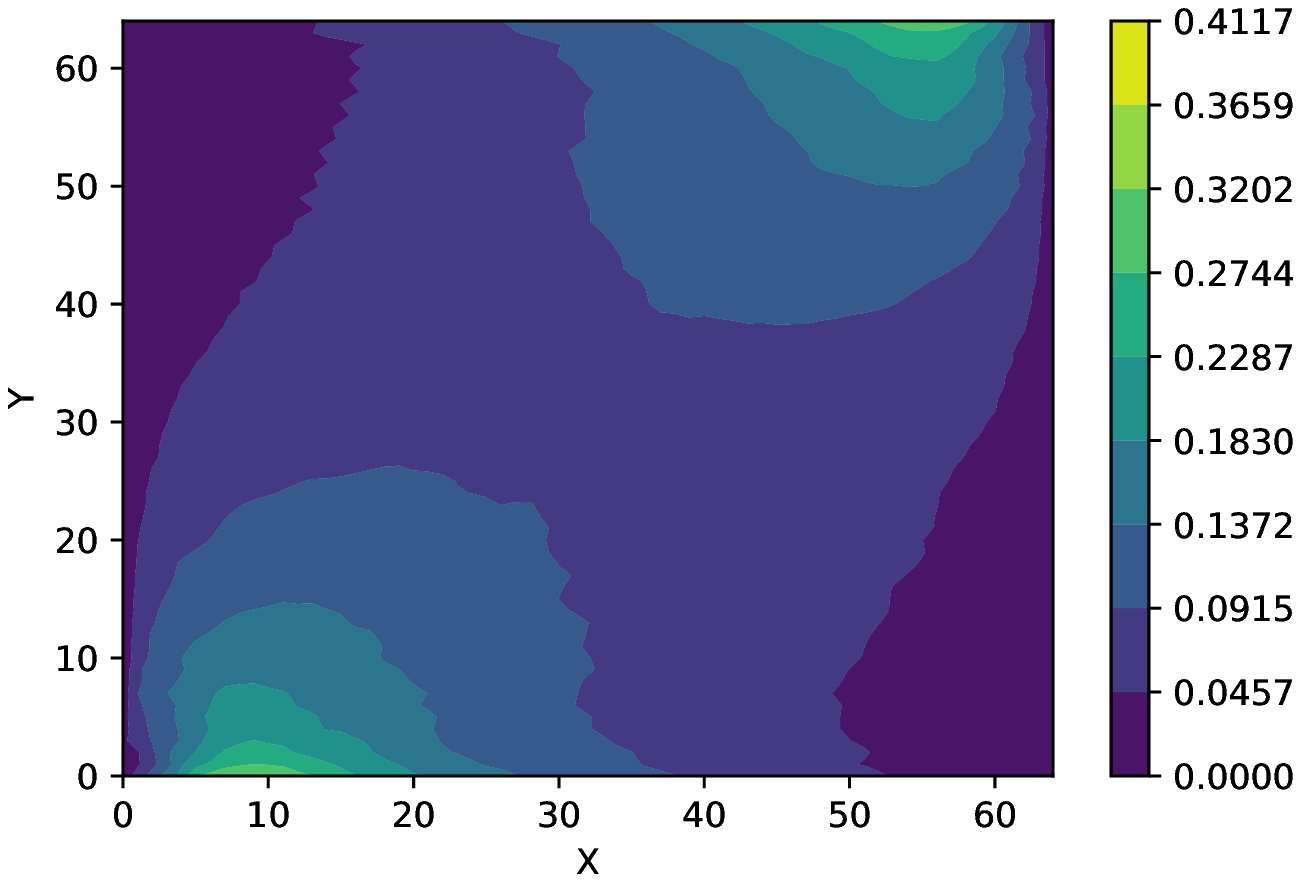}}}%
    \caption{Aleatoric uncertainty propagation of GLU-Net for KLE4225 trained on 512 training samples. For each row descriptions, refer to Fig. \ref{fig:alea1}.}%
    \label{fig:alea3}%
\end{figure}

Fig. \ref{fig:alea1} shows the mean and variance for the KLE50 case obtained using the pressure and velocity fields obtained using crude MCS and the proposed approach. The GLU-net results correspond to the case where 256 training samples are used. 
For almost all the cases, the mean and variances obtained using the proposed GLU-net are in excellent agreement with the benchmark results obtained using MCS; the only exception being the variance of the pressure field. This is expected as the $R^2-$score for pressure field was found to be relatively lower in Fig. \ref{fig:r2-score}. Nonetheless, the zone of concentration at the bottom left, and top right corners are correctly identified even for the variance prediction. 
Figs. \ref{fig:alea2} and \ref{fig:alea3} show the mean and variance of the pressure and velocity fields obtained using MCS and the proposed approach. For both cases, 512 training samples have been used. Again, the proposed approach yields result having an excellent match with the benchmark results obtained using crude MCS. For quantitative assessment, the $R^2-$score for mean and variance predictions are reported in Table \ref{tab:tab1}. For the mean prediction, we obtain an almost perfect prediction with $R^2-$score in the range of 0.99. For the standard deviation of velocity fields also, $R^2-$score in the range 0.94 - 0.97 is obtained. However, for the standard deviation of the pressure field, the $R^2-$score (0.76-0.84) is on a relatively lower side.

\begin{table}[]
    \centering
    \caption{${R^2}$-Scores of the Aleatoric Uncertainty across all dimensions}
    \label{tab:tab1}
    \begin{tabular}{llllllll}
    \hline
    \textbf{Data} &\textbf{${\mu _p}$} & \textbf{$\sigma _p^2$} &\textbf{${\mu _{{u_x}}}$} & \textbf{$\sigma _{{u_x}}^2$} &\textbf{${\mu _{{u_y}}}$} & \textbf{$\sigma _{{u_y}}^2$}  \\ \hline
    KLE50 - 256   & 0.99&0.81 &0.99 &0.97 &0.99 &0.98\\
    KLE500 - 512  & 0.99&0.76 &0.99 &0.97 &0.98 &0.97\\
    KLE4225 - 512 & 0.98&0.84 &0.99 &0.94 &0.99 &0.94\\ \hline 
    \end{tabular}
\end{table}

\begin{figure}[h!]
    \centering
    \subfloat[\centering Pressure (64 training samples)]{{\includegraphics[width=0.3\textwidth]{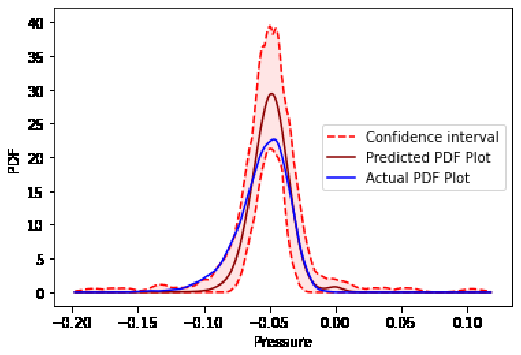}}}%
    \hspace{0.2cm}
    \subfloat[\centering X-component of velocity (64 training samples)]{{\includegraphics[width=0.3\textwidth]{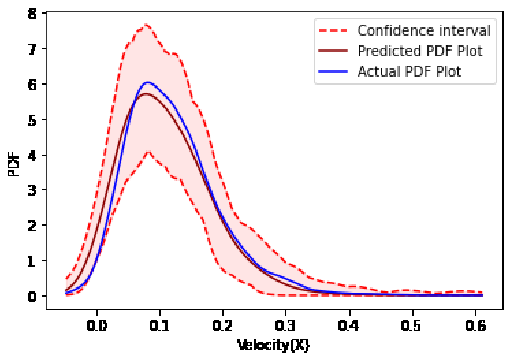}}}%
    \hspace{0.2cm}
    \subfloat[\centering Y-component of velocity (clu training samples)]{{\includegraphics[width=0.3\textwidth]{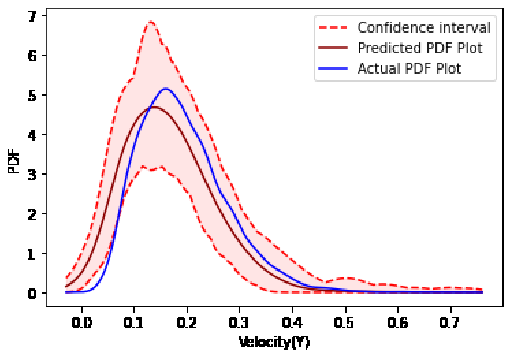}}}%
    \hspace{0.2cm}
    \subfloat[\centering Pressure (256 training samples)]{{\includegraphics[width=0.3\textwidth]{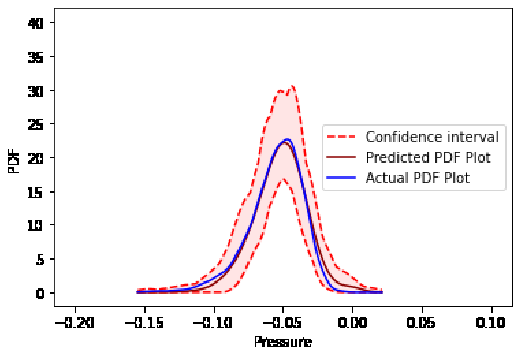}}}%
    \hspace{0.2cm}
    \subfloat[\centering X-component of velocity (256 training samples)]{{\includegraphics[width=0.3\textwidth]{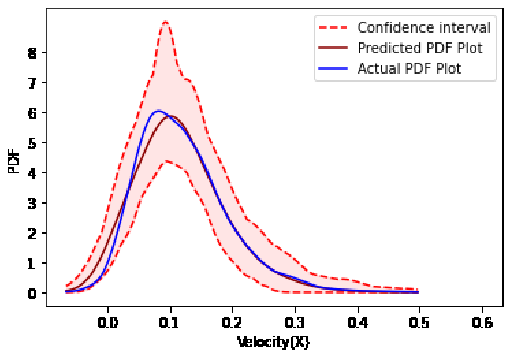}}}%
    \hspace{0.2cm}
    \subfloat[\centering Y-component of velocity (256 training samples)]{{\includegraphics[width=0.3\textwidth]{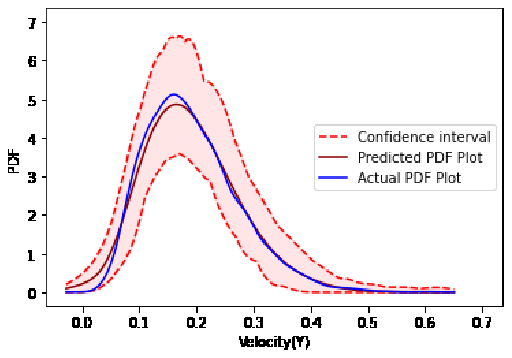}}}%
    \caption{PDF of pressure and velocity for KLE50 using the proposed GLU-net and MCS. The first row corresponds to results obtained using GLU-net trained with 64 training samples. The second row corresponds to results obtained using GLU-net trained with 256 training samples.}%
    \label{fig:PDF 1}%
\end{figure}

Finally, we focus on the capability of the proposed GLU-net in accurately predicting the PDF of the pressure and velocities. The procedure is mostly similar to that of obtaining the mean and standard deviation of contours; however, in this case, we also illustrate the epistemic uncertainty in the trained GLU-net model. Fig. \ref{fig:PDF 1} shows the PDF of the pressure and velocities for KLE50 at pixel location $(0.77,0.6)$ obtained using crude MCS and the proposed GLU-net. Two cases where the GLU-net model is trained using 64 and 256 training samples are shown. While both the GLU-net models yield reasonable results, the one trained with 256 training  samples is comparatively superior and matches almost exactly with the MCS results. The shaded portion indicates the epistemic uncertainty due to limited and noisy data.   

\begin{figure}[h!]
    \centering
    \subfloat[\centering Pressure (128 training samples)]{{\includegraphics[width=0.3\textwidth]{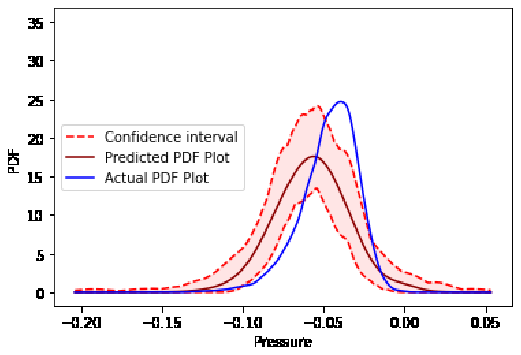}}}%
    \hspace{0.2cm}
    \subfloat[\centering X-component of velocity (128 training samples)]{{\includegraphics[width=0.3\textwidth]{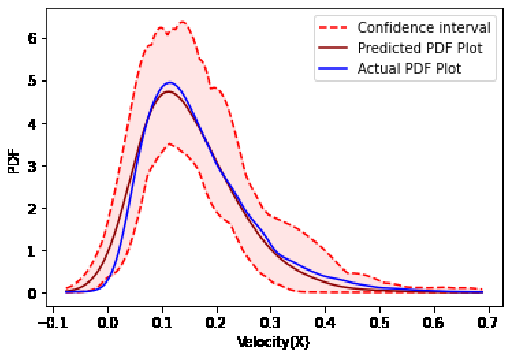}}}%
    \hspace{0.2cm}
    \subfloat[\centering Y-component of velocity (128 training samples)]{{\includegraphics[width=0.3\textwidth]{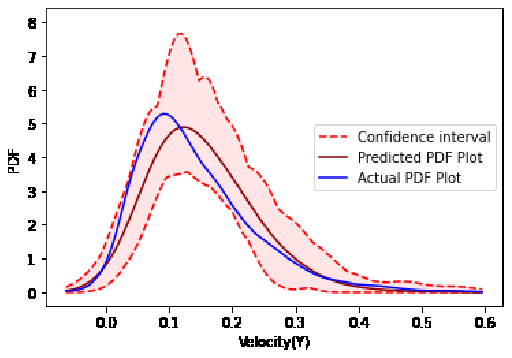}}}%
    \hspace{0.2cm}
    \subfloat[\centering Pressure (512 training samples)]{{\includegraphics[width=0.3\textwidth]{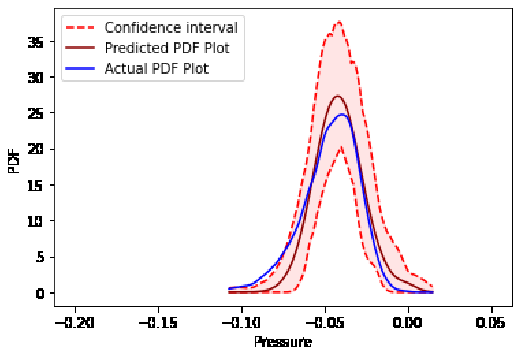}}}%
    \hspace{0.2cm}
    \subfloat[\centering X-component of velocity (512 training samples)]{{\includegraphics[width=0.3\textwidth]{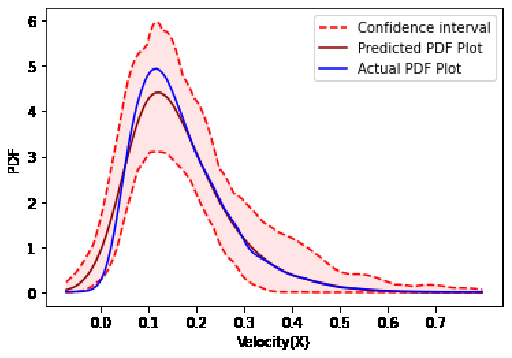}}}%
    \hspace{0.2cm}
    \subfloat[\centering Y-component of velocity (512 training samples)]{{\includegraphics[width=0.3\textwidth]{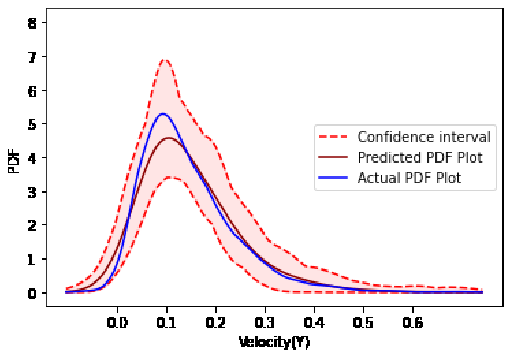}}}%
    \caption{PDF of pressure and velocity for KLE500 using the proposed GLU-net and MCS. The first row corresponds to results obtained using GLU-net trained with 128 training samples. The second row corresponds to results obtained using GLU-net trained with 512 training samples.}%
    \label{fig:PDF 2}%
\end{figure}

\begin{figure}[t!]%
    \centering
    \subfloat[\centering Pressure (128 training samples)]{{\includegraphics[width=0.3\textwidth]{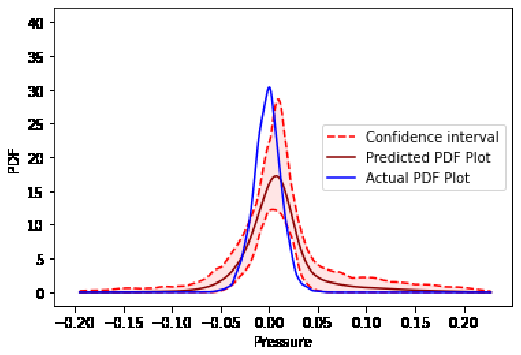}}}%
    \hspace{0.2cm}
    \subfloat[\centering X-component of velocity (128 training samples)]{{\includegraphics[width=0.3\textwidth]{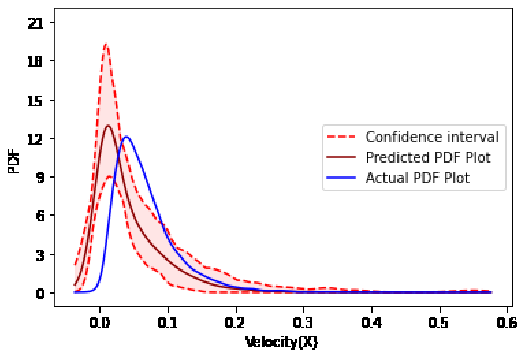}}}%
    \hspace{0.2cm}
    \subfloat[\centering Y-component of Velocity (128 training samples)]{{\includegraphics[width=0.3\textwidth]{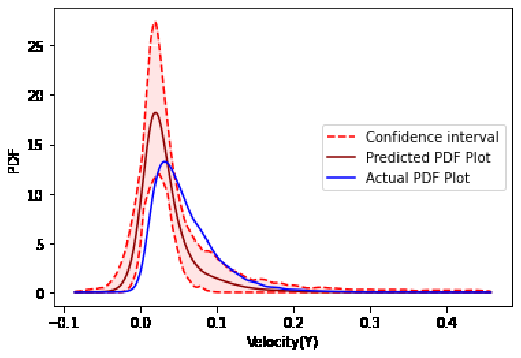}}}%
    \hspace{0.2cm}
    \subfloat[\centering Pressure (512 training samples)]{{\includegraphics[width=0.3\textwidth]{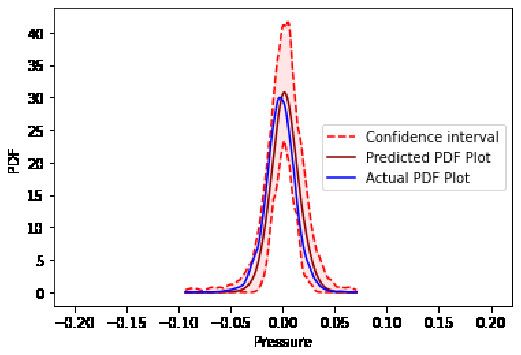}}}%
    \hspace{0.2cm}
    \subfloat[\centering X-component of velocity (512 training samples)]{{\includegraphics[width=0.3\textwidth]{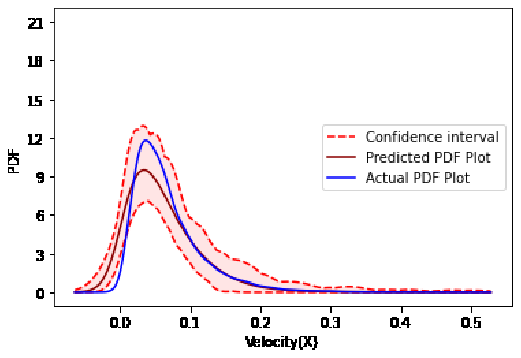}}}%
    \hspace{0.2cm}
    \subfloat[\centering Y-component of velocity (512 training samples)]{{\includegraphics[width=0.3\textwidth]{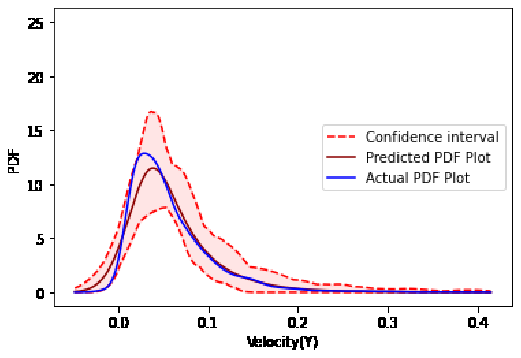}}}%
    \caption{PDF of pressure and velocity for KLE4225 using the proposed GLU-net and MCS. The first row corresponds to results obtained using GLU-net trained with 128 training samples. The second row corresponds to results obtained using GLU-net trained with 512 training samples.}%
    \label{fig:PDF 3}%
\end{figure}

Figs. \ref{fig:PDF 2} and \ref{fig:PDF 3} illustrate the PDF of the pressure and velocities for KLE500 and KLE4225, respectively. For both cases, results corresponding to GLU-net trained with 128 and 512 training samples are presented. For KLE500, we observe that the GLU-net with 128 training samples yields overconfident prediction as indicated by the fact that the true PDF for pressure resides outside the confidence interval. GLU-net with 512 training samples yields superior results. Similar observations also hold for the PDFs corresponding to the KLE4225.

\section{Conclusion}\label{sec:conc}
In this paper, we propose a novel deep learning architecture for solving high-dimensional uncertainty quantification problems. The proposed approach, referred to as the Gated Linear Network induced U-net (GLU-net), combines the well-known U-net architecture with the Gaussian Gated Linear Network (GGLN). The proposed approach treats the underlying problem as an image to image regression problem and hence, is highly efficient. One of the primary advantages of the proposed architecture resides in its inherent capability to quantify uncertainty due to noisy and limited data. Additionally, the architecture of GLU-net is significantly less complex with fewer parameters than the contemporary state-of-the-art architectures available in the literature. 

The proposed approach is used for solving the Darcy flow problem with a random permeability field. For controlling the intrinsic dimensionality of the problem, Karhunen Loève expansion with 50, 500, and 4225 stochastic dimensions have been considered. Additionally, case studies with a different number of training samples have been carried out. For all the cases, the proposed GLU-net yields reasonable results for the pressure and the velocity fields with $R^2-$score of 0.9 and above. The proposed approach is also robust, as indicated by convergence studies carried out by increasing the number of training samples. Overall, the proposed approach does an excellent job of quantifying the propagation of uncertainty from the input to the output.

Despite the excellent performance in quantifying the uncertainty propogation, the proposed GLU-net is limited to regular grids only, because the proposed approach uses a convolutional block. One possible solution is to use a graph neural network. Also, among the three fields for the Darcy flow problem, the performance of the proposed approach in predicting the pressure field is relatively bad. This can be addressed by using a separate network for the pressure.

\section*{Acknowledgment}\label{sec:ack}
SC acknowledges the financial support received from IIT Delhi in form of seed grant.

\appendix
\section{Gaussian Gated Linear Network}\label{app:1}
\subsection{Weight Update and Weight Projection}
This subsection is not used in our work, because we use G-GLN module under a larger architecture. But we include them under our code base and are points to be kept noted when G-GLN is used as a single module directly(when we know about the target $y$ and the input $x$, being sent to the G-GLN). Now, for the weight update following the gradient descent, let us get the derivative of the vector form of loss function(with $\mu$ and $\eta$ vectors) with a particular weight $\eta_i$,
\begin{equation}\label{eq:eq17}
   \frac{{\partial l(y;\cdot)}}{{\partial {\eta _i}}} =  - ||\eta ||_1^{ - 1} + (y - {\eta ^T}\mu /||\eta |{|_1})(y - 2{\mu _i} + {\eta ^T}\mu /||\eta |{|_1})    
\end{equation}
but we need $\frac{{\partial l(y; \cdot )}}{{\partial {w_i}}}$, which can be directly obtained by the definition of $\eta$ vector, and using expressions of $\mu_PoG$ and $\sigma _{PoG}^2$ so the final expression would be, 
\begin{equation}\label{eq:eq18}
{\nabla _w}l(y;w) = diag(\frac{1}{{{\sigma ^2}}})[(y - {\mu _{PoG}})({1_{m,1}}(y + {\mu _{PoG}} - 2\mu ) - {1_{m,1}}\sigma _{PoG}^2]    
\end{equation}
where ${1_{a,b}}$ represents an identity matrix of shape $(a,b)$.

Now, a selected weight for a given neuron at a layer $w_i$ can be updated as,
\begin{equation}\label{eq:eq19}
{w_{i,t}} = PROJ({w_{i,t - 1}} - \eta \nabla {l_i}(y;{x'})    
\end{equation}
where $\eta$ is a learning rate defined for the G-GLN, ${x'}$ is the context input and $\nabla {l_i}(y;{x'})$ is the gradient $\frac{{\partial {l_i}(y;{x'})}}{{\partial {w_i}}}$, $t$ being the time of updation and PROJ is the projection of the weight into the set $W$,$W: = \{ w \in {[0,b]^m}:||w|{|_1} \ge \varepsilon \}$ defined in the Gaussian Mixing section. PROJ can be rewritten as the below three sets of conditions, weight being in the set $[0,b]$, ${\mu _{PoG}} \in [{\mu _{\min }},{\mu _{\max }}]$ and $\sigma _{PoG}^2 \in [\sigma _{\min }^2,\sigma _{\max }^2]$, obtained from the PoG expressions. Finally, the authors of \cite{veness2019gated} allow negative constraints on the weights being ${w_i} \in [ - 1000,1000]$ and only other constraint on $\sigma _{PoG}^2$ being $\sigma _{PoG}^2 \in [0.5,{10^5}]$. Having seen the trained weights, we expect that the absolute value of each weight isn't much higher because of our input and the output being close to 0.

\subsection{Convexity of the Loss function under weight constraints}\label{app:app2}
The Hessian matrix of a convex function is positive semi-definite(PSD). Let us consider $g(\eta ) = {\eta ^T}\mu /||w|{|_1}$, then, ${g'}(\eta ) = \frac{{\partial g}}{{\partial {\eta _j}}} = {\mu _j}||\eta ||_1^{ - 1} - {\eta ^T}\mu ||\eta ||_1^{ - 2}$. $\eta$ is already defined as ${\eta _i}: = {w_i}/\sigma _i^2,\eta = ({\eta_1},...{\eta_m})$. Now, for the second derivative, upon calculation gives
\begin{equation}\label{eq:eq20}
\frac{{{\partial ^2}l(y,\cdot)}}{{\partial {\eta _i}\partial {\eta _j}}} = ||\eta ||_1^{ - 2} + 2||\eta ||_1^{ - 1}\left( {{\mu _j} - {\eta ^T}\mu /||\eta |{|_1}} \right)\left( {{\mu _i} - {\eta ^T}\mu /||\eta |{|_1}} \right)    
\end{equation}
Now, the Hessian matrix would be the general form of the above, 
\begin{equation}\label{eq:eq21}
{\nabla ^2}l(y;\eta ) = ||\eta ||_1^{ - 2}{1_{m,m}} + w||\eta ||_1^{ - 1}\left( {\mu  - {\eta ^T}\mu /||\eta |{|_1}{1_{m,1}}} \right){\left( {{\mu _i} - {\eta ^T}\mu /||\eta |{|_1}} \right)^T}    
\end{equation}
The first term of Eq \ref{eq:eq21} is PSD. Now, for the second term $\left( {\mu  - {\eta ^T}\mu /||\eta |{|_1}{1_{m,1}}} \right){\left( {{\mu _i} - {\eta ^T}\mu /||\eta |{|_1}} \right)^T}$ is PSD by letting $a = \left( {\mu  - {\eta ^T}\mu /||\eta |{|_1}{1_{m,1}}} \right)$ and ${u^T}a{a^T}u = {(u\cdot a)^2} \ge 0$, $\forall u \in {\Re ^m}$.Now, sum of these two PSD matrixes is also PSD. Now since $l(y;\eta)$ is convex in $\eta$ which can be extended to  $l(y;w)$ being $w$.

\end{document}